%% file: medima-template.tex
\documentclass[times,twocolumn,final,longtitle,nonatbib,numafflabel]{elsarticle}

\AtBeginDocument{%
  \def\thefnote{\myfnsymbol{fnote}}}
\makeatletter
\def\myfnsymbol#1{\expandafter\@myfnsymbol\csname c@#1\endcsname}
\def\@myfnsymbol#1{\ifcase #1\or $\dagger$\or $\#$\else \@ctrerr\fi}
\def\fntext[#1]#2{\g@addto@macro\@fnotes{%
   \refstepcounter{fnote}\elsLabel{#1}%
   \def\thefootnote{\thefnote}
   \global\setcounter{footnote}{\c@fnote}%
   \footnotetext{#2}}}
\makeatother

\usepackage{medima}
\usepackage{framed,multirow}

\usepackage{amssymb}
\usepackage{latexsym}

\usepackage{pifont}
\newcommand{\cmark}{\ding{51}}%
\newcommand{\xmark}{\ding{55}}%

\usepackage{url}
\usepackage[table]{xcolor}

\usepackage{hyperref}

\usepackage{makecell}

\definecolor{goldenyellow}{rgb}{1.0, 0.87, 0.0}
\definecolor{bronze}{rgb}{0.8, 0.5, 0.2}
\definecolor{silver}{rgb}{0.25, 0.28, 0.33}  

\usepackage{newtxtext} 

\newcommand{\mainheading}[1]{\vspace{1em}\noindent\textbf{\fontsize{12pt}{14pt}\selectfont #1}\vspace{1em}}

\makeatletter
\renewcommand\section{\@startsection {section}{1}{\z@}%
  {12pt \@plus 4pt \@minus 2pt}%
  {6pt \@plus 2pt \@minus 2pt}%
  {\normalfont\bfseries\fontsize{10pt}{12pt}\selectfont}}
\makeatother

\usepackage{titletoc}

\usepackage{tocloft}

\usepackage{caption}
\DeclareCaptionFont{tenpt}{\fontsize{10}{12}\selectfont}
\usepackage{etoolbox}
\AtBeginEnvironment{tabular}{\fontsize{8}{10}\selectfont}
\AtBeginEnvironment{tabularx}{\fontsize{8}{10}\selectfont}
\AtBeginEnvironment{longtable}{\fontsize{8}{10}\selectfont}

\usepackage{setspace}

\makeatletter
\let\c@author\relax
\makeatother

\usepackage[style=vancouver,backend=biber,maxnames=6,minnames=3]{biblatex}
\AtEveryBibitem{%
  \ifboolexpr{
    test {\ifnumgreater{\value{listcount}}{3}}
    and
    test {\ifnumless{\value{listcount}}{7}}
  }
  {\printnames{author}}
  {}
}

\journal{NEJM AI}

\begin{document}

\singlespacing

\verso{Yang, Musio, and Ma \textit{et~al.}}

\begin{frontmatter}

\title{The TopCoW Challenge --- Topology-Aware Circle of Willis Segmentation for CT and MR Angiography}

\input{sections/authors.tex}


\begin{abstract}
\textbf{Background:}
The Circle of Willis (CoW) is an important network of arteries connecting major circulations of the brain. Its vascular architecture is believed to influence the risk, severity, and outcome of serious neurovascular diseases. However, characterizing the highly variable CoW anatomy remains a manual and time-consuming expert task. The CoW is commonly imaged by two non-invasive angiographic imaging modalities --- magnetic resonance angiography (MRA) and computed tomography angiography (CTA) --- yet few datasets with annotated CoW anatomy exist, and there have been no established benchmarks for comparing CoW segmentation algorithms.
\\\textbf{Methods:}
We organized the TopCoW benchmark challenge alongside the release of an annotated CoW dataset with 125 paired MRA and CTA scans from the same patients. Voxel-level annotations for 13 vessel components were created using virtual reality technology and verified by clinical experts. Participants submitted algorithms for CoW segmentation and variant classification, which we evaluated on internal and external test sets comprising 226 scans from over five centers. The benchmark includes voxel-level segmentation, CoW component detection, CoW variant classification, and two clinical application tasks.
\\\textbf{Results:}
We received submissions from over 250 participants across six continents. Top-performing teams achieved over 90\% Dice scores for CoW segmentation, over 80\% F1 scores for detecting key vessel components, and over 70\% balanced accuracy in CoW variant classification across nearly all test sets. The best algorithms also supported clinically relevant downstream tasks by accurately classifying fetal-type posterior cerebral arteries and localizing aneurysms in relation to CoW anatomy. Notably, modality-agnostic models and topological optimization emerged as key factors in improving segmentation performance.
\\\textbf{Conclusions:}
This benchmark demonstrated the utility of CoW segmentation algorithms for some downstream clinical applications with explainability. The annotated datasets and top-performing algorithms are published alongside this benchmark to foster methodological advances and clinical tool development.
\end{abstract}


\end{frontmatter}


\clearpage


\begin{refsection}

\input{sections/introduction.tex}

\input{sections/material_n_methods.tex}

\input{sections/results.tex}

\input{sections/discussions.tex}




\section*{Data Availability}
\label{sec:data}
The TopCoW training dataset, consisting of 250 annotated images, and our multi-center test sets of 86 annotated images are available in our public Zenodo repository (15 GB) at \url{https://zenodo.org/records/15692630}.

\section*{Code Availability}
\label{sec:code}
The Docker images from the best-performing teams, along with the scripts to facilitate local execution, are available in our public Zenodo repository (45 GB) at \url{https://zenodo.org/records/15665435}.

For code availability of individual team submissions, please see Supplementary \ref{suppl:all_methods}.

The implementation of our evaluation metric code is open sourced at \url{https://github.com/CoWBenchmark/TopCoW_Eval_Metrics}.

\section*{Author Contributions}

\textbf{Contributor Roles}:
(1) Conceptualization; (2) Data curation; (3) Formal analysis; (4) Funding acquisition; (5) Investigation; (6) Methodology; (7) Project administration; (8) Resources; (9) Software; (10) Supervision; (11) Validation; (12) Visualization; (13) Writing - original draft; (14) Writing - review \& editing.
\textbf{Authors (abbreviated, in manuscript order) with CRediT roles in parentheses:} 
KY (1,2,3,5,6,9,13), FM (1,2,3,5,6,9,13), YM (1,2,6,11,13), NJ (1,2,4,5,8,14), JCP (6,12,14), RAM (6,12), LH (6,12), HBL (6,14), IEH (6,8), AS (6,14), SS (6,9), HH (6,9,14), CP (6,9), EDLR (6,8,14), BW (6,14), DW (6,9), FK (6,9), FN (6,9), MJM (6,9), IE (6), DR (8,10), INV (6,8), YMR (8,14), BKV (8), HJK (8), PS (9,14), WL (9), TM (9), MRR (9), YK (9), FI (9), KMH (9,10), CZ (8,9), HZ (8), PB (1,2,4,6,11,14), JH (6,11), CW (6,11), LW (6,11), JB (6,11), EC (6,11,14), HB (6,11,14), HLH (8), AM (6,11), JH (6,11), AS (11), BW (8,10,11), JSK (8,10,11), EOR (8,11,14), RW (11), EM (9), LLG (9,14), KO (9,14), DL (9), OUA (9,14), AH (9,14), JR (9), DR (9), ST (9), AK (9), DF (9), AQ (9), MM (9), SN (9), ND (9), JCH (9), DL (9,14), FG (9), DF (9), MAZ (9,14), CL (9), HZ (9), ZZ (9), MZ (9,14), XY (9), HZ (9), GZY (9), YG (9), SR (9), JH (9), HP (9), JC (9), MW (9), HM (9), NM (9), FA (9), CY (9), REH (9), CL (9), JS (9), AC (9), XL (9), UMLTE (9), VA (9), LG (9), AO (9), PC (9), AG (9), MD (9,14), OC(9), JL (9), HH (9), YC (9) , ZL (9), YL (9), SZ (9), TRP (9), AHS (9), VMT (9), MO (9), HW (9), MMB (9), YS (9), SH (1,4,8,10), SW (1,4,6,8,11), BM (1,4,8,10,14)

\section*{Acknowledgments}
\input{sections/acknowledgement}

\printbibliography

\end{refsection}


\begin{refsection}

\include{supplement.tex}

\printbibliography[title={Supplementary References}]

\end{refsection}

\end{document}

%% file: sections/authors.tex


\author[uzhDqbm]{Kaiyuan Yang\corref{coFirst}}
\cortext[coFirst]{K.Y., F.M., and Y.M. contributed equally}

    \address[uzhDqbm]{
        Department of Quantitative Biomedicine, University of Zurich, Zurich, Switzerland
    }

\author[uzhDqbm,zhaw]{Fabio Musio\corref{coFirst}}

    \address[zhaw]{
        Institute of Computational Life Sciences, Zurich University of Applied Sciences (ZHAW), Waedenswil, Switzerland
    }

\author[USZneuroRad,whuZhongNan]{Yihui Ma\corref{coFirst}}

    \address[USZneuroRad]{
        Department of Neuroradiology, University Hospital of Zurich, Zurich, Switzerland
    }

    \address[whuZhongNan]{
        Department of Neurosurgery, Zhongnan Hospital of Wuhan University, Wuhan, China
    }

\author[zhaw]{Norman Juchler}

\author[cornellRad]{Johannes C. Paetzold}

    \address[cornellRad]{
        Department of Radiology at Weill Cornell Medicine, Cornell University, New York, USA
    }

\author[iTermHelmholtz,tumCS]{Rami Al-Maskari}

    \address[iTermHelmholtz]{
        Institute for Tissue Engineering and Regenerative Medicine (iTERM), Helmholtz Munich, Neuherberg, Germany
    }
    \address[tumCS]{
       School of Computation, Information and Technology, Technical University of Munich, Germany
    }

\author[iTermHelmholtz]{Luciano Höher}

\author[uzhDqbm,harvardMedicalSchool]{Hongwei Bran Li}

    \address[harvardMedicalSchool]{
        Athinoula A. Martinos Center for Biomedical Imaging, Harvard Medical School, Boston, USA
    }

\author[uzhDqbm]{Ibrahim Ethem Hamamci}
\author[uzhDqbm]{Anjany Sekuboyina}

\author[uzhDqbm]{Suprosanna Shit}

\author[uzhDqbm]{Houjing Huang}

\author[uzhDqbm]{Chinmay Prabhakar}

\author[uzhDqbm]{Ezequiel de la Rosa}

\author[uzhDqbm]{Bastian Wittmann}

\author[uzhDqbm,tumCS]{Diana Waldmannstetter}

\author[uzhDqbm,tumCS,tumMed,HelmholtzAI]{Florian Kofler}

    \address[tumMed]{
        School of Medicine and Health, TUM Klinikum, Technical University of Munich, Germany
    }
    \address[HelmholtzAI]{
        Helmholtz AI, Helmholtz Munich, Neuherberg, Germany
    }

\author[uzhDqbm,tumCS,tumMed]{Fernando Navarro}

\author[tumCS,tumMCML,imperialCollege]{Martin J. Menten}

    \address[tumMCML]{
        Munich Center for Machine Learning, Munich, Germany
    }
    \address[imperialCollege]{
        Department of Computing, Imperial College London, London, UK
    }

\author[tumCS]{Ivan Ezhov}

\author[tumCS,tumMed,tumMCML,imperialCollege]{Daniel Rueckert}

\author[crownUMCISI]{Iris N. Vos}

    \address[crownUMCISI]{
        Image Sciences Institute, UMC Utrecht, Utrecht, The Netherlands
    }

\author[crownUMCNeuro]{Ynte M. Ruigrok}

    \address[crownUMCNeuro]{
        Department of Neurology and Neurosurgery, University Medical Center Utrecht, Utrecht, The Netherlands
    }

\author[crownUMCRad]{Birgitta K. Velthuis}

    \address[crownUMCRad]{
        Department of Radiology, University Medical Center Utrecht, Utrecht, The Netherlands
    }

\author[crownUMCISI]{Hugo J. Kuijf}

\author[Team_NexToU]{Pengcheng Shi}
\author[Team_NexToU]{Wei Liu}
\author[NewTingMaPlace]{Ting Ma}

    \address[Team_NexToU]{
        Electronic \& Information Engineering School, Harbin Institute of Technology (Shenzhen), Shenzhen, China
    }
    \address[NewTingMaPlace]{
        Biomedical Engineering School, Harbin Institute of Technology (Shenzhen), Shenzhen, China
    }

\author[Team_DKFZ_Dkfz,Team_DKFZ_Math]{Maximilian R. Rokuss}
\author[Team_DKFZ_Dkfz,Team_DKFZ_Math,Team_DKFZ_HIDSS]{Yannick Kirchhoff}
\author[Team_DKFZ_Dkfz,Team_DKFZ_HI]{Fabian Isensee}
\author[Team_DKFZ_Dkfz,Team_DKFZ_Rad]{Klaus Maier-Hein}
    \address[Team_DKFZ_Dkfz]{
        Division of Medical Image Computing, German Cancer Research Center (DKFZ), Heidelberg, Germany
    }
    \address[Team_DKFZ_Math]{
        Faculty of Mathematics and Computer Science, Heidelberg University, Germany
    }
    \address[Team_DKFZ_HIDSS]{
        HIDSS4Health - Helmholtz Information and Data Science School for Health, Karlsruhe/Heidelberg, Germany
    }
    \address[Team_DKFZ_HI]{
        Helmholtz Imaging, German Cancer Research Center, Heidelberg, Germany
    }
    \address[Team_DKFZ_Rad]{
        Pattern Analysis and Learning Group, Department of Radiation Oncology, Heidelberg University Hospital
    }

\author[Team_UW]{Chengcheng Zhu}

    \address[Team_UW]{
        Department of Radiology, University of Washington, Seattle, WA, USA
    }

\author[Team_UW_SJTU]{Huilin Zhao}

    \address[Team_UW_SJTU]{
        Department of Radiology, Ren Ji Hospital, Shanghai Jiao Tong University School of Medicine, Shanghai, China
    }

\author[genevaHUG]{Philippe Bijlenga\fnref{clinician}}
\author[genevaHUG]{Julien Hämmerli\fnref{clinician}}
\fntext[clinician]{Clinical committee}

    \address[genevaHUG]{
        Department of Clinical Neurosciences, Division of Neurosurgery, Geneva University Hospitals, Geneva,
        Switzerland
    }

\author[genevaHUG]{Catherine Wurster\fnref{clinician}}

\author[uszNeurology]{Laura Westphal\fnref{clinician}}

    \address[uszNeurology]{
        Department of Neurology, University Hospital of Zurich, Zurich, Switzerland
    }

\author[joeroenAffliation]{Jeroen Bisschop\fnref{clinician}}

    \address[joeroenAffliation]{
        Department of Physiology, University of Toronto, Canada
    }

\author[uszNeuroSurgery]{Elisa Colombo\fnref{clinician}}

    \address[uszNeuroSurgery]{
        Department of Neurosurgery, University Hospital of Zurich, Zurich, Switzerland
    }

\author[uszNeurology]{Hakim Baazaoui\fnref{clinician}}
\author[uszNeurology]{Hannah-Lea Handelsmann\fnref{clinician}}

\author[singaporeNUH]{Andrew Makmur\fnref{clinician}}

    \address[singaporeNUH]{
        Department of Diagnostic Imaging, National University Hospital, Singapore
    }

\author[singaporeNUH]{James Hallinan\fnref{clinician}}

\author[amrish]{Amrish Soundararajan\fnref{clinician}}

    \address[amrish]{
        University of Chicago, USA
    }

\author[tumMed]{Benedikt Wiestler\fnref{clinician}}

\author[tumMed]{Jan S. Kirschke\fnref{clinician}}

\author[tumMed]{Evamaria O. Riedel\fnref{clinician}}

\author[bernSpital]{Roland Wiest\fnref{clinician}}

    \address[bernSpital]{
        Department of Diagnostic and
        Interventional Neuroradiology, University
        Hospital Berne and University of Berne, Berne,
        Switzerland
    }


\author[Team_2i_mtl]{Emmanuel Montagnon\fnref{participant}}
\fntext[participant]{Participant of the challenge, ordered alphabetically by team name}

    \address[Team_2i_mtl]{
        Centre de Recherche du Centre Hospitalier de l’Université de Montréal (CRCHUM), Montréal, Québec, Canada
    }

\author[Team_2i_mtl]{Laurent Letourneau-Guillon\fnref{participant}}

\author[Team_ARG,Team_ARG_uni]{Kwanseok Oh\fnref{participant}}
\author[Team_ARG]{Dahye Lee\fnref{participant}}

    \address[Team_ARG]{
        DEEPNOID Inc., Seoul, South Korea
    }
    \address[Team_ARG_uni]{
        Department of Artificial Intelligence, Korea University, Seoul, South Korea
    }

\author[Team_CLAIM]{Orhun Utku Aydin\fnref{participant}}
\author[Team_CLAIM]{Adam Hilbert\fnref{participant}}
\author[Team_CLAIM]{Jana Rieger\fnref{participant}}
\author[Team_CLAIM]{Dimitrios Rallios\fnref{participant}}
\author[Team_CLAIM]{Satoru Tanioka\fnref{participant}}
\author[Team_CLAIM]{Alexander Koch\fnref{participant}}
\author[Team_CLAIM]{Dietmar Frey\fnref{participant}}

    \address[Team_CLAIM]{
        Charité Lab for AI in Medicine (CLAIM), Charité Universitätsmedizin Berlin, Berlin, Germany
    }

\author[Team_DeepLearnAI]{Abdul Qayyum\fnref{participant}}
\author[Team_DeepLearnAI2]{Moona Mazher\fnref{participant}}
\author[Team_DeepLearnAI]{Steven Niederer\fnref{participant}}

    \address[Team_DeepLearnAI]{
        National Heart and Lung Institute, Faculty of Medicine, Imperial College London, London, UK
    }
    \address[Team_DeepLearnAI2]{
        Centre for Medical Image Computing, Department of Computer Science, University College London, London, UK
    }

\author[Team_DKFZ_Dkfz,Team_DKFZ_Math,Team_DKFZ_HIDSS]{Nico Disch\fnref{participant}}
\author[Team_DKFZ_Dkfz]{Julius C. Holzschuh\fnref{participant}}

\author[Team_DLaBella29]{Dominic LaBella\fnref{participant}}

    \address[Team_DLaBella29]{
        Department of Radiation Oncology, Duke University Medical Center, Durham, NC, USA
    }

\author[Team_EURECOM]{Francesco Galati\fnref{participant}}

    \address[Team_EURECOM]{
        EURECOM, Biot, France
    }

\author[Team_EURECOM]{Daniele Falcetta\fnref{participant}}

\author[Team_EURECOM]{Maria A. Zuluaga\fnref{participant}}

\author[Team_gbCoW]{Chaolong Lin\fnref{participant}}

    \address[Team_gbCoW]{
        Institute of Medical Technology, Peking University Health Science Center, Beijing, China
    }

\author[Team_gbCoW]{Haoran Zhao\fnref{participant}}

\author[Team_gl]{Zehan Zhang\fnref{participant}}

    \address[Team_gl]{
        Hangzhou Genlight MedTech Co., Ltd., China
    }

\author[Team_IMR,DptAutoSJTU]{Minghui Zhang\fnref{participant}}
\author[Team_IMR,DptAutoSJTU]{Xin You\fnref{participant}}
\author[Team_IMR]{Hanxiao Zhang\fnref{participant}}
\author[Team_IMR]{Guang-Zhong Yang\fnref{participant}}
\author[Team_IMR,DptAutoSJTU]{Yun Gu\fnref{participant}}

    \address[Team_IMR]{
        Institute of Medical Robotics, Shanghai Jiao Tong University, Shanghai, China
    }
    \address[DptAutoSJTU]{
        Department of Automation, Shanghai Jiao Tong University, Shanghai, China
    }

\author[Team_IWantToGoToCanada]{Sinyoung Ra\fnref{participant}}

    \address[Team_IWantToGoToCanada]{
        Department of Artificial Intelligence, Sungkyunkwan University, Seoul, South Korea
    }

\author[Team_IWantToGoToCanada]{Jongyun Hwang\fnref{participant}}

\author[Team_IWantToGoToCanada_2]{Hyunjin Park\fnref{participant}}

    \address[Team_IWantToGoToCanada_2]{
        Department of Electrical and Computer Engineering, Sungkyunkwan University, Seoul, South Korea
    }

\author[Team_junqiangchen]{Junqiang Chen\fnref{participant}}

    \address[Team_junqiangchen]{
        Shanghai MediWorks Precision Instruments Co., Ltd., China
    }

\author[Team_lWM,lWM_poland]{Marek Wodzinski\fnref{participant}}
\author[Team_lWM]{Henning Müller\fnref{participant}}

    \address[Team_lWM]{
        Institute of Informatics, HES-SO Valais-Wallis, Switzerland
    }

    \address[lWM_poland]{
        Department of Measurement and Electronics, AGH University of Krakow, Poland
    }

\author[Team_Nantes,Nantes2]{Nesrin Mansouri\fnref{participant}}
\author[Team_Nantes,Nantes2]{Florent Autrusseau\fnref{participant}}

    \address[Team_Nantes]{
        Institut du Thorax (ITX), Université Nantes, Nantes, France
    }

    \address[Nantes2]{
        Laboratoire de Thermique et Energie de Nantes (LTeN), Université Nantes, Polytech’Nantes, Nantes, France
    }


\author[Team_NIC-VICOROB]{Cansu Yalcin\fnref{participant}}

    \address[Team_NIC-VICOROB]{
        Research Institute of Computer Vision and Robotics (ViCOROB), Universitat de Girona, Catalonia, Spain
    }

\author[Team_NIC-VICOROB]{Rachika E. Hamadache\fnref{participant}}
\author[Team_NIC-VICOROB]{Clara Lisazo\fnref{participant}}
\author[Team_NIC-VICOROB]{Joaquim Salvi\fnref{participant}}
\author[Team_NIC-VICOROB]{Adrià Casamitjana\fnref{participant}}
\author[Team_NIC-VICOROB]{Xavier Lladó\fnref{participant}}

\author[Team_NIC-VICOROB]{Uma Maria Lal-Trehan Estrada\fnref{participant}}
\author[Team_NIC-VICOROB]{Valeriia Abramova\fnref{participant}}

\author[VICOROB_Houston]{Luca Giancardo\fnref{participant}}

    \address[VICOROB_Houston]{
Center for Precision Health, McWilliams School of Biomedical Informatics, University of Texas Health Science Center at Houston, USA
    }

\author[Team_NIC-VICOROB]{Arnau Oliver\fnref{participant}}

\author[Team_pamaad_A]{Paula Casademunt\fnref{participant}}
\author[Team_pamaad_A]{Adrian Galdran\fnref{participant}}
\author[zhaw,Team_pamaad_M]{Matteo Delucchi\fnref{participant}}
\author[Team_pamaad_A]{Oscar Camara\fnref{participant}}

    \address[Team_pamaad_A]{
        Physense, BCN-Medtech, Department of Communication and Information Technologies, Universitat Pompeu Fabra, Barcelona, Spain
    }
    \address[Team_pamaad_M]{
        Department of Mathematical Modeling and Machine Learning, University of Zurich, Zurich, Switzerland
    }

\author[Team_refrain_1,Team_refrain_2]{Jialu Liu\fnref{participant}}

    \address[Team_refrain_1]{
        Laboratory of Brain Atlas and Brain-inspired Intelligence, Institute of Automation, Chinese Academy of Sciences, Beijing, China
    }

    \address[Team_refrain_2]{
        School of Artificial Intelligence, University of Chinese Academy of Sciences, Beijing, China
    }

\author[Team_refrain_1,Team_refrain_2]{Haibin Huang\fnref{participant}}
\author[Team_refrain_1,Team_refrain_2]{Yue Cui\fnref{participant}}

\author[Team_sjtu]{Zehang Lin\fnref{participant}}

    \address[Team_sjtu]{
        School of Computer and Information Engineering, Xiamen University of Technology, Xiamen, China
    }

\author[DptAutoSJTU]{Yusheng Liu\fnref{participant}}

\author[Team_sjtu]{Shunzhi Zhu\fnref{participant}}

\author[Team_UBVTL_canon,Team_UBVTL_neuro]{Tatsat R. Patel\fnref{participant}}
\author[Team_UBVTL_canon,Team_UBVTL_neuro]{Adnan H. Siddiqui\fnref{participant}}
\author[Team_UBVTL_canon,Team_UBVTL_patho]{Vincent M. Tutino\fnref{participant}}
    \address[Team_UBVTL_canon]{
        Canon Stroke and Vascular Research Center, University at Buffalo, NY, USA
    }
    \address[Team_UBVTL_neuro]{
        Department of Neurosurgery, University at Buffalo, NY, USA
    }
    \address[Team_UBVTL_patho]{
        Department of Pathology and Anatomical Sciences, University at Buffalo, NY, USA
    }

\author[Team_UW]{Maysam Orouskhani\fnref{participant}}
\author[Team_UW]{Huayu Wang\fnref{participant}}
\author[Team_UW]{Mahmud Mossa-Basha\fnref{participant}}

\author[Team_ysato]{Yuki Sato\fnref{participant}}
    \address[Team_ysato]{
        LPIXEL Inc., Tokyo, Japan
    }

\author[zhaw]{Sven Hirsch\corref{coLast}}
\cortext[coLast]{S.H., S.W., and B.M. are co-corresponding authors}
\ead{hirc@zhaw.ch}

\author[uszNeurology]{Susanne Wegener\corref{coLast}}
\ead{susanne.wegener@usz.ch}

\author[uzhDqbm]{Bjoern Menze\corref{coLast}}
\ead{bjoern.menze@uzh.ch}


%% file: sections/introduction.tex
\section{Introduction}
\label{sec:intro}

The Circle of Willis (CoW) is an important anastomotic network of arteries connecting the anterior and posterior circulations of the brain, as well as the left and right cerebral hemispheres.\supercite{osborn2013osborn} Due to its centrality, the CoW is commonly involved in pathologies like aneurysms and stroke. Clinically, the vascular architecture of the CoW is believed to impact the occurrence and severity of stroke,\supercite{liebeskind2003collateral,chuang2013configuration,van2015completeness,kim2016BMJ} pose a potential risk for aneurysm formation,\supercite{rinaldo2016relationship,hindenes2023anatomical} and affect the neurologic events and clinical outcomes of neurosurgeries.\supercite{yang2017relationship,banga2018incomplete} Accurate CoW characterization is therefore of great clinical relevance.

Clinicians have articulated an unmet demand for efficient software tools to analyze the angio-architecture of the CoW. Assessing the anatomy of the CoW from angiographic images remains a manual, time-consuming task requiring specialist judgment. The CoW anatomy involves vessels that vary in diameters from around 1 to 4 mm.\supercite{krabbe1998circle} The vessels are often tortuous and difficult to identify in isolation; anatomical identification frequently depends on their spatial relationships. Furthermore, the CoW exhibits numerous natural variants, with key arteries frequently hypoplastic or absent.\supercite{krabbe1998circle,iqbal2013comprehensive} Such complexity and heterogeneity of the anatomy make characterizing CoW vasculature challenging.

The CoW vessels are commonly diagnosed by two non-invasive angiographic imaging modalities, namely magnetic resonance angiography (MRA) and computed tomography angiography (CTA). Despite recent releases of public MRA datasets,\supercite{bullitt2005vessel,IXIdataset,CAS2023,chatterjee2024smile,mou2024costa} vessel annotations were binary with no anatomical labeling of the CoW. Annotated datasets on the other important modality, CTA, did not exist. Prior work on the CoW characterization was developed mainly as a labeling task built upon binary vessel masks, skeletons or graphs,\supercite{bogunovic2013anatomical,robben2013anatomical,robben2016simultaneous,chen2020automated,hong2023automated,vos2025evaluation} and with two recent studies tackling it as a multiclass segmentation task.\supercite{dumais2022eicab,hilbert2022anatomical} However, only private CoW annotations were used, and the modality was restricted to MRA only. There were no benchmarks to compare CoW segmentation algorithms.  Furthermore, past studies have not sufficiently explored the difficulties associated with the complex and variable CoW anatomies in clinical settings, leaving their clinical relevance uncertain.

To this end, we organized a benchmark on ``Topology-Aware Anatomical Segmentation of the Circle of Willis for CTA and MRA" (TopCoW) in 2023 and 2024. TopCoW was the first public challenge on CoW anatomical segmentation featuring a paired dataset with voxel-level vessel annotations on both MRA and CTA. The main tasks benchmarked were CoW segmentation and CoW variant classification. Algorithms were crowd-sourced from global participants and evaluated on both internal and external multi-center test datasets. CoW segmentation algorithms were additionally evaluated on two clinical downstream tasks: classifying fetal-type posterior cerebral artery (PCA) and locating intracranial aneurysms using CoW segmentation. For transparency and continued development, we publicly released our annotated datasets and top-performing algorithms on Zenodo (see links in \hyperref[sec:data]{data} and \hyperref[sec:code]{code} availability sections).

%% file: sections/material_n_methods.tex
\section{Methods}
\label{sec:methods}

\subsection{Datasets}

The TopCoW challenge cohort was composed of patients admitted to the Stroke Center of the University Hospital Zurich (USZ) in 2018 and 2019 with suspicion of stroke-related neurological conditions. The inclusion criteria for the TopCoW data were: 1) both MRA and CTA scans were available and in good quality for that patient; 2) at least the MRA or CTA allowed for an assessment of the CoW anatomy; 3) no large aneurysms inside the CoW region of interest (ROI); 4) certain rare CoW variants that could not be characterized by our 13 annotated CoW components were excluded. All TopCoW data were anonymized, defaced, re-oriented, and cropped to the braincase region. More information on the TopCoW dataset can be found in Supplementary Material, \ref{sec:data_cohort_suppl}-\ref{sec:notes2023dataupdate}.

In addition to the internal test sets from the TopCoW dataset, we gathered and annotated 86 MRA and CTA scans from four external multi-center test datasets to evaluate the robustness of the algorithms. The external test datasets were sourced from existing public datasets and annotated in the same way as the TopCoW dataset. Two external CTA test sets were from the ISLES'24 challenge training set (ISLES)\supercite{de2024isles,riedel2026ischemic} and the 2021 Large IA (intracranial aneurysm) segmentation dataset.\supercite{LargeIAzenodo,LargeIAcellpattern2021} Two external MRA test sets were from the 2022 Lausanne OpenNeuro aneurysm dataset (Lausanne)\supercite{LausanneOpenNeuro,LausanneNeuroInfoPaper2023} and the IXI Hammersmith Hospital dataset (IXI-HH).\supercite{IXIdataset} The external data comprised scans from multiple manufacturers. The statistical summary of the image information for the training, internal test, and external test data is shown in Supplementary Material, \ref{sec:external_data_char}.

\subsection{Data Annotation}

\begin{figure*}[ht!]
    \centering
    \includegraphics[width=0.79\textwidth]{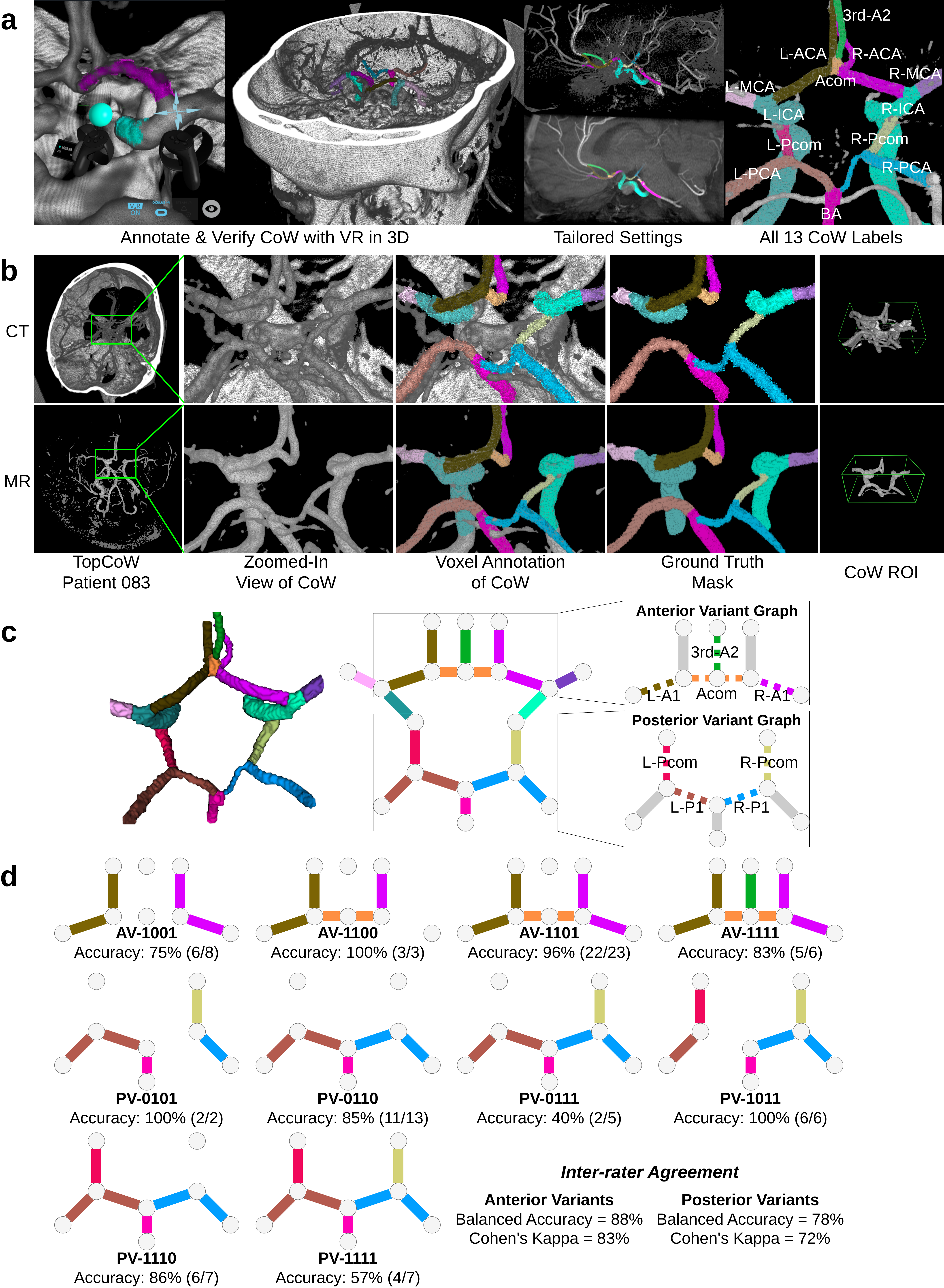}
    \caption{
    \textbf{CoW annotation overview.}
    (a) Left shows data annotation using VR. Middle shows tailored settings by adjusting opacity, threshold, and window dynamically for suitable visualization during annotation. The right image shows the 13 anatomical labels for the CoW anatomy.
    (b) TopCoW dataset provides paired images from two modalities, CTA and MRA, for each patient. Voxel annotations of the CoW vessels and a 3D bounding box of the CoW ROI are labelled for each modality.
    (c) The CoW segmentation mask is converted into a CoW variant graph annotation. Each CoW can be classified by an anterior variant (AV) and a posterior variant (PV) graph. Both AV and PV are identified by a four-edge graph, with 0 being absent and 1 being present in the edge-list.
    (d) Inter-rater agreement for CoW variant classification on 40 TopCoW CTA test cases. Accuracy for each variant is shown along with the balanced accuracy and Cohen's Kappa score.
    }
    \label{fig:New_VR_Var_rater_fig}
\end{figure*}

For each 3D angiography image, we provided three types of annotations regarding the CoW: the voxel-level multiclass segmentation mask of the CoW, a 3D bounding box for the CoW ROI, and the CoW variant graph. Virtual reality (VR) was used to efficiently annotate and verify CoW anatomy in 3D. Fig.~\ref{fig:New_VR_Var_rater_fig}a shows the workflow and view from VR. VR annotations were performed using syGlass,\supercite{pidhorskyi2018syglass} as previously described.\supercite{kaltenecker2023virtual} Fig.~\ref{fig:New_VR_Var_rater_fig}a right shows an MRA example with all 13 CoW vessel classes of multiclass segmentation annotation: left and right internal carotid artery (ICA), left and right anterior cerebral artery (ACA), left and right middle cerebral artery (MCA), anterior communicating artery (Acom), left and right posterior communicating artery (Pcom), left and right posterior cerebral artery (PCA), and basilar artery (BA). Occasionally, the anterior part of the CoW can have a third A2 artery arising from the Acom, and we labeled it with class 3rd-A2. Fig.~\ref{fig:New_VR_Var_rater_fig}b shows an example segmentation mask and ROI annotations for both MRA and CTA from a TopCoW patient. The CoW annotation protocol was developed by a senior neurosurgeon (Y.M., over 10 years of experience) and reviewed by another senior neurosurgeon (P.B., over 15 years) and a senior neurologist (S.W., over 15 years). Two doctoral researchers (K.Y. and F.M.) performed voxel-level annotations after training by Y.M. The annotators each labeled half of the dataset partitioned by patient. Uncertain cases were adjudicated by Y.M., and a subset was resolved through expert consensus involving additional neurosurgeons and neurologists.  More information on the annotation protocol is in Supplementary Material, \ref{sec:anno_protocol}.

The CoW variant graph for classification was derived from the segmentation mask and encompasses the anterior variant (AV) and posterior variant (PV) graphs. These variants were defined based on the presence of L-A1, Acom, 3rd-A2, and R-A1 vessels for AV, and L-Pcom, L-P1, R-P1, and R-Pcom vessels for PV. Hypoplastic vessels were categorized as present. Fig.~\ref{fig:New_VR_Var_rater_fig}c shows the AV and PV graph composition. Distributions of the CoW variants for AV and PV in all our datasets are shown in Supplementary Material, \ref{sec:cow_var_distribution}.

To estimate the variability of the annotations used for benchmarking and the upper bounds of algorithmic performance, we analyzed inter-rater agreement for CoW variant classification and voxel-level annotation. For CoW variant classification, senior neurosurgeon P.B. independently re-labeled the AV and PV classes for 40 CTA patients from the TopCoW test set. These cases were selected using stratified sampling across all available CoW variants in the internal test data, prioritizing rare anatomical variants. As shown in Fig.~\ref{fig:New_VR_Var_rater_fig}d, the selected sample included four AV and six PV classes. Labels from P.B. were compared with the annotations used in the benchmark. Balanced accuracies between the raters were 88\% for AV and 78\% for PV. Cohen's Kappa scores were 83\% for AV and 72\% for PV, suggesting good agreement.  Detailed results on the voxel-level segmentation agreement can be found in Supplementary Material, \ref{sec:inter_rater_voxel_seg}.

\subsection{Challenge Structure}

The featured task of TopCoW across both iterations was multiclass segmentation. The first 2023 challenge had a task of merged-binary segmentation, which was deemed sufficiently solved and replaced by CoW ROI detection and CoW variant graph classification tasks in the 2024 iteration. The dataset expanded considerably in 2024 (125 training/70 internal test patients). Teams were ranked by average metric performance on internal data, with top performers validated on external datasets. Our challenge had two modality tracks, MRA and CTA, for algorithm evaluations. For the internal test set, algorithm input was a pair of MRA and CTA, and teams could choose one or both modalities. The input during the external test phase was single-modality which was compatible with all top submissions. (See Supplementary Material, Fig.~\ref{fig:nejm_flowchart} and Sections \ref{sec:algo_submit}–\ref{sec:ranking_robustness} for details.)

\subsection{Evaluation of Segmentation Algorithms}

The submitted algorithms were in the form of isolated Docker containers and were evaluated within the CoW ROI using the following metrics:

\paragraph{\textbf{Voxel-Level Multiclass Segmentation}}
For voxel-level metrics, the multiclass CoW segmentation predictions were evaluated using Dice similarity coefficient (Dice score),\supercite{dice1945measures} centerline Dice (clDice),\supercite{shit2021cldice} Hausdorff distance at 95th percentile,\supercite{taha2015metrics,maier2024metrics} and connected component error.\supercite{hu2019topology,menten2023skeletonization}

\paragraph{\textbf{Beyond Segmentation I: Key CoW Component Detection}}
The first beyond segmentation metric was the average F1 score for detection of Acom, Pcoms, and 3rd-A2. Positive detection was defined as at least 25\% intersection over union between the predicted and ground truth masks. This relatively lenient cutoff is consistent with prior studies on small-object detection in medical imaging with a similar range (10-30\%).\supercite{baumgartner2022accurate,ceballos2024vessel}

\paragraph{\textbf{Beyond Segmentation II: CoW Variant Classification}}
The second beyond segmentation metric was variant balanced accuracy (VarBalAcc) for CoW variant graph classification. The VarBalAcc was calculated for both anterior and posterior variants. The variant class was determined using the AV and PV edge-list of the variant graph, as shown in Fig.~\ref{fig:New_VR_Var_rater_fig}c. The segmentation mask was converted to edge-list based on presence of the Acom, Pcoms, and 3rd-A2 labels, and whether the ACA and PCA were connected to their respective neighboring labels - ICA and BA - for the A1 and P1 edges.

\paragraph{\textbf{Clinical Application I: Fetal PCA Classification}}
We extracted centerlines and diameters of CoW segmentation masks using a prior workflow\supercite{musio2024CoWGraph,musio2026circle} with Supplementary \ref{sec:details_on_fetal} to provide more details. The diameters along the Pcom and P1 segments were used to determine the CoW anatomical variant called the fetal PCA variant.\supercite{osborn2013osborn} We compared the diameters of the Pcom and P1 at the 25th percentile. The CoW was classified as having a fetal PCA variant if the Pcom was slightly larger in diameter ($>\!1.05\times$ the diameter of P1). Fetal PCA was assessed separately for the left and right sides. The same set of 40 TopCoW CTA cases used in inter-rater agreement for variant classification were labeled for fetal PCA class by the senior neurosurgeon (P.B). The fetal PCA labels from P.B. were treated as ground truth.

\paragraph{\textbf{Clinical Application II: Locating Aneurysm}}
We selected 12 patients with intracranial aneurysms along their CoW vessels from the aforementioned external LargeIA dataset, which included aneurysm ground truth annotations. The aneurysm locations were then labeled by a senior neurosurgeon (Y.M.). CoW segmentation algorithms were applied to the images of the aneurysm patients, and the resulting CoW predictions were overlaid with the provided aneurysm ground truth. The predicted location of the aneurysm was determined by identifying the CoW labels adjacent to or overlapping with the aneurysm mask.

%% file: sections/results.tex
\section{Results}
\label{sec:results_sec}

\subsection{The TopCoW Dataset}

In total, 200 paired MRA and CTA scans from unique patients were curated and randomly split by patient into 125 training, 5 validation, and 70 test subjects. The cohort was 40\% female with a median age of 74 years (interquartile range 61 to 82; Table~\ref{table:clinical_char}). The primary imaging indication was ischemic stroke (61\%), followed by stroke or transient ischemic attack (TIA) mimic (15.5\%) and TIA (14.0\%).

\input{tables/table_clinical_characteristics.tex}

\subsection{Winning Algorithm Design}

\begin{figure*}[ht!]
    \centering
    \includegraphics[width=0.75\textwidth]{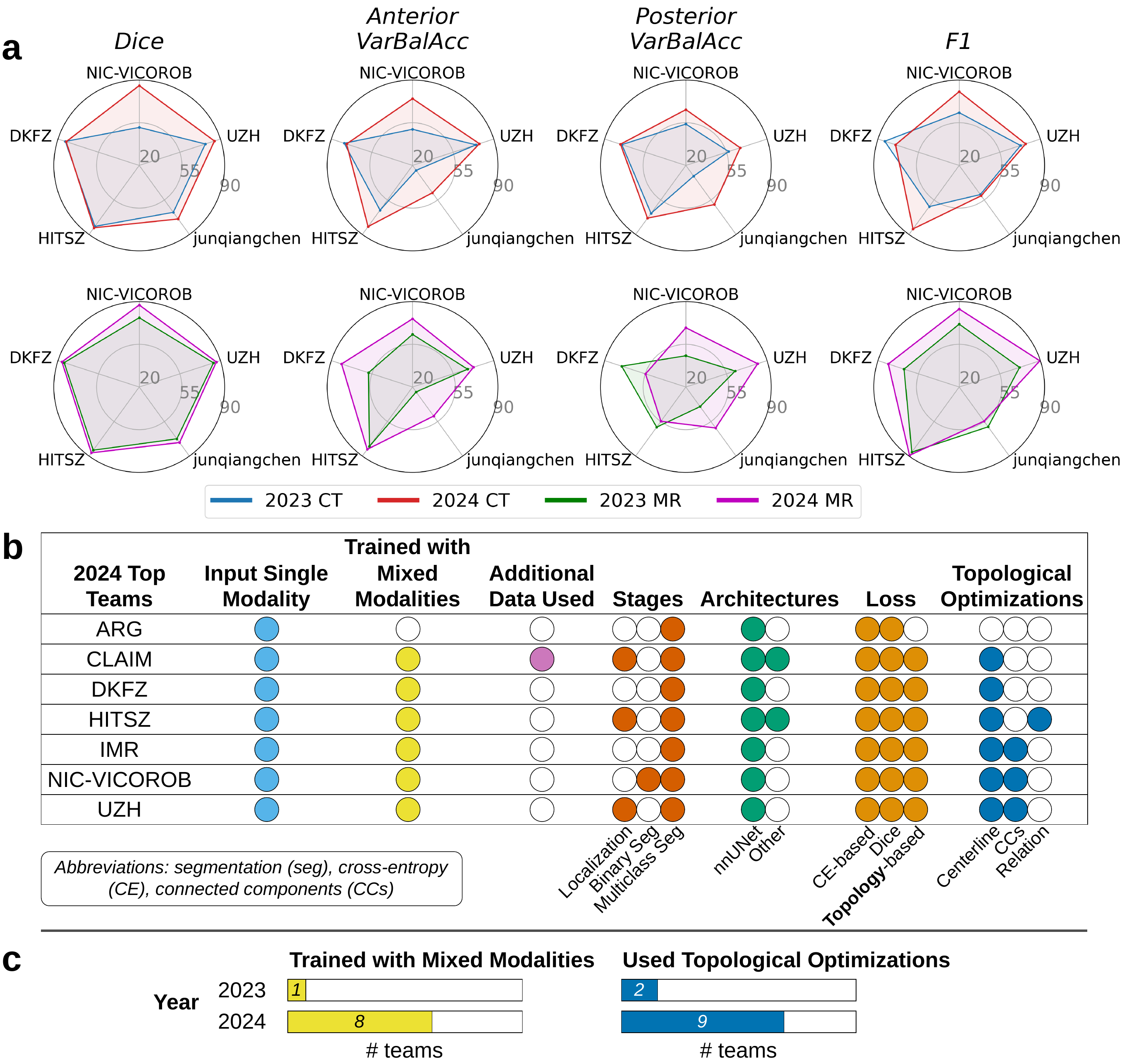}
    \caption{
    \textbf{Progression of TopCoW submissions and winning strategies.}
    (a) Performance of the five teams that participated in both years on the same 34 patients in both years' test sets. The metrics shown are class-average Dice, variant-balanced accuracy (VarBalAcc) of the anterior and posterior CoW variant classifications, and average F1 score for detection of Acom, Pcoms, and 3rd-A2. The upper row shows CTA performance; the lower shows MRA.
    (b) Key characteristics of the segmentation algorithms from the top six teams from both tracks in alphabetical team name order.
    (c) Two methodological breakthroughs for segmentation in 2023 got picked up by many more teams in the following year.
    }
    \label{fig:Common34_KeyTable}
\end{figure*}

Twenty-five teams out of over 250 registrants from six continents submitted to the challenge. As shown in Fig.~\ref{fig:Common34_KeyTable}a, the performance improved for the five teams participating in both years, particularly for the CTA modality. Fig.~\ref{fig:Common34_KeyTable}b summarizes key design choices of the segmentation algorithms from the top teams in 2024. Most teams utilized a unified architecture with shared parameters to process single-modality inputs from a mixed-modality training pool, making their models modality-agnostic. Only one team used additional, independently prepared training data, which included some of the external MRA test images but without our ground truth labels. Over half adopted a two-stage pipeline, such as localization followed by segmentation, which allowed the model to localize segmentation training to a sub-volume, reduce false positives in the background, and conserve computational resources for ensembling. All top teams based their architectures on nnUNet.\supercite{isensee2021nnUNet} Two methodological shifts emerged from 2023 to 2024: broader adoption of mixed-modality training and topological optimization as shown in Fig.~\ref{fig:Common34_KeyTable}c.

\subsection{Voxel-Level Multiclass Segmentation}

\begin{figure*}[p]
    \centering
    \includegraphics[width=0.98\textwidth]{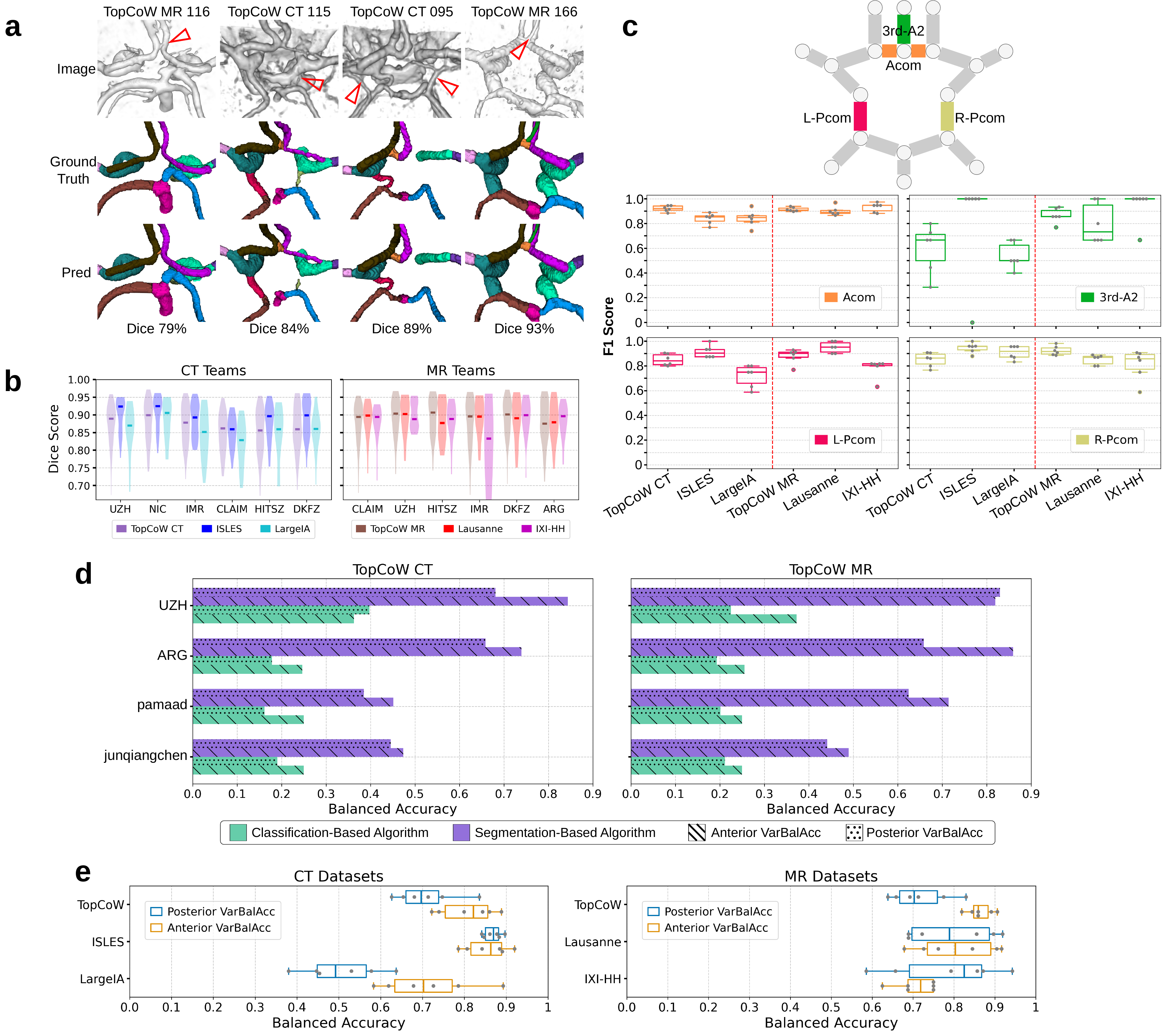}
    \caption{
    \textbf{Segmentation performance.}
    (a) Qualitative results for the multiclass segmentation task. Ground truth masks are shown alongside predictions from team `UZH' for four selected cases from the internal test set, ordered by increasing class-average Dice score. Red arrowheads in the image indicate challenging features for each case (MR-116: indeterminate Acom vessel; CT-115: hypoplastic R-Pcom; CT-095: venous contamination; MR-166: 3rd-A2 variant).
    (b) Class-average Dice scores from the top six teams on CTA and MRA internal and external test sets. Team `NIC-VICOROB' is abbreviated as `NIC'. Team `CLAIM' used additional training data which included some external MRA test images from Lausanne and IXI-HH without our ground truth labels. Team `IMR' had five cases in IXI-HH with Dice below the shown range. The charts show violin plots with the middle bar indicating the median.
    (c) Detection of key CoW components for all test datasets by the top six teams from each modality. F1 scores from the six teams were displayed in box plots.
    (d) Four teams submitted both classification-based and segmentation-based algorithms for classifying the CoW variants. Classification-based algorithms were trained to predict CoW variant classes directly. Segmentation-based algorithms were purely for CoW segmentation and later evaluated for CoW variant classification performance.
    (e) CoW variant classification performance from the top six segmentation teams on internal and external test sets. The box plots show the anterior and posterior VarBalAcc scores from the top six teams.
    }
    \label{fig:mergedSegResHori}
\end{figure*}

Fig.~\ref{fig:mergedSegResHori}a shows qualitative segmentation results on the TopCoW internal test sets from a top team. Two patients per modality were selected, representing a wide range of CoW variants and class-average Dice scores. The predictions accurately segmented various complex CoW anatomies, capturing vessel curvatures and class boundaries.

Fig.~\ref{fig:mergedSegResHori}b shows class-average Dice scores from the top six teams on internal and external test sets. All top teams showed good generalization across the external test sets, with median Dice scores above 80\% across modalities and test sets. The IXI-HH performance drop was unique to team `IMR' due to their 0–700 intensity truncation for MRA training data preprocessing, which reduced generalization to a subset of IXI-HH images with high mean intensities. Other voxel-level metrics showed similarly strong performance and generalization trends in Supplementary \ref{sec:detailed_2024_INternal_results}-\ref{sec:detailed_2024_EXternal_results}.

\subsection{Beyond Segmentation I: Key CoW Component Detection}

The presence or absence of four key CoW components — the communicating arteries (Acom, R-Pcom, L-Pcom) and the 3rd-A2 segment — directly determines many CoW variant types. Accurate detection of these four components is a clinically relevant feature of segmentation algorithms. We assessed their detection performance across all test sets in Fig.~\ref{fig:mergedSegResHori}c, showing F1 scores from the top six teams of each modality in boxplots. Detection of the communicating arteries was consistent across test sets and modalities, with top teams achieving over 75\% F1 scores on most sets.

\subsection{Beyond Segmentation II: CoW Variant Classification}

Notably, four teams took part in both the segmentation task and the classification task with segmentation-based and classification-based algorithms respectively. As shown in Fig.~\ref{fig:mergedSegResHori}d, segmentation-based methods outperformed classification-based ones by at least a factor of two on the internal test sets.

Fig.~\ref{fig:mergedSegResHori}e shows the VarBalAcc from the top six segmentation teams for anterior and posterior variants across all test sets. The top teams were able to generalize well to external MRA datasets with both anterior and posterior VarBalAcc averaging around 80\%, and to external CTA datasets with above 70\% VarBalAcc in general, which are near human-level performance.

\subsection{Clinical Application I: Fetal PCA Classification}

\begin{figure*}[ht!]
    \centering
    \includegraphics[width=0.72\textwidth]{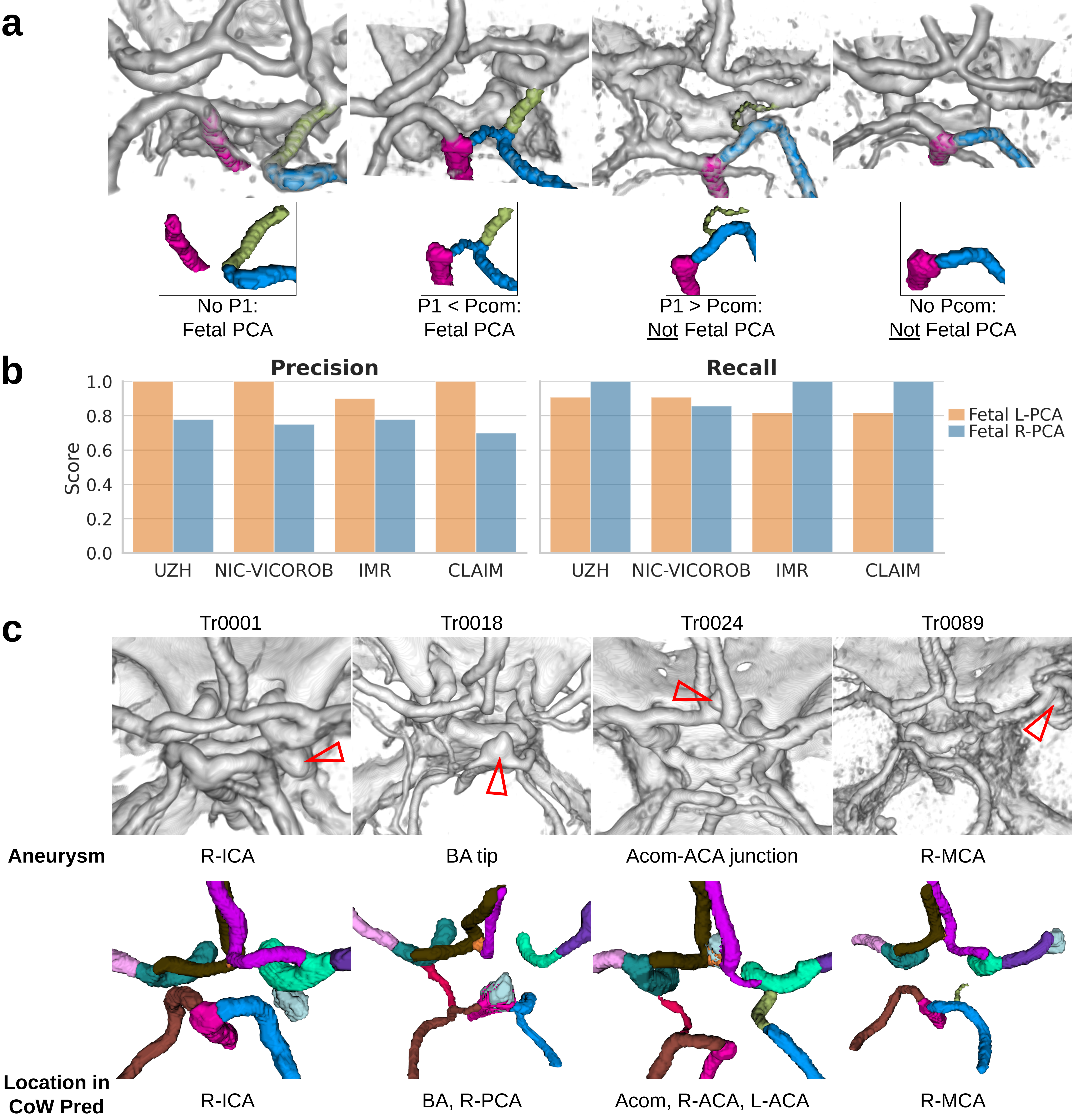}
    \caption{
    \textbf{Two clinical applications of CoW segmentation.}
    (a) Segmentation masks of the ipsilateral Pcom and P1 segments were used to determine whether a fetal PCA class was present on that side. Upper row shows CTA images overlaid with segmentation masks related to fetal PCA. Lower row shows zoomed-in view of the segmentation masks. Examples shown illustrate the fetal R-PCA classification.
    (b) Fetal PCA classification performance in terms of precision and recall scores from the top four teams.
    (c) Upper row shows CTA images with aneurysms indicated by red arrowheads. Lower row shows the aneurysm masks in silver color overlaid with predicted CoW segmentation masks from team `UZH'. The ``Location in CoW Pred" indicates the predicted CoW labels that the aneurysm overlapped with or was adjacent to.
    }
    \label{fig:fetal_aneu}
\end{figure*}

One potential clinical use of CoW segmentation models is fetal PCA classification, which is important for surgical planning and interpretation of perfusion imaging;\supercite{osborn2013osborn} neurosurgeons and neuroradiologists routinely assess fetal PCA status from CTA and MRA images. As shown in Fig.~\ref{fig:fetal_aneu}a, diameters along the ipsilateral Pcom and P1 masks were used to predict the fetal PCA class. This allowed us to convert the top teams' segmentation outputs into fetal PCA labels. Fig.~\ref{fig:fetal_aneu}b shows that these outputs achieved around 80\% or higher precision and recall for both fetal L-PCA and fetal R-PCA classification.

\subsection{Clinical Application II: Locating Aneurysm}

Another important potential clinical use is locating aneurysms using CoW segmentation as reference. Clinicians routinely need to locate intracranial aneurysms relative to CoW anatomy in MRA and CTA images, as aneurysms most often occur along CoW vessels.\supercite{osborn2013osborn} 12 patients with aneurysms at various locations were segmented by the top four teams. Fig.~\ref{fig:fetal_aneu}c shows the aneurysm ground truth overlaid with the CoW segmentations from a top team for four representative cases. CoW vessel labels overlapping or adjacent to the aneurysm were used to detect its location. Team `UZH' correctly located the aneurysms in all 12 patients, teams `NIC-VICOROB' and `CLAIM' in 11 out of 12, and team `IMR' in 10 out of 12 patients.

%% file: tables/table_clinical_characteristics.tex
\begin{table}[!htbp]
\caption{Clinical characteristics of the TopCoW cohort, including age, sex, and type of event. Age statistics exclude one case with an implausible age of 0 years. Sex statistics exclude one case with no recorded sex information.
Abbreviations: standard deviations (SD), inter-quartile range (IQR), minimum (min), and maximum (max).}
\label{table:clinical_char}
\centering
\begin{tabular}{lc}
\hline
\textbf{Variable} & \textbf{Values (n=200)} \\
\hline
\multicolumn{2}{l}{\textbf{Age}} \\
Mean (SD) & 71.27 (14.08) \\
Median [IQR] & 74.00 [61.05--82.35] \\
Min--Max & 24.20--91.80 \\
\hline
\multicolumn{2}{l}{\textbf{Sex}} \\
Male & 119 (59.5\%) \\
Female & 80 (40.0\%) \\
\hline
\multicolumn{2}{l}{\textbf{Type of event}} \\
Ischemic stroke & 122 (61.0\%) \\
Stroke or TIA mimic & 31 (15.5\%) \\
Transient ischemic attack & 28 (14.0\%) \\
Intracerebral hemorrhage & 9 (4.5\%) \\
Retinal infarct & 6 (3.0\%) \\
Cerebral sinus vein thrombosis & 2 (1.0\%) \\
Amaurosis fugax & 2 (1.0\%) \\
\hline
\end{tabular}
\end{table}

%% file: sections/discussions.tex
\section{Discussion}
\label{sec:discussions_sec}

\subsection{Benefits of Mixed Modality Training for CTA}

Training a unified model with mixed modalities was an effective strategy, especially for the more difficult CTA modality. Between MRA and CTA, CTA consistently showed lower metric scores. This may be due to veins and bones near the CoW seen in CTA, and less detailed soft tissue in the background compared to MRA. CTA benefited significantly more from mixed modality training, showing greater performance gains than MRA. This is a key finding that may inform other CTA segmentation tasks.

\subsection{Importance of Topological Optimizations}

Topological optimizations enabled CoW segmentation algorithms to be used in topology-dependent downstream clinical tasks, such as CoW variant and fetal PCA classification. These applications require segmented vessels to preserve key topological properties. Top submissions applied a range of topological optimizations, such as topology-preserving loss functions,\supercite{kirchhoff2024skeleton,shi2024cbDice,zhang2023towards} connected component post-processing,\supercite{imr2024topology,hamadache2026topology} and class neighborhood relations.\supercite{gupta2022learning,shi2023nextou} Collectively, these methods improved connectivity, centerline accuracy, and overall topological correctness, making the segmentations suitable for topology-sensitive downstream tasks.

\subsection{Comparison with CROWN Results}

Concurrent to the TopCoW challenge in 2023, the CROWN challenge\supercite{vos2025evaluation} also benchmarked the CoW variant classification task. The key difference was that CROWN did not provide any CoW segmentation annotations and instead formulated the task purely as image classification. We compared the performance of our algorithms with those reported in CROWN. Based on the merged common set of variants, the top two CROWN teams achieved 24–30\% anterior and 28–50\% posterior balanced accuracy, whereas the top two TopCoW MRA teams reached 82–89\% anterior and 75–82\% posterior balanced accuracy. Similar findings were observed in our analysis (Fig.~\ref{fig:mergedSegResHori}d), where segmentation-based algorithms — trained and optimized for the CoW multiclass segmentation task — outperformed classification-based methods on the variant classification task by at least two-fold.

\subsection{Explainability via Segmentation}

CoW segmentation algorithms also provide explainability - a key feature in clinical settings. We showed that top-performing models could effectively detect key CoW components, classify CoW variants and fetal-type PCA, and locate intracranial aneurysms. Importantly, the conversion of segmentation masks into downstream outputs was fully transparent and interpretable, unlike black-box approaches. When clinicians require a ``confidence" score or rationale behind a prediction — whether for detection, classification, or localization — they can better interpret and contextualize the results by inspecting the CoW segmentation output.

\subsection{VR to Handle Complex Anatomy}

As one of the first challenges to use VR generated annotations at-scale, our VR-based expert-in-the-loop annotation protocol proved very effective and can overcome the otherwise time-consuming annotation process for a complex multiclass anatomical segmentation problem. The 3D depth view offered efficient annotation/verification capabilities well suited for curvilinear structures like the CoW vessels, enabling fast and accurate annotation of complex CoW anatomies and rare variants. This allowed us to prepare such a densely annotated large-scale dataset that covered many clinically relevant CoW anatomies. This VR-driven efficiency mirrors recent findings where VR increased the speed and accuracy of 3D annotations.\supercite{kaltenecker2023virtual}

\subsection{Clinical Implications, Limitations, and Future Work}

Accurate CoW variant characterization is relevant for neurovascular practice because variant patterns may influence collateral circulation assessment and procedural planning. In this benchmark, the segmentation-based approach provided voxel-level vessel delineation and transparent derivation of key components, variant classes, fetal PCA classification, and aneurysm localization. These outputs can inform endovascular catheter navigation and support evaluation of flow redistribution during microsurgical interventions, including bypass strategies and temporary vessel occlusion. The benchmarked algorithms may therefore help automate CoW analysis and enable more standardized pre-procedural planning, although this benchmark does not directly evaluate effects on procedural outcomes or safety. Nevertheless, several limitations of our study should be acknowledged. Due to the large heterogeneity of CoW anatomy, not all CoW variants were included in our annotation scheme. Expanding labels to include rarer variants could improve model robustness and applicability. The current CoW variant graph could also be refined into more detailed types involving hypoplasia and vessel diameters using the same workflow applied in our fetal PCA classification, where diameters were extracted along the centerlines of P1 and Pcom. Our dataset exhibited class imbalance, with certain vessel variants represented by only 2-3 training examples, which may reduce prediction reliability for uncommon but clinically important variants. External validation was restricted to the top six teams per modality due to resource constraints; however, as this represents approximately half of all submissions in the 2024 iteration, the subset provides a representative assessment of the state of the art. Finally, while this work solely focuses on the CoW, future efforts could focus on segmenting the full brain vessel anatomy, including all major arteries and veins visible in angiograms, enabling a more comprehensive vascular analysis.

%% file: sections/acknowledgement.tex
The challenge is supported by the Digitalization Initiative of the Zurich Higher Education Institutions (DIZH) and the Helmut Horten Foundation. We thank Hrvoje Bogunović for helpful discussions and suggestions during the early planning stages. We also thank Nathan Spencer and Michael Morehead from syGlass for the technical assistance for the VR setup, and James Meakin and Chris van Run from grand-challenge.org for the technical support for the challenge infrastructure.

Ynte Ruigrok has received funding from the European Research Council (ERC) under the European Union's Horizon 2020 research and innovation program (grant agreement No. 852173). Hakim Baazaoui received funding from the Koetser Foundation and the ``Young Talents in Clinical Research" program of the SAMS and of the G. \& J. Bangerter-Rhyner Foundation. Paula Casademunt, Oscar Camara, Philippe Bijlenga, Julien Haemmerli, Sven Hirsch, Norman Juchler, Fabio Musio, and Matteo Delucchi received funding from GEMINI (funded by EU Horizon Europe R\&I program No 101136438). Ting Ma received funding from the National Natural Science Foundation of P.R. China (62276081).

Team \textbf{2i\_mtl} was supported by Grants from the Quebec Bio-Imaging Network (Project No. 21.24) and start-up funds from the Centre de Recherche du CHUM and Departement de radiologie, radio-oncologie et medecine nucleaire, Universite de Montreal/Bayer. Laurent Letourneau-Guillon is supported by a Clinical Research Scholarship-Junior 1 Salary Award (311203) from the Fonds de Recherche du Quebec en Sante and Fondation de l’Association des Radiologistes du Quebec.
Team \textbf{ARG} was supported by the Institute of Information \& Communications Technology Planning \& Evaluation (IITP) grant funded by the Korea government (MSIT) No. RS-2022-II220959 ((Part 2) Few-Shot Learning of Causal Inference in Vision and Language for Decision Making).
Team \textbf{CLAIM} acknowledges funding from the German Federal Ministry of Education and Research (ANONYMED Project, no. 16KISA042K) and from the European Unions Horizon 2020 research and innovation programme (VALIDATE Project, no. 777107, coordinator Dietmar Frey). Computation has been performed on the HPC for Research cluster of the Berlin Institute of Health. They also acknowledge the contribution of MRCLEAN investigators by providing access to data from the MRCLEAN trial.
Team \textbf{DKFZ} was supported by the Helmholtz Association under the joint research school ``HIDSS4Health - Helmholtz Information and Data Science School for Health" and part of their work was funded by Helmholtz Imaging (HI), a platform of the Helmholtz Incubator on Information and Data Science. Julius C. Holzschuh has been funded by a fellowship of the DKFZ Clinician Scientist Program, supported by the Dieter Morszeck Foundation.
Team \textbf{EURECOM} was partially funded by the French government, through the 3IA Cote d’Azur Investments in the Future project managed by the ANR (ANR-19-P3IA-0002) and by the ANR JCJC project I-VESSEG (22-CE45-0015-01).
Team \textbf{IMR} was supported in part by National Key R\&D Program of China (Grant Number: 2022ZD0212400), Natural Science Foundation of China (Grant Number: 62373243) and the Science and Technology Commission of Shanghai Municipality, China (Grant Number: 20DZ2220400), Shanghai Municipal Science and Technology Major Project (No.2021SHZDZX0102).
Team \textbf{IWantToGoToCanada} was supported by the National Research Foundation (NRF-2020M3E5D2A01084892), Institute for Basic Science (IBS-R015-D1), ITRC support program (IITP-2023-2018-0-01798), AI Graduate School Support Program (2019-0-00421), ICT Creative Consilience program (IITP-2023-2020-0-01821), and the Artificial Intelligence Innovation Hub program (2021-0-02068).
Team \textbf{lWM} wants to acknowledge the Polish HPC infrastructure PLGrid support (No. PLG/2023/016239).
Team \textbf{NantesU} was partially supported by the French ANR project ``eCAN" and INSERM CoPoC \#MAT-PI-22155-A-01 (RVF23037NSA).
Team \textbf{NIC-VICOROB} was supported by the Ministerio de Ciencia e Innovacion (DPI2020-114769RB-I00) as well as by ICREA under the ICREA Academia programme, and also partly supported the Ministerio de Ciencia e Innovacion (DPI2020-114769RB-I00). Members of the 2024 team received the support from the PID2020-114769RBI00 and the PID2023-146187OB-I00 projects funded by the Ministerio de Ciencia, Innovación y Universidades.
Team \textbf{Pamaad} P. Casademunt is supported by the European Union's Horizon 2020 grant agreement No.101136438 (GEMINI project), by the Agència de Gestió d’Ajuts Universitaris i de Recerca (grant No. 2024 FI-1 00419), and the Maria de Maeztu grant of excellence. A. Galdran is supported by grant RYC2022-037144-I, funded by MCIN/AEI/10.13039/501100011033 and by FSE+. They would like to thank the HPC team from ZHAW, particularly Pascal Häussler and Stefan Weber, for their generous allocation of computational resources and technical assistance.
Team \textbf{UB-VTL} wants to acknowledge the computational resources provided by the Center of Computational Research (CCR) at University of Buffalo.
Team \textbf{UW} was supported by the United States National Institute of Health (grants R01HL162743 and R00HL136883).

Dr. Catherine Wurster contributed to this work during its early stages and passed away unexpectedly before publication.

%% file: supplement.tex

\newpage
\clearpage
\pagenumbering{roman}  
\setcounter{page}{1}    
\renewcommand{\thesection}{S\arabic{section}} 
\setcounter{section}{0} 
\renewcommand{\thefigure}{S\arabic{figure}} 
\setcounter{figure}{0}  
\renewcommand{\thetable}{S\arabic{table}}   
\setcounter{table}{0}   


\startcontents[supplement]

\onecolumn
\begin{center}
    \mainheading{Supplementary Material}\\
    \vspace{0.5em}
    For the paper \emph{``The TopCoW Challenge --- Topology-Aware Circle of Willis Segmentation for CT and MR Angiography"}\\
    \vspace{0.7em}
\end{center}
\printcontents[supplement]{}{1}{}
\twocolumn

\clearpage

\begin{figure*}[ht!]
    \centering
    \includegraphics[width=0.73\textwidth]{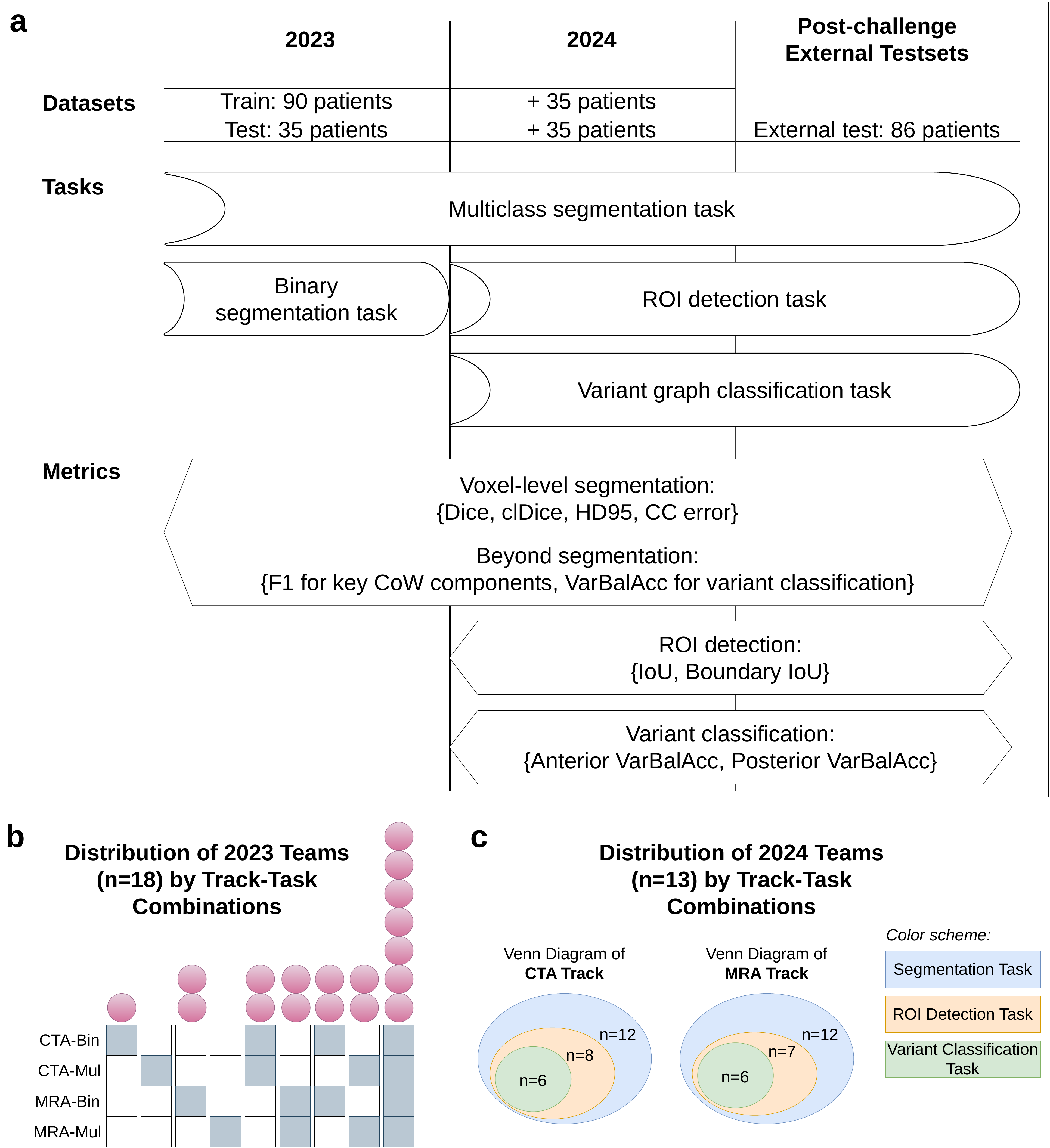}
    \caption{
    Structure of the 2023 and 2024 challenges and associated tasks.
    \textbf{(a)} Schematic flowchart showing the overview of the datasets, tasks, and evaluation metrics for both years.
    There were two tracks, MRA and CTA, for each task.
    Abbreviations: connected component (CC), centerline Dice (clDice), Hausdorff distance at 95th percentile (HD95), intersection over union (IoU), region of interest (ROI), variant balanced accuracy (VarBalAcc).
    Notes: The 2023 training data additionally released 20 MRA scans from another challenge; 2023 originally used a subset of the segmentation metrics; an additional beyond segmentation metric, topology match rate (TMR), was used for analysis and discussed in the supplementary.
    \textbf{(b)} Histogram of 2023 teams and track-task combinations. The tracks can be CTA or MRA. The tasks can be binary (Bin) or multiclass (Mul) segmentation. Each histogram bin is a vertical column of four possible participation (grey is true, white is false). The circles represent the number of teams for that bin. 7/18 teams took part in all four track-task combinations in 2023.
    \textbf{(c)} The Venn diagram illustrates a perfect nested hierarchy in 2024 task participation: all teams performing CoW variant classification task also competed in CoW ROI detection task, and all detection teams participated in the core segmentation task. This distribution emerged spontaneously, as there were no mandatory requirements to participate in multiple tasks.
    }
    \label{fig:nejm_flowchart}
\end{figure*}

The supplementary material begins with a schematic Fig.~\ref{fig:nejm_flowchart} outlining the datasets, tasks, evaluation metrics, and analyses for both years. Subsequent sections are organized by thematic headers for ease of navigation.

\input{sections/suppl/data.tex}
\input{sections/suppl/annotation_protocol.tex}
\input{sections/suppl/interrater}
\input{sections/suppl/variant_distribution.tex}
\input{sections/suppl/eval_metrics}
\input{sections/suppl/ranking}
\input{sections/suppl/fetal}
\input{sections/suppl/description_all_methods.tex}

\input{sections/suppl/results24.tex}

\input{sections/suppl/infertime}
\input{sections/suppl/aneu}
\input{sections/suppl/results23.tex}

%% file: sections/suppl/data.tex
\section{TopCoW Imaging Characteristics}
\label{sec:data_cohort_suppl}


Siemens scanners were typically (MR 95\%, CT 99.5\%) used for both modalities.
CT image acquisition was performed on Siemens (SOMATOM) and GE (Revolution CT) devices.
MR images were acquired using Siemens (Skyra, Avanto Fit, Aera) and Philips (Ingenia, Achieva dStream) devices.
MRA scans were imaged using Time-of-Flight (TOF) with magnetic field strength of typically 3 Tesla (T) (98\%).
Detailed image acquisition parameters are provided in Table~\ref{table:img_acquisition}.

\input{tables/table_image_acquisition.tex}

\section{Inclusion and Exclusion of CoW Variants}
\label{sec:suppl_inclusion_exclusion_cow}

We tried to include as many diverse CoW variants as possible in our challenge dataset.
From our observation, the following variants were \textbf{included} and annotated in our training and test dataset:
\begin{itemize}
\itemsep0em
    \item[\cmark] with or without Acom (\textit{Note: Acom determines A1/A2})
    \item[\cmark] double Acom
    \item[\cmark] with or without Pcom (\textit{Note: Pcom determines P1/P2})
    \item[\cmark] the triple ACA variant or 3rd-A2
    \item[\cmark] aplastic or hypoplastic A1 or P1 segments
    \item[\cmark] fetal PCA variants
    \item[\cmark] when CoW vessels (e.g. ACA, PCA, Acom) have fenestrations
\end{itemize}

However, there are some rare variants of which the topology cannot be characterized by our CoW multiclass labels.
These variants are much less common than the ones we have included,
and our thirteen CoW segment labels are insufficient to describe the complex anatomy of these rare variants.
Here is a list of such CoW variants that we had observed and \textbf{excluded} from our dataset:
\begin{itemize}
\itemsep0em
    \item[\xmark] azygos ACA or when the left and right ACAs are fused
    \item[\xmark] anterior choroidal artery (AChA) course and supply replacing an ipsilateral fetal PCA
    \item[\xmark] duplicated PCA
    \item[\xmark] persistent primitive trigeminal artery between ICA and BA
\end{itemize}

\section{Anonymization, Defacing, and Pre-processing}
\label{sec:defacingCrop}
The TopCoW data used had been approved by the local ethical committee. The data were anonymized (removal and anonymization of relevant DICOM patient information).
Additional de-facing and cropping procedures were performed to ensure patient privacy in the image data after converting the DICOM to NIfTI format using dcm2niix. \supercite{rordendcm2niix}
Specifically, we masked out the face using TotalSegmentator \supercite{wasserthal2023totalsegmentator} for CTA
and shear-cutted the facial regions with quickshear method \supercite{quickshearbib} for MRA,
and then cropped the image data using brain mask from TotalSegmentator for CTA or HD-BET \supercite{isensee2019hdbet} for MRA
to include only the braincase region.

Other than the defacing and cropping-to-braincase of the nifti image data, the only pre-processing of the data was to re-orient the image to LPS+ orientation.
No further pre-processing of the data was performed to keep the data as close to the original clinical setting as possible. 

The anonymized image data were approved to be released under the “Open use. Must provide the source. Use for commercial purposes requires permission of the data owner." license from the OpenData Swiss Terms of use \url{https://opendata.swiss/en/terms-of-use}.

\section{Notes on the Updates to the 2023 TopCoW Data}
\label{sec:notes2023dataupdate}
Between the two years of the challenge, the training dataset grew from 90 to 125 patients, and the test dataset grew from 35 to 70 patients, while the validation set size remained unchanged at five images.

In 2023, due to technical reasons related to image data type, a few test images frequently
caused inference runtime error in participant Docker containers. In order for the participants to be aware of such technical issues, we moved one representative patient to the validation set in 2024 so participants could pick up on related technical mistakes early on
before submitting for the final test set.

In 2023, in addition to the TopCoW challenge training data, we also released a small specific subset (20 MRA scans) of the CROWN challenge \supercite{vos2025evaluation} data with our annotations (multiclass CoW voxel mask and CoW ROI).
This data was not included in the 2024 data release, but it is available upon request.

In 2024, we also refined the segmentation mask and bounding box labels for some of the 2023 data, both for training and the test data, and re-released the updated 2023 data in the 2024 release.
We trimmed floating blobs of disconnected components smaller than 13 voxels for each CoW mask labels, which leads to more accurate ground-truth Betti-0 numbers for segmentation masks.
We slightly adjusted some CoW ROI 3D bounding box labels for some 2023 cases, making them more harmonized and consistent with our bounding box definition.

\section{External Multi-Center Test Data \& Data Characteristics}
\label{sec:external_data_char}
To evaluate the generalizability of the algorithms, we additionally gathered and annotated 86 MRA and CTA scans from four external multi-center test datasets.
These external test datasets were from existing public datasets without CoW annotations.
Two external CTA test sets were from the TUM University Hospital in Germany of the public ISLES'24 challenge training set (ISLES) \supercite{de2024isles,riedel2026ischemic} and various hospitals in China of the public Large IA Segmentation dataset (LargeIA) \supercite{LargeIAzenodo, LargeIAcellpattern2021}.
Two external MRA test sets were from the Lausanne University Hospital in Switzerland of a public OpenNeuro dataset (Lausanne) \supercite{LausanneOpenNeuro, LausanneNeuroInfoPaper2023}
and the Hammersmith Hospital in UK of the public IXI dataset (IXI-HH) \supercite{IXIdataset}.
The inclusion criteria were similar to that of the TopCoW dataset.
For ISLES, we chose 26 CTA patients whose CoW were not occluded within the ROI.
For LargeIA dataset, we chose 20 CTAs that do not have aneurysms inside the CoW ROI.
For Lausanne and IXI-HH datasets, we chose 20 MRAs each from the healthy control group.
The external datasets comprised scans from the following manufacturers: ISLES (Siemens and Philips), LargeIA (Siemens, Neusoft, and GE Healthcare), Lausanne (17/20 Siemens and 3/20 Philips), and IXI-HH (20/20 Philips).
All external datasets were annotated in the same fashion as the TopCoW dataset for the CoW benchmark.

\begin{figure*}[ht!]
    \centering
    \includegraphics[width=0.73\textwidth]{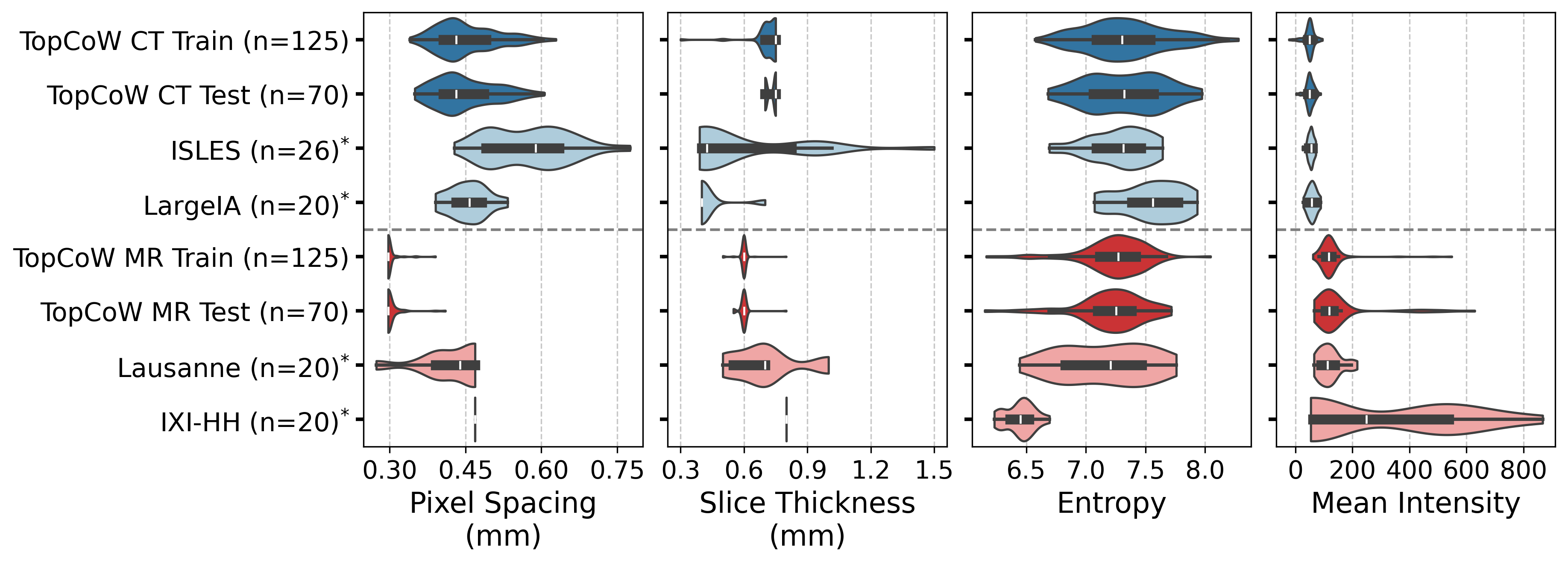}
    \caption{
    Data characteristics of the TopCoW training, internal test, and external multi-center test datasets.
    CTA datasets are in blue violin plots, and MRA datasets are in red.
    Datasets were compared in terms of pixel spacing, slice thickness, entropy, and mean intensity inside the ROI.
    One extreme outlier in mean intensity was removed from TopCoW MRA training set for visualization purposes.
    `n' is number of cases.
    `*' indicates external test datasets.
    }
    \label{fig:data_stats_violin}
\end{figure*}

Fig.~\ref{fig:data_stats_violin} shows the statistical summary of the image information of the training, internal test, and external test data within the ROI.
We compared the voxel dimension, entropy, and image intensity.
TopCoW data had similar training and test distribution.
ISLES and LargeIA datasets had much thinner slice thickness than TopCoW CTA.
Lausanne and IXI-HH datasets had much bigger pixel spacing in the X-Y dimension compare with TopCoW MRA.
Voxel dimensions were quite different among the external datasets.
IXI-HH had a marked lower entropy, which may be due to its dated nature as the images were acquired from around 20 years ago.
IXI-HH also had a very different mean intensity distribution compared to other datasets, with many MR images having ultra-high intensity values.
Overall, the TopCoW internal test images were in-distribution while the external test datasets were out-of-distribution, which is useful for evaluating generalizability.

%% file: tables/table_image_acquisition.tex
\begin{table}[!htbp]
\caption{Imaging characteristics of the TopCoW cohort, including manufacturer, magnetic field strength for MR, and resolution in millimeters.
SD is standard deviations.}
\label{table:img_acquisition}
\centering
\begin{tabular}{lcc}
    \hline
    \textbf{Variable} & \textbf{MR (n=200)} & \textbf{CT (n=200)} \\
    \hline
    \textbf{Manufacturer} & & \\
    Siemens & 190 (95\%) & 199 (99.5\%) \\
    Philips & 10 (5\%) & \\
    GE & & 1 (0.5\%)\\
    \hline
    \textbf{Tesla} & & \\
    3 T & 196 (98\%) & - \\
    1.5 T & 4 (2\%) & - \\
    \hline
    \textbf{Resolution} & & \\
    Spacing (SD) & 0.30 (0.02) & 0.45 (0.06) \\
    Thickness (SD) & 0.60 (0.03) & 0.72 (0.06) \\
    \hline
\end{tabular}
\end{table}

%% file: sections/suppl/annotation_protocol.tex
\section{Details on Annotation \& Verification Protocol}
\label{sec:anno_protocol}

The CoW annotation protocol was designed by a senior neurosurgeon (Y.M., over 10 years of experience) and reviewed by a senior neurosurgeon (P.B., over 15 years of experience) and a senior neurologist (S.W., over 15 years of experience).
Y.M. used around 35 initial patients to educate and train the annotators (K.Y. and F.M.) on the CoW anatomical knowledge and the annotation protocol.
Around another 40 patients from subsequent cases that the annotators were uncertain of were reviewed and verified by Y.M..
Second opinions and verifications were also obtained from other neurosurgeons (P.B., J.H., C.W., E.C.) and neurologists (S.W., L.W., H.B.) for around 15 patients.
All annotated data used in the benchmark were manually verified by at least one annotator.

The annotation protocol on how to segment vessel components and boundaries at bifurcation points such as ACA-ICA-MCA, ACA-Acom, PCA-Pcom, Pcom-ICA, etc., were discussed and agreed upon by the clinical experts.
For example, we marked the superior tip of the ACA-ICA-MCA bifurcations to be part of ICA, and similarly for BA-PCA bifurcation, we marked the tip to be of BA.
We also included the infundibulum as the origin of certain vessel components such as Pcom.
The annotation protocol also covered CoW variants such as fetal PCA, triple ACA etc.

For the TopCoW dataset annotation, since the TopCoW data had both CTA and MRA modalities for the same patients,
the anatomy of the CoW was first inspected in both CTA and MRA
to diagnose the anatomical components.
Then the CoW vessels were annotated or verified for each modality.

CoW annotations for the initial 260 images were manually labeled.
The remaining data used in the benchmark were pre-labeled using a model that was based on the 2023 submission from team `DKFZ' and trained on the initial 260 annotated images.
All images were manually corrected and verified.

For vessels extending beyond the CoW ROI such as the ACA, MCA, and PCA, we typically only labelled until the first major bifurcation occurs,
and we only labelled the main vessel instead of any minor branches.
For the CTA modality, the ICAs were not labelled through the anterior clinoid and sphenoid bone regions, but were labeled starting from the C7 segment in Bouthillier classification system.\supercite{bouthillier1996segments}
For MRA, we labelled the entire curvature of the ICA in the CoW region even in bone regions.

The CoW ROI was drawn manually by the same voxel-level annotators after labeling the vessels and was defined as the 3D bounding box containing the volume required for the diagnosis of the CoW variant with a padding. For higher sensitivity, we padded the bounding box with roughly the diameter of the ICA to include slightly more regions in the ROI.

The annotation of the CoW variant graph was derived from the segmentation mask and encompasses the anterior variant (AV) and posterior variant (PV) graphs.
The AV graph was determined by the presence of four vessel edges: L-A1, Acom, 3rd-A2, and R-A1.
The corresponding edge-list was defined by 0 or 1 according to the edge presence.
For example, AV-1001 is the anterior variant that has L-A1 present, Acom and 3rd-A2 absent, and R-A1 present.
Similarly, the PV graph was determined by a four-element edge-list of L-Pcom, L-P1, R-P1, and R-Pcom.

%% file: sections/suppl/interrater.tex
\section{Inter-Rater Agreement for Voxel-Level Segmentation}
\label{sec:inter_rater_voxel_seg}
\begin{figure*}[!htbp]
    \centering
    \includegraphics[width=0.44\textwidth]{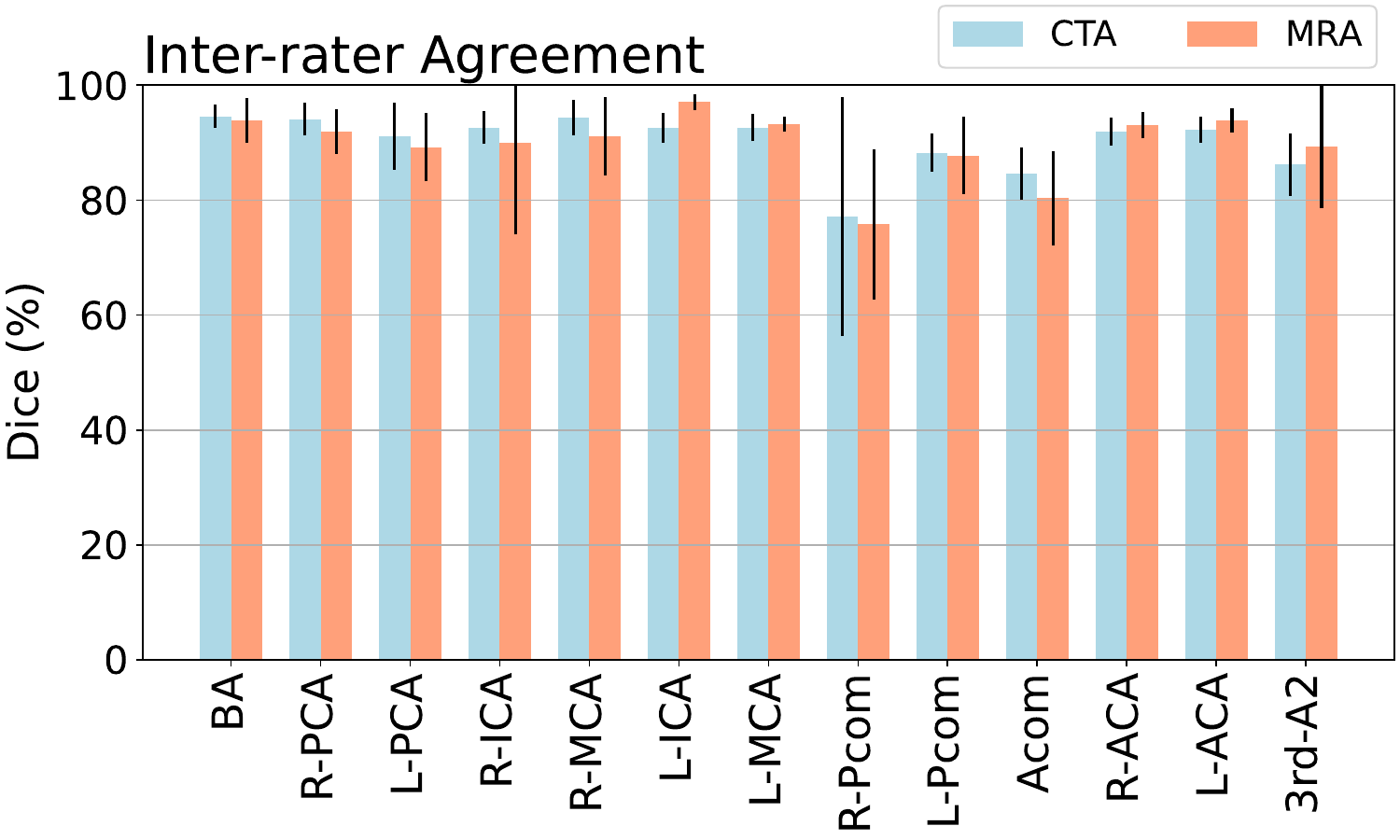}
    \caption{
    Inter-rater agreement for voxel-level segmentation on a subset of the internal test set ($n=5$) for all 13 CoW component classes in Dice scores $\pm$ standard deviations.
    Note that the $n$ for R-Pcom, L-Pcom, and 3rd-A2 are $3,3,3$ for CTA and $4,4,3$ for MRA, respectively.
    }
    \label{fig:interrater_fig}
\end{figure*}

Voxel-level annotations were done on a subset of 5 patients from the TopCoW test set by the two manual annotators (K.Y. and F.M.). These five patients were selected because they each contained most or all of the CoW multiclass labels. Fig.~\ref{fig:interrater_fig} shows the Dice scores for all 13 CoW labels for the CoW anatomical annotations from both annotators. CoW component classes had Dice scores of around 90\% or above, while R-Pcom, L-Pcom, Acom, and 3rd-A2 had slightly lower Dice at 76-89\%. 

Table~\ref{table:interrater_agr_table} shows the per case performance of the inter-rater agreement.
The $90\%$ class-average Dice per case indicated good agreement in raters' multiclass voxel annotations.
The merged binary mask had around $95\%$ Dice which showed good agreement for binary segmentation annotations.
Besides the Dice similarity coefficient,
we also compared the raters' annotations in terms of centerline Dice or clDice \supercite{shit2021cldice} and errors in connected components using zero-th Betti number $\beta_0$.
The clDice and $\beta_0$ errors both had very good inter-rater agreement at near maximum scores as shown in Table~\ref{table:interrater_agr_table}.

\input{tables/table_interrater.tex}

%% file: tables/table_interrater.tex
\begin{table*}[!htbp]
\caption{
Inter-rater agreement results (mean $\pm$ standard deviation) on segmentation of a subset of the internal test set ($n=5$) for MRA and CTA in terms of binary and class average Dice scores, centerline Dice (clDice) scores, binary and class average errors of the zero-th Betti number $\beta_0$. The binary Dice, binary $\beta_0$ errors, and \textit{clDice} scores were computed on the merged binary class.
The class-average \textit{Dice} scores and $\beta_0$ errors were computed for each class separately and the average was taken per case. The arrow indicates the favorable direction.
}
\label{table:interrater_agr_table}
\centering
    \begin{tabular}{llllll}
    \hline
 \multicolumn{6}{c}{\textbf{Inter-rater agreement on segmentation}}\\
    \hline
    \textbf{Modality} & \makecell[l]{\bfseries Binary\\\bfseries Dice (\%) $\uparrow$} & \makecell[l]{\bfseries Per case\\\bfseries class-average\\\bfseries Dice (\%) $\uparrow$} & \bfseries clDice (\%) $\uparrow$ & \makecell[l]{\bfseries Binary\\\bfseries $\beta_0$ error $\downarrow$} & \makecell[l]{\bfseries Per case\\\bfseries class-average\\\bfseries $\beta_0$ error $\downarrow$} \\
    \hline
    CTA & $94.86\pm2.24$ & $90.91\pm3.35$ & $99.72\pm0.24$ & $0\pm0$ & $0\pm0$ \\
    MRA & $96.49\pm2.01$ & $90.21\pm4.02$ & $99.49\pm0.80$ & $0.4\pm0.55$ & $0.07\pm0.07$ \\
    \hline
    \end{tabular}
\end{table*}

%% file: sections/suppl/variant_distribution.tex
\section{CoW Variant Distributions}
\label{sec:cow_var_distribution}

We document the prevalence of all present CoW anterior and posterior variants from our training and test data
in
Table~\ref{table:dataset_ant_variant_distribution}
and 
Table~\ref{table:dataset_post_variant_distribution}.

\input{tables/table_2024_dataset_anterior_variant_distribution}
\input{tables/table_2024_dataset_posterior_variant_distribution}

%% file: tables/table_2024_dataset_anterior_variant_distribution.tex

\begin{table*}[!htbp]
\caption{Distribution of CoW anterior variants (AV) across the TopCoW train and test sets as well as the external datasets. The AV is identified by a four-edge graph, with 0 being absent and 1 being present in the edge-list. The edges are: L-A1, Acom, 3rd-A2, R-A1. Values are reported as absolute counts, with relative percentages in parentheses.}
\label{table:dataset_ant_variant_distribution}
\centering
    \begin{tabular}{l|lllll|l}
    \hline
    \multicolumn{7}{c}{\bfseries Distribution of CoW anterior variants}\\
    \hline
    \bfseries Dataset & \bfseries AV-0101 & \bfseries AV-1001 & \bfseries AV-1100 & \bfseries AV-1101 & \bfseries AV-1111 & \bfseries Total \\
    \hline
    TopCoW CT Train & 3 (2.4\%) & 19 (15.2\%) & 5 (4.0\%) & 83 (66.4\%) & 15 (12.0\%) & 125 \\
    TopCoW CT Test & 0 & 14 (20.0\%) & 3 (4.3\%) & 47 (67.1\%) & 6 (8.6\%) & 70 \\
    ISLES & 0 & 7 (26.9\%) & 1 (3.8\%) & 17 (65.4\%) & 1 (3.8\%) & 26 \\
    LargeIA & 0 & 4 (20.0\%) & 0 & 14 (70.0\%) & 2 (10.0\%) & 20 \\
    \hline
    TopCoW MR Train & 3 (2.4\%) & 19 (15.2\%) & 5 (4.0\%) & 82 (65.6\%) & 16 (12.8\%) & 125 \\
    TopCoW MR Test & 0 & 12 (17.1\%) & 3 (4.3\%) & 48 (68.6\%) & 7 (10.0\%) & 70 \\
    Lausanne & 0 & 4 (20.0\%) & 0 & 14 (70.0\%) & 2 (10.0\%) & 20 \\
    IXI-HH & 1 (5.0\%) & 1 (5.0\%) & 0 & 16 (80.0\%) & 2 (10.0\%) & 20 \\
    \hline
    \end{tabular}
\end{table*}

%% file: tables/table_2024_dataset_posterior_variant_distribution.tex

\begin{table*}[!htbp]
\caption{Distribution of CoW posterior variants (PV) across the TopCoW train and test sets as well as the external datasets. The PV is identified by a four-edge graph, with 0 being absent and 1 being present in the edge-list. The edges are: L-Pcom, L-P1, R-P1, R-Pcom. Values are reported as absolute counts, with relative percentages in parentheses.}
\label{table:dataset_post_variant_distribution}
\centering
    \begin{tabular}{l|lllllllll|l}
    \hline
    \multicolumn{11}{c}{\bfseries Distribution of CoW posterior variants}\\
    \hline
    \bfseries Dataset & \bfseries PV-0101 & \bfseries PV-0110 & \bfseries PV-0111 & \bfseries PV-1001 & \bfseries PV-1010 & \bfseries PV-1011 & \bfseries PV-1101 & \bfseries PV-1110 & \bfseries PV-1111 & \bfseries Total \\
    \hline
    TopCoW CT Train & 3 (2.4\%) & 46 (36.8\%) & 21 (16.8\%) & 2 (1.6\%) & 2 (1.6\%) & 3 (2.4\%) & 5 (4.0\%) & 16 (12.8\%) & 27 (21.6\%) & 125 \\
    TopCoW CT Test & 2 (2.9\%) & 36 (51.4\%) & 12 (17.1\%) & 0 & 0 & 6 (8.6\%) & 0 & 7 (10.0\%) & 7 (10.0\%) & 70 \\
    ISLES & 3 (11.5\%) & 12 (46.2\%) & 3 (11.5\%) & 0 & 0 & 1 (3.8\%) & 0 & 2 (7.7\%) & 5 (19.2\%) & 26 \\
    LargeIA & 0 & 7 (35.0\%) & 5 (25.0\%) & 3 (15.0\%) & 0 & 1 (5.0\%) & 0 & 2 (10.0\%) & 2 (10.0\%) & 20 \\
    \hline
    TopCoW MR Train & 3 (2.4\%) & 47 (37.6\%) & 19 (15.2\%) & 0 & 2 (1.6\%) & 3 (2.4\%) & 4 (3.2\%) & 15 (12.0\%) & 32 (25.6\%) & 125 \\
    TopCoW MR Test & 2 (2.9\%) & 33 (47.1\%) & 12 (17.1\%) & 0 & 0 & 4 (5.7\%) & 0 & 6 (8.6\%) & 13 (18.6\%) & 70 \\
    Lausanne & 1 (5.0\%) & 5 (25.0\%) & 3 (15.0\%) & 0 & 0 & 1 (5.0\%) & 0 & 3 (15.0\%) & 7 (35.0\%) & 20 \\
    IXI-HH & 0 & 7 (35.0\%) & 2 (10.0\%) & 0 & 0 & 1 (5.0\%) & 0 & 2 (10.0\%) & 8 (40.0\%) & 20 \\
    \hline
    \end{tabular}
\end{table*}

%% file: sections/suppl/eval_metrics.tex
\section{Algorithm Submission}
\label{sec:algo_submit}
Our challenge had two tracks for algorithm submissions, namely a CTA track and an MRA track. The main tasks were to multiclass segment the anatomical components of the CoW and to classify the CoW variant graph. The input to the algorithm was the whole 3D image volume, and the evaluation was conducted within the CoW ROI. For all tasks, the input to the submitted algorithm was initially intended to be a pair of CTA and MRA images from a patient due to the paired-modality feature of the TopCoW dataset. Algorithms that only needed one of the modalities could simply ignore the other modality input. In practice, all of the participating algorithms worked with single-modality input, and thus the submissions were also able to be evaluated with the single-modality external test datasets.

The submitted algorithms must be fully-automatic in the form of isolated Docker containers. For internal test data, the Docker containers were run in the cloud on the submission platform that provided an Nvidia T4 GPU with 16GB GPU memory. Submitted algorithms were limited to a runtime of 12-15 minutes per test case for inference on the cloud. Each team was given only one opportunity to upload their containers for the hidden test set. For external test sets, the Docker containers were run locally on a laptop with an RTX 3080 GPU with 16GB GPU memory.


\section{Evaluation Metrics for All Tasks}
\label{sec:all_eval_metrics}

For the 2023 iteration, there were two tasks and both were for segmentation:
multiclass segmentation of the anatomical components of the CoW and binary segmentation of the CoW vessels.
For binary segmentation task, the binary vessel label was generated by merging the multiclass CoW labels.
In 2024, we kept the multiclass segmentation task but discontinued the binary segmentation task from 2023.
We introduced two new tasks in 2024: a detection task for the CoW ROI and a classification task for the CoW variant graphs.

Here we describe in details the metrics used for all the tasks.
The segmentation tasks in 2023 were evaluated using three metrics:
Dice similarity coefficient \supercite{dice1945measures},
centerline Dice (clDice), \supercite{shit2021cldice}
and connected component or zero-th Betti number ($\beta_0$) error \supercite{hu2019topology,menten2023skeletonization}.
In 2024, we added the following segmentation metrics: 
Hausdorff distance at 95th percentile (HD95) \supercite{taha2015metrics,maier2024metrics},
average F1 score for detection of Acom, Pcoms, and 3rd-A2,
variant balanced accuracy (VarBalAcc) of CoW variant graph classification,
and balanced CoW topology match rate (TMR).
The last metric was a custom composite metric combining detection, connectivity, and graph classification metrics:
For labels in anterior or posterior variants,
the predicted segmentation needs to satisfy correct detection,
correct neighborhood connectivity (connected to valid vessel classes),
no 0-th Betti number errors,
and not left-right flipped
in order to be counted as a match in the topology.

The CoW ROI detection task was evaluated using two metrics:
intersection over union (IoU)
and boundary IoU with boundary distance of 20\% of the box dimension \supercite{cheng2021boundaryiou, reinke2024understanding}
between the ground-truth bounding box and the predicted box.

The separate CoW variant classification task required the algorithms to output the variant graph directly instead of a segmentation mask.
The CoW variant classification task shared the same evaluation metric as the second beyond segmentation metric,
which was the VarBalAcc for both anterior and posterior variants.

Our evaluation code with documentation was open sourced at
\href{https://github.com/CoWBenchmark/TopCoW_Eval_Metrics}{https://github.com/CoWBenchmark/TopCoW\_Eval\_Metrics}.

%% file: sections/suppl/ranking.tex
\section{Ranking Analysis for 2024 Segmentation Algorithms}
\label{sec:ranking_robustness}

The algorithms were evaluated on the internal test sets using all the metrics, and their rank positions for each metric were averaged to reach a ranking for the leaderboards. The top performing teams on the averaged rank for CTA or MRA tracks were referred to as ``top teams”. The top teams were further evaluated on the external multi-center test sets for generalizability.

We also created 10 bootstraps of the internal test sets and calculated the rankings on the bootstrapped sets.
Table~\ref{table:task_1_ranking_bootstrapped}
shows the ranking on the original non-bootstrapped set and the average ranking of 10 bootstrapped test sets.
The rankings after bootstraps were stable,
and supported the selection of the top teams.

\input{tables/table_2024_task_1_ranking_bootstrapped}

%% file: tables/table_2024_task_1_ranking_bootstrapped.tex

\begin{table*}[!htbp]
\caption{Mean position (mean $\pm$ standard deviation) of the ranking for the multiclass segmentation task for both CTA and MRA.
The ranking shown is either from the non-bootstrapped internal test set (leaderboard) or from the average across 10 bootstrap runs.
For the leaderboard columns,  the mean position is computed as the average rank position across the 9 evaluation metrics for the segmentation task, and the standard deviation (SD) is across the 9 metric-specific ranks.
For the bootstrap columns, the mean is the average of the mean positions across the 10 runs, and the SD is the root-mean-square of the within-bootstrap SDs across the runs.
The top three values for each column are marked with gold, silver, and bronze colors.
If a team only submitted to one of the tracks, the columns of the other track are filled with `-'.}
\label{table:task_1_ranking_bootstrapped}
\centering
    \begin{tabular}{lll|ll}
        \hline
        \multicolumn{5}{c}{\bfseries TopCoW multiclass segmentation ranking}\\
        \hline
        & \multicolumn{2}{c}{\bfseries CTA}& \multicolumn{2}{c}{\bfseries MRA}\\
        \cline{2-3} \cline{4-5}
        \bfseries Team & \makecell[l]{\bfseries Leaderboard\\\bfseries mean position} & \makecell[l]{\bfseries Bootstrapped\\\bfseries mean position} & \makecell[l]{\bfseries Leaderboard\\\bfseries mean position} & \makecell[l]{\bfseries Bootstrapped\\\bfseries mean position} \\
        \hline
        ARG & $6.00\pm1.25$ & $6.30\pm1.39$ & $5.56\pm1.34$ & $5.72\pm1.31$ \\
        CLAIM & \cellcolor{bronze!30}$3.33\pm2.16$ & \cellcolor{silver!20}$3.28\pm1.97$ & \cellcolor{goldenyellow!60}$2.22\pm1.40$ & \cellcolor{goldenyellow!60}$2.18\pm1.29$ \\
        DeepLearnAI & $10.11\pm1.10$ & $9.92\pm1.09$ & $7.44\pm1.26$ & $7.52\pm1.48$ \\
        DKFZ & $5.44\pm1.71$ & $5.12\pm1.64$ & $4.89\pm1.79$ & $4.57\pm1.78$ \\
        DLaBella29 & $7.33\pm1.49$ & $7.17\pm1.56$ & - & - \\
        HITSZ & $4.67\pm1.70$ & $4.90\pm1.68$ & \cellcolor{bronze!30}$3.00\pm0.94$ & \cellcolor{bronze!30}$3.30\pm1.21$ \\
        IMR & \cellcolor{bronze!30}$3.33\pm2.00$ & $3.58\pm1.77$ & $4.56\pm2.06$ & $4.40\pm2.29$ \\
        junqiangchen & $9.89\pm0.99$ & $9.97\pm0.88$ & $10.00\pm0.00$ & $9.80\pm0.54$ \\
        NIC-VICOROB & \cellcolor{silver!20}$3.22\pm2.15$ & \cellcolor{bronze!30}$3.30\pm2.07$ & $6.11\pm2.23$ & $6.51\pm1.97$ \\
        pamaad & $10.00\pm1.05$ & $10.01\pm1.21$ & $8.22\pm0.92$ & $8.03\pm1.17$ \\
        UB-VTL & $11.78\pm0.42$ & $11.79\pm0.44$ & - & - \\
        UZH & \cellcolor{goldenyellow!60}$2.78\pm1.47$ & \cellcolor{goldenyellow!60}$2.62\pm1.79$ & \cellcolor{silver!20}$2.89\pm2.56$ & \cellcolor{silver!20}$2.87\pm2.23$ \\
        \hline
    \end{tabular}
\end{table*}

%% file: sections/suppl/fetal.tex
\section{Details on Fetal PCA Classification}
\label{sec:details_on_fetal}

\begin{figure*}[!ht]
    \centering
    \includegraphics[width=0.75\textwidth]{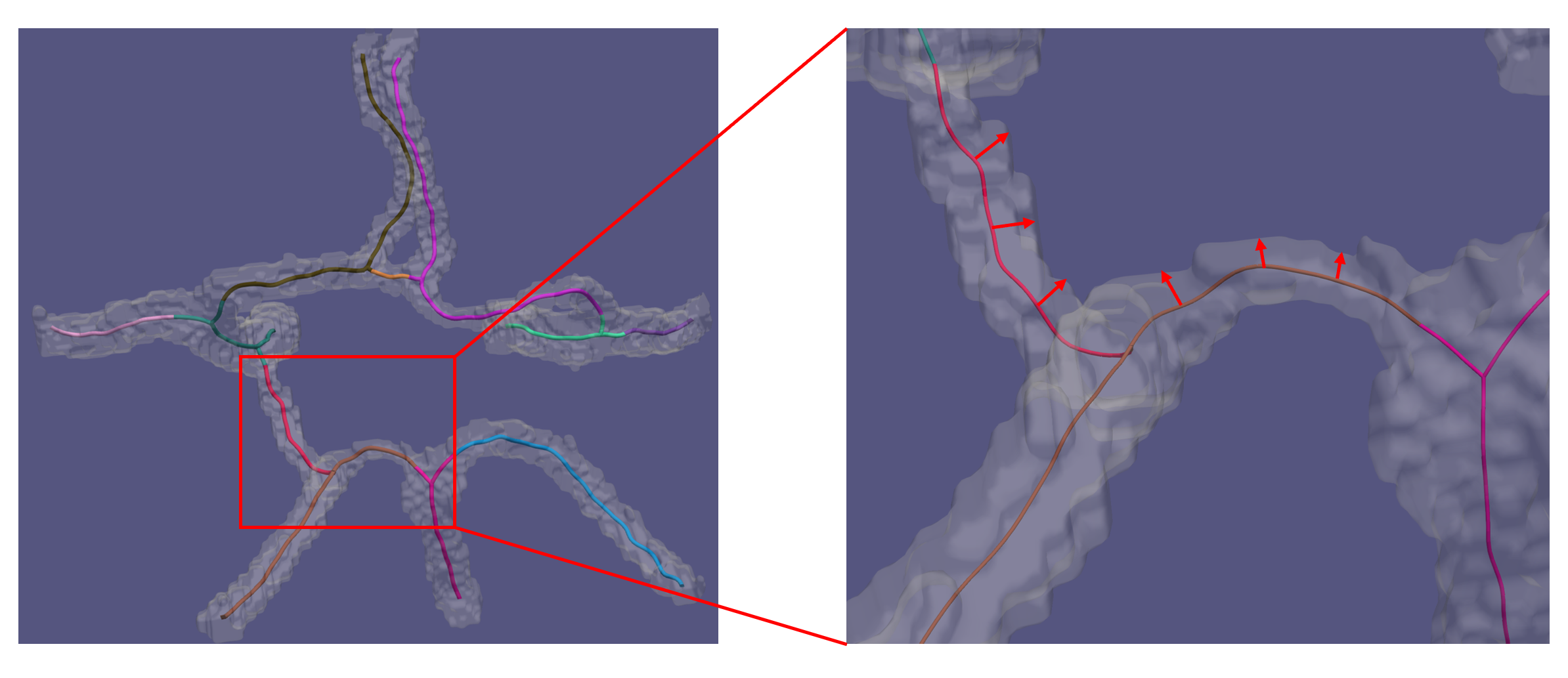}
    \caption{The radius estimation of the Pcom and the P1 segments was based on the centerline graph and the surface mesh of the segmentation mask. The centerline graphs were extracted using \textit{Voreen} \supercite{meyer2009voreen} with the radius given as an attribute for each edge of the graph.
    }
    \label{fig:radius_estimation_fetal}
\end{figure*}

Fig.~\ref{fig:radius_estimation_fetal} shows the intermediate workflow in \textit{Voreen} \supercite{meyer2009voreen} for computing diameters along Pcom and P1 centerlines from segmentation masks for fetal PCA classification.

%% file: sections/suppl/description_all_methods.tex

\section{Descriptions of Submitted Algorithms and Teams}
\label{suppl:all_methods}

In this section we summarize the methods and algorithms of all the participating teams ordered alphabetically by team names:

\paragraph{\textbf{2i\_mtl}}
The team included Emmanuel Montagnon and Laurent Letourneau-Guillon. The team took part in the CT binary task in 2023. They employed a two-stage approach consisting of a patch-based 3D AttentionUNet \supercite{oktay2018AttentionUNet} followed by a 3D autoencoder to mitigate false positives. The autoencoder received as input both an image patch and the AttentionUNet mask prediction.

\paragraph{\textbf{agaldran}}
The submissions were made by Adrian Galdran. He took part in all four tracks and tasks in 2023. His approach was based on a self-adapting 3D dynamic UNet provided by the MONAI library. \supercite{monai2020, cardoso2022monai}  
Internal cross-validation on the volumetric patch size over various sizes was performed.

\paragraph{\textbf{ARG}}
The full team name was `ARG-DEEPNeuro'.
The team consisted of Kwanseok Oh and Dahye Lee. They took part in all three tasks for both MRA and CTA tracks in 2024. For multiclass segmentation task and CoW detection task, they used the 3D nnUNet to segment the CoW vessels. In segmentation task, they added the Generalized Surface Loss \supercite{celaya2023generalized} for CTA track to the default loss functions (Dice and CE) and applied class weight multiplication. For detection task, they trimmed disconnected components before extracting bounding box coordinates. Specifically for the segmentation and detection tasks on the CTA track, models were trained using both MRA and CTA modalities. For variant graph classification task, they used a nnUNet encoder combined with a self-attention mechanism tailored for anterior and posterior predictions.
\textbf{The segmentation algorithm has been included in \href{https://zenodo.org/records/15665435}{our Zenodo Docker release (records/15665435)}.}

\paragraph{\textbf{CLAIM}}
The team was composed of Orhun Utku Aydin, Adam Hilbert, Jana Rieger, Dimitrios Rallios, Satoru Tanioka, and Dietmar Frey.
They participated in all three tasks for both MRA and CTA tracks in 2024. All models were trained on mixed MRA and CTA scans. For multiclass segmentation task, they used a combined dataset of TopCoW and 500 additional scans enriched for rare anatomical variants. A two-stage approach was employed: an initial ROI detection step using their task 2 detection model, followed by segmentation with the 3D nnUNet within the enlarged ROI. To improve topological connectedness, they incorporated the Skeleton Recall (SkelRecall). loss \supercite{kirchhoff2024skeleton} For CoW detection task, they used YOLOv8 by Ultralytics \supercite{Jocher_Ultralytics_YOLO_2023} trained on 2D slices to detect ROIs, which were aggregated into 3D boxes. For CoW graph classification task, they implemented a rule-based graph extraction algorithm operating on the output of segmentation task. Their segmentation algorithm and a list of all the additional dataset references are available at \url{https://github.com/claim-berlin/TopCoW_2024_MRA_winning_solution}.
\textbf{The segmentation algorithm has been included in \href{https://zenodo.org/records/15665435}{our Zenodo Docker release (records/15665435)}.}

\paragraph{\textbf{DeepLearnAI}}
The team was composed of Abdul Qayyum, Moona Mazher, and Steven A. Niederer.
They participated in the multiclass segmentation task for both MRA and CTA tracks in 2024. They used a two-stage approach: the first stage employed self-supervised learning using DINOv2 \supercite{oquab2023dinov2} to pretrain representations, and the second stage used a 3D xLSTM-UNet model \supercite{dutta2024segmentation} for downstream segmentation. The pipeline was implemented using MONAI.

\paragraph{\textbf{DKFZ}}
Also known as team `WilliWillsWissen' in 2023.
The submissions were made by Maximilian R. Rokuss, Yannick Kirchhoff, Nico Disch, Julius C. Holzschuh, Fabian Isensee and Klaus Maier-Hein.
The team took part in the segmentation tasks for both MRA and CTA tracks in both 2023 and 2024.
They used a patch-based 3D nnUNet-with various adaptations and trained on both modalities.
To improve connectedness they incorporated the clDice \supercite{shit2021cldice} and a recall on the skeleton of the vessels (SkelRecall); and they employed extensive cross-validation ensemble per model as well as a subsequent ensemble of differently trained models for each submission.
\textbf{The segmentation algorithm has been included in \href{https://zenodo.org/records/15665435}{our Zenodo Docker release (records/15665435)}.}

\paragraph{\textbf{DLaBella29}}
The submissions were made by Dominic LaBella, who participated in the multiclass segmentation task for both MRA and CTA tracks in 2024. He used a 3D SegResNet \supercite{myronenko20183d} via MONAI and Auto3DSeg. A connectivity check ensured R/L-Pcom contact with the PCA and ICA, and postprocessing included removal of distant or small components.

\paragraph{\textbf{EURECOM}}
The team consisted of Francesco Galati, Daniele Falcetta and Maria A. Zuluaga. They took part in both binary segmentation tasks in 2023 and applied a single model strategy for multi-domain vessel segmentation. They employed an adapted version of their A2V framework \supercite{galati2023A2V} consisting of a single encoder-generator architecture for image reconstruction, translation, and ultimately segmentation with a shared latent space for both modalities. In a first step they extracted brain masks required by A2V using SynthSeg. \supercite{billot2023synthseg} The code is available at \url{https://github.com/i-vesseg/MultiVesSeg}.

\paragraph{\textbf{gbCoW}}
The team was made up of Chaolong Lin and Haoran Zhao. They took part in the MRA binary and multiclass tasks in 2023 using the nnUNet framework for patch-based 3D segmentation. They trained a single model on the multiclass labels only; the binary masks were obtained from their multiclass predictions.

\paragraph{\textbf{gl}}
The submissions were made by Zehan Zhang. He submitted algorithms to both multiclass segmentation tasks in 2023 following a multi-step approach consisting of 1) the extraction of a custom ROI using a dataset specific atlas and affine registration, 2) binary segmentation and 3) subsequent multiclass segmentation with a 2-channel input (image ROI and binary mask). For the segmentation, both the 3D MedNexT \supercite{roy2023mednext} and UX-Net \supercite{lee2022uxnet} architectures were employed as an esemble. The inference code is available at \url{https://github.com/zzh980123/TopCoW_Algo_Submission}.

\paragraph{\textbf{HITSZ}}
Also known as team `NexToU' in 2023.
The team consisted of Pengcheng Shi, Wei Liu and Ting Ma.
They participated in the segmentation tasks for both MRA and CTA tracks in both 2023 and 2024. They used a two-stage pipeline: a low-resolution model trained on binary labels using the 3D nnUNet, followed by a full-resolution model trained on multiclass labels using their own NexToU architecture. \supercite{shi2023nextou} Both stages were trained on MRA and CTA data. Besides the default losses, they used the cbDice loss, \supercite{shi2024cbDice} which is both topology-aware and diameter-balanced, and a hierarchical topological interaction loss. The code for NexToU is at \url{https://github.com/PengchengShi1220/NexToU}; the cbDice loss function is available at \url{https://github.com/PengchengShi1220/cbDice}.
\textbf{The segmentation algorithm has been included in \href{https://zenodo.org/records/15665435}{our Zenodo Docker release (records/15665435)}.}

\paragraph{\textbf{IMR}}
The team was composed of Minghui Zhang, Xin You, Hanxiao Zhang, Guang-Zhong Yang, and Yun Gu.
They participated in all three tasks for both MRA and CTA tracks in 2024. All models were based on the 3D nnUNet and trained on both modalities. They optimized for the multi-modality training by intensity normalization using intensity truncation, with different settings of the truncation values applied for CTA (-1000 to 1800) and MRA (0 to 700) scans. For multiclass segmentation task, they further developed the connectivity-aware loss (CAL) based on \supercite{zhang2023towards} to improve topological completeness, and applied a topology-aware refinement postprocessing step to repair disconnected vessel components. Detection task was reformulated as segmentation followed by bounding box extraction. Classification task was inferred directly from segmentation outputs without a separate classification head. A detailed method description can be found on their ArXiv pre-print. \supercite{imr2024topology}
\textbf{The segmentation algorithm has been included in \href{https://zenodo.org/records/15665435}{our Zenodo Docker release (records/15665435)}.}

\paragraph{\textbf{IWantToGoToCanada}}
The team consisted of Sinyoung Ra, Jongyun Hwang and Hyunjin Park. They took part in the CTA binary and multiclass tasks in 2023. The nnUNet was used to extract the binary segmentation mask. For the subsequent multiclass segementation a 3D Swin-UNETR \supercite{hatamizadeh2021SwinUNETR} architecture was employed with both the image and the binary mask as input.

\paragraph{\textbf{junqiangchen}}
The submissions were made by Junqiang Chen, taking part in all tasks and tracks in both 2023 and 2024. In 2023, a two-stage approach was employed using the VNet3D \supercite{milletari2016VNet} for both stages: In the first stage a custom ROI was extracted based on a binary segmentation, in the second stage the segmentation was performed on the ROI only.
In 2024, all models were trained using mixed MRA and CTA data. For segmentation and detection tasks, he used a 3D VNet architecture to segment the CoW vessels. For classification task, a 3D ResNet was used. The code is available on \url{https://github.com/junqiangchen/PytorchDeepLearing}.

\paragraph{\textbf{lWM}}
The team was composed of Marek Wodzinski and Henning Müller. They took part in all four tracks and tasks in 2023 using a patch-based 3D ResidualUNet with a focus on data preprocessing and augmentation.

\paragraph{\textbf{NantesU}}
Nesrin Mansouri and Florent Autrusseau participated in the segmentation task for the MRA track only in 2024, using the 3D nnUNet. To address class imbalance, they generated 504 synthetic MRA images focusing on underrepresented arteries, expanding the dataset from 125 to 629 images. Their synthetic modeling approach was based on prior work. \supercite{autrusseau2022toward, nader2024vascular}

\paragraph{\textbf{NIC-VICOROB}}
Also known as team `NIC-VICOROB-1' in 2023.
The team was made up of Cansu Yalcin, Rachika E. Hamadache, Clara Lisazo, Joaquim Salvi, Adrià Casamitjana, and Xavier Llad\'{o}. In 2023, the team took part in all tracks and tasks using a patch-based 3D nnUNet. Working in two stages with a 2-channel input consisting of both the image and the binary mask improved the segmentation results for both the CTA and MRA multiclass tasks.
In 2024, they participated in segmentation and detection tasks for both tracks. Models based on 3D nnUNnet were trained on MRA only for the MRA track, and on both modalities for the CTA track. For multiclass segmentation task, a two-stage pipeline was similar to 2023 was used: multiclass segmentation followed by binary segmentation for postprocessing. For detection task, CoW detection was treated as a segmentation task using binarized masks. For topological optimization they used the Skeleton Recall (SkelRecall) loss and did postprocessing on the predictions including background filling, small component removal, and connectivity restoration.
\textbf{The segmentation algorithm has been included in \href{https://zenodo.org/records/15665435}{our Zenodo Docker release (records/15665435)}.}

\paragraph{\textbf{NIC-VICOROB-2}}
In 2023, NIC-VICOROB had a second team consisting of Uma Maria Lal-Trehan Estrada, Valeriia Abramova, Luca Giancardo and Arnau Oliver, taking part in all four tracks and tasks. For the binary segmentation tasks they employed a patch-based 3D AttentionUNet. For the multiclass segmentation tasks they employed a two-stage approach using a 2D AttentionUNet with full axial slices and binary segmentation masks, obtained from the 3D AttentionUNet, as input. The 2D approach was chosen due to GPU memory limitations. 

\paragraph{\textbf{pamaad}}
The team was composed of Paula Casademunt, Adrian Galdran, and Matteo Delucchi.
They participated in all three tasks for both MRA and CTA tracks in 2024. For segmentation and detection tasks, they used 3D nnUNet with a novel centerlineCE loss. \supercite{acebes2024centerline} For segmentation task, they used a two stage approach involving an initial segmentation for custom ROI cropping followed by multiclass segmentation on the ROI. For classification task, they used a Video Swin Transformer \supercite{liu2022VideoSwinT} adapted to 3D medical data.

\paragraph{\textbf{refrain}}
The submissions were made by Jialu Liu, Haibin Huang and Yue Cui. They submitted algorithms for the MRA binary and multiclass task, employing a 3D nnUNet for both tasks.  A template atlas was used to extract a custom ROI via registration. Furthermore, they used data augmentation to balance the training set with respect underrepresented CoW variants and applied segment specific loss weighting with higher weights for R-Pcom, L-Pcom, Acom and 3rd-A2. The code is available on \url{https://github.com/Vessel-Segmentation/Topcow_private}.

\paragraph{\textbf{sjtu\_eiee\_2-426lab}}
The submissions were made by Zehang Lin, Yusheng Liu and Shunzhi Zhu, taking part in the CTA binary and multiclass tasks. They used a two-stage approach using the 3D nnUNet: A first binary segmentation for a custom ROI extraction followed by a segmentation on the extracted ROI only.

\paragraph{\textbf{SynthCLAIM}}
CLAIM had a second team that included Alexander Koch.
They participated in multiclass segmentation task of the MRA track only in 2024. To generate training data, they trained a StyleGAN \supercite{karras2019styleGan} on the TopCoW MRA scans and synthesized 10,000 TOF-MRA volumes. These were pseudo-labeled using the nnUNet segmentation model from team CLAIM. A two-stage pipeline was then trained exclusively on this synthetic dataset: first, a 2D YOLO-based \supercite{redmon2016yolo} model was trained to detect the CoW ROI slice-wise; second, a 3D nnUNet performed multiclass segmentation within the detected ROIs. To enhance topological accuracy, the nnUNet incorporated the Skeleton Recall (SkelRecall) loss.

\paragraph{\textbf{UB-VTL}}
The team consisted of Tatsat R. Patel, Adnan H. Siddiqui, and Vincent M. Tutino. 
In 2023, they participated in binary segmentation tasks employing a patch-based 3D BRAVE-NET \supercite{hilbert2020bravenet} taking as input a normal patch and a low-res patch for more context. As modifications, they added residual connections, used parametric rectified linear units (PReLu) as activations and worked with the centerline Dice (clDice) as a loss function.
In 2024, they participated in the segmentation and detection tasks for the CTA track only. For segmentation task, they used a two-stage pipeline based on BRAVE-NET. Stage 1 performed binary segmentation of CoW vessels using the clDice loss; stage 2 used a modified BRAVE-NET with the stage 1 binary mask as additional input to perform multiclass segmentation. For detection task, they used a two-stage atlas-based method without deep learning: CTA images were registered to a CT atlas, \supercite{ctatlas2020ubvtl} and the atlas ROI was computed using the STAPLE \supercite{warfield2004staple} algorithm. During inference, the atlas ROI was registered to test images, and Cartesian ray tracing was used to predict CoW coordinates.

\paragraph{\textbf{UW}}
The submission was made by Maysam Orouskhani, Huayu Wang, Mahmud Mossa-Basha and Chengcheng Zhu. They took part in the MRA binary task in 2023 using the patch-based 3D nnUNet framework with a modified 3-component loss function consisting of Dice, Cross-Entropy (CE) and TopK loss. The code, models and trained weights can be accessed via \url{https://github.com/orouskhani/TopCow2023}.

\paragraph{\textbf{UZH}}
Also known as team `Organizers' in 2023.
The team was composed of Houjing Huang, Fabio Musio, Chinmay Prabhakar, Suprosanna Shit, and Kaiyuan Yang.
In 2023, the team participated in the multiclass tasks of both tracks using a two-stage approach: stage-1 detection of the CoW ROIs with nnDetection \supercite{baumgartner2021nndetection} based on the binary vessel labels and stage-2 multiclass segmentation on the ROIs with 3D nnUNet. Additionally, inter-modal registration was used as a data augmentation strategy, registering all the image pairs and thereby doubling the size of the training set for both modalities.
In 2024, the team upgraded the detection module with a nnUNet segmentation model for the binary `brick' masks enclosed by the CoW ROI bounding box coordinates.
All segmentation models were now trained on mixed modalities.
Inter-modal registration for data augmentation was kept and used to double the training set.
The stage-2 segmentation also added the Skeleton Recall (SkelRecall) loss.
For classification task, again a two-stage method was applied: binary vessel segmentation using nnUNet, followed by edge prediction. For the edge prediction, the binary segmentation mask was converted into a graph using Voreen, \supercite{meyer2009voreen, drees2021scalable} and then edge labels were predicted based on topological properties of the graph.
\textbf{The segmentation algorithm has been included in \href{https://zenodo.org/records/15665435}{our Zenodo Docker release (records/15665435)}.}

\paragraph{\textbf{ysato}}
The submission was made by Yuki Sato, taking part in the MRA binary task. The author employed a non-deep learning approach based on recursive algorithm consisting of auto vessel thresholding and region growing with a rule-based automated seed point selection. It was the only non-deep learning based submission to our challenge. Accordingly, the inference time per case was very short ($\sim$15s) and the computations could be done on a CPU.

%% file: sections/suppl/results24.tex
\section{Detailed Results of the 2024 Tasks for Internal Test}
\label{sec:detailed_2024_INternal_results}

Table~\ref{table:task_1_seg_results_cta_mra_internal_2024},
Table~\ref{table:task_2_det_results_cta_mra_internal_2024},
and 
Table~\ref{table:task_3_edg_results_cta_mra_internal_2024}
show the results of the 2024 multiclass segmentation task,
the detection task for the CoW ROI,
and the classification task for the CoW variant graphs
on the TopCoW internal test data.

\input{tables/table_2024_task_1_seg_results_ct_mr_internal}

\input{tables/table_2024_task_2_det_results_ct_mr_internal}

\input{tables/table_2024_task_3_edg_results_ct_mr_internal}

\section{Detailed Results of the 2024 Tasks for External Test}
\label{sec:detailed_2024_EXternal_results}

Table~\ref{table:task_1_seg_results_cta_mra_external_2024},
Table~\ref{table:task_2_det_results_cta_mra_external_2024}, and
Table~\ref{table:task_3_edg_results_cta_mra_external_2024}
show the results of the 2024 tasks on the external test data.

\input{tables/table_2024_task_1_seg_results_ct_mr_external}

\input{tables/table_2024_task_2_det_results_ct_mr_external}

\input{tables/table_2024_task_3_edg_results_ct_mr_external}

%% file: tables/table_2024_task_1_seg_results_ct_mr_internal.tex
\begin{table*}[!htbp]
\caption{Results (mean $\pm$ standard deviation) of the CoW multiclass segmentation task from 2024 on the TopCoW internal test set in terms of class-average Dice similarity coefficient, centerline Dice (clDice) on merged binary mask, class-average 0-th Betti number ($\beta_0$) error, class-average Hausdorff Distance 95th Percentile (HD95), average F1 score for detection of communicating arteries and 3rd-A2, variant-balanced accuracy (VarBalAcc) of graph classification,
and CoW variant topology match rate (TMR) for both the anterior and posterior variants. The arrow indicates the favorable direction. The top three values for each metric are marked in gold, silver, and bronze colors. Teams marked with a `*' were late submissions due to technical issues. The bottom two rows for each modality are the results for the top-3 teams ensemble and the organizer teams (`UZH', `DKFZ', `HITSZ') ensemble.}
\label{table:task_1_seg_results_cta_mra_internal_2024}
\centering
\resizebox{1\textwidth}{!}{
    \begin{tabular}{llllllcccc}
        \hline
        \multicolumn{10}{c}{\bfseries Multiclass segmentation performance on the 2024 TopCoW internal test data}\\
        \hline
        \multicolumn{10}{c}{\bfseries CTA (n=70)}\\
        \hline
        \bfseries Team & \makecell[l]{\bfseries Per case\\\bfseries class-avg\\\bfseries Dice (\%) $\uparrow$} & \bfseries clDice (\%) $\uparrow$ & \makecell[l]{\bfseries Per case\\\bfseries class-avg\\\bfseries $\beta_0$ error $\downarrow$} &  \makecell[l]{\bfseries Per case\\\bfseries class-avg\\\bfseries HD95 $\downarrow$} & \makecell[l]{\bfseries Average F1\\\bfseries score (\%) $\uparrow$} & \makecell[l]{\bfseries Anterior\\\bfseries VarBalAcc (\%) $\uparrow$} & \makecell[l]{\bfseries Posterior\\\bfseries VarBalAcc (\%) $\uparrow$} & \makecell[l]{\bfseries Anterior\\\bfseries TMR (\%) $\uparrow$} & \makecell[l]{\bfseries Posterior\\\bfseries TMR (\%) $\uparrow$}\\
        \hline
        ARG & $85.05\pm6.74$ & $98.75\pm1.28$ & $0.13\pm0.14$ & $4.85\pm5.03$ & $77.19\pm9.16$ & $73.94$ & $65.87$ & $39.98$ & $42.79$ \\
        CLAIM & $84.92\pm5.63$ & \cellcolor{goldenyellow!60}$99.00\pm1.15$ & $0.06\pm0.09$ & \cellcolor{silver!20}$3.03\pm3.87$ & \cellcolor{bronze!30}$83.11\pm9.63$ & $73.94$ & \cellcolor{goldenyellow!60}$83.60$ & $45.21$ & \cellcolor{goldenyellow!60}$69.97$ \\
        DeepLearnAI & $73.39\pm22.52$ & $96.38\pm3.55$ & $0.23\pm0.24$ & $11.61\pm10.71$ & $21.63\pm37.47$ & $66.45$ & $16.67$ & $24.01$ & $12.96$ \\
        DKFZ & $86.04\pm6.43$ & \cellcolor{bronze!30}$98.90\pm1.27$ & $0.10\pm0.10$ & $3.97\pm4.33$ & $73.76\pm17.47$ & $79.95$ & $65.41$ & $48.13$ & $41.87$ \\
        DLaBella29 & $84.14\pm6.28$ & $98.16\pm2.09$ & $0.07\pm0.08$ & $5.20\pm5.01$ & $78.09\pm10.56$ & $60.03$ & $52.78$ & $32.28$ & $43.06$ \\
        HITSZ & $85.03\pm6.55$ & $98.39\pm2.22$ & $0.10\pm0.11$ & $3.76\pm4.20$ & $79.50\pm9.44$ & \cellcolor{silver!20}$86.03$ & \cellcolor{bronze!30}$71.36$ & \cellcolor{bronze!30}$48.51$ & $49.80$ \\
        IMR & \cellcolor{bronze!30}$87.13\pm5.68$ & $98.60\pm1.64$ & \cellcolor{bronze!30}$0.06\pm0.09$ & \cellcolor{goldenyellow!60}$2.88\pm3.71$ & \cellcolor{goldenyellow!60}$86.01\pm8.17$ & \cellcolor{goldenyellow!60}$88.88$ & $62.63$ & $46.59$ & $49.93$ \\
        junqiangchen & $74.81\pm7.19$ & $96.96\pm2.56$ & $0.34\pm0.24$ & $9.72\pm6.51$ & $48.62\pm30.14$ & $47.34$ & $44.51$ & $7.98$ & $16.73$ \\
        NIC-VICOROB & \cellcolor{goldenyellow!60}$88.01\pm5.98$ & \cellcolor{silver!20}$98.94\pm1.41$ & \cellcolor{silver!20}$0.04\pm0.06$ & $3.45\pm4.43$ & $74.97\pm26.96$ & $72.21$ & \cellcolor{silver!20}$74.74$ & \cellcolor{silver!20}$57.05$ & \cellcolor{silver!20}$62.96$ \\
        pamaad & $64.66\pm29.64$ & $98.23\pm2.02$ & $0.31\pm0.34$ & $11.34\pm10.83$ & $57.44\pm16.13$ & $45.09$ & $38.49$ & $12.42$ & $13.96$ \\
        UB-VTL & $65.60\pm9.29$ & $88.87\pm7.02$ & $1.10\pm0.72$ & $17.92\pm7.87$ & $0.00\pm0.00$ & $19.64$ & $12.04$ & $8.93$ & $10.65$ \\
        UZH & \cellcolor{silver!20}$87.37\pm5.89$ & $98.57\pm2.42$ & \cellcolor{goldenyellow!60}$0.04\pm0.05$ & \cellcolor{bronze!30}$3.22\pm4.25$ & \cellcolor{silver!20}$85.78\pm5.32$ & \cellcolor{bronze!30}$84.37$ & $68.06$ & \cellcolor{goldenyellow!60}$73.91$ & \cellcolor{bronze!30}$57.74$ \\
        \hline
        Top-3 ensemble & $88.49\pm5.80$ & $98.84\pm1.54$ & $0.05\pm0.08$ & $3.04\pm4.05$ & $84.11\pm10.34$ & $84.71$ & $68.72$ & $52.36$ & $56.02$ \\
        Orgs ensemble & $87.56\pm6.04$ & $98.89\pm1.44$ & $0.08\pm0.09$ & $3.23\pm3.96$ & $80.21\pm12.40$ & $86.56$ & $63.49$ & $70.15$ & $42.79$ \\
        \hline
        \multicolumn{10}{c}{}\\
        \multicolumn{10}{c}{\bfseries MRA (n=70)}\\
        \hline
        \bfseries Team & \makecell[l]{\bfseries Per case\\\bfseries class-avg\\\bfseries Dice (\%) $\uparrow$} & \bfseries clDice (\%) $\uparrow$ & \makecell[l]{\bfseries Per case\\\bfseries class-avg\\\bfseries $\beta_0$ error $\downarrow$} &  \makecell[l]{\bfseries Per case\\\bfseries class-avg\\\bfseries HD95 $\downarrow$} & \makecell[l]{\bfseries Average F1\\\bfseries score (\%) $\uparrow$} & \makecell[l]{\bfseries Anterior\\\bfseries VarBalAcc (\%) $\uparrow$} & \makecell[l]{\bfseries Posterior\\\bfseries VarBalAcc (\%) $\uparrow$} & \makecell[l]{\bfseries Anterior\\\bfseries TMR (\%) $\uparrow$} & \makecell[l]{\bfseries Posterior\\\bfseries TMR (\%) $\uparrow$}\\
        \hline
        ARG & $87.69\pm6.34$ & $98.88\pm1.51$ & $0.12\pm0.13$ & $3.72\pm4.56$ & $87.12\pm6.15$ & \cellcolor{bronze!30}$85.94$ & $65.87$ & $41.00$ & $41.60$ \\
        CLAIM & $87.60\pm5.98$ & \cellcolor{goldenyellow!60}$99.16\pm1.34$ & \cellcolor{bronze!30}$0.05\pm0.07$ & \cellcolor{goldenyellow!60}$1.50\pm2.53$ & \cellcolor{silver!20}$91.51\pm4.50$ & \cellcolor{silver!20}$89.14$ & \cellcolor{silver!20}$77.49$ & \cellcolor{silver!20}$60.12$ & \cellcolor{bronze!30}$59.75$ \\
        DeepLearnAI & $82.70\pm20.38$ & $98.74\pm1.80$ & $0.18\pm0.23$ & $5.40\pm9.77$ & $66.91\pm39.05$ & $66.67$ & $70.65$ & $29.17$ & $35.81$ \\
        DKFZ & $89.07\pm5.63$ & \cellcolor{silver!20}$99.06\pm1.40$ & $0.10\pm0.11$ & $2.48\pm3.40$ & $86.16\pm5.62$ & $84.45$ & $63.85$ & $48.81$ & $41.98$ \\
        DLaBella29* & $84.63\pm10.18$ & $96.13\pm5.44$ & $0.08\pm0.10$ & $6.18\pm7.82$ & $75.14\pm5.22$ & $71.43$ & $50.67$ & $37.80$ & $50.17$ \\
        HITSZ & \cellcolor{silver!20}$89.36\pm5.45$ & \cellcolor{bronze!30}$99.04\pm1.44$ & $0.09\pm0.11$ & \cellcolor{bronze!30}$2.46\pm3.39$ & $90.04\pm2.59$ & \cellcolor{bronze!30}$85.94$ & $71.30$ & \cellcolor{bronze!30}$53.27$ & $52.60$ \\
        IMR & $87.88\pm6.88$ & $98.43\pm2.23$ & $0.09\pm0.13$ & $2.57\pm3.66$ & \cellcolor{bronze!30}$91.23\pm2.03$ & \cellcolor{goldenyellow!60}$90.62$ & $69.24$ & $52.31$ & $49.05$ \\
        junqiangchen & $77.06\pm7.01$ & $98.06\pm1.98$ & $0.33\pm0.26$ & $8.67\pm5.96$ & $49.97\pm29.93$ & $48.96$ & $44.13$ & $8.33$ & $7.09$ \\
        NantesU* & $79.53\pm23.84$ & $98.33\pm2.02$ & $0.21\pm0.50$ & $5.52\pm9.02$ & $84.51\pm3.18$ & $74.55$ & $66.57$ & $42.56$ & $36.00$ \\
        NIC-VICOROB & $80.26\pm22.14$ & $98.79\pm1.59$ & $0.08\pm0.11$ & $6.40\pm8.58$ & $77.44\pm14.13$ & $60.64$ & $71.45$ & $35.79$ & $50.41$ \\
        NIC-VICOROB* & \cellcolor{goldenyellow!60}$89.58\pm5.44$ & $98.96\pm1.50$ & \cellcolor{goldenyellow!60}$0.03\pm0.05$ & \cellcolor{silver!20}$2.31\pm3.30$ & $89.86\pm7.51$ & $82.96$ & \cellcolor{bronze!30}$72.31$ & $52.38$ & \cellcolor{silver!20}$61.79$ \\
        pamaad & $77.64\pm22.41$ & $98.65\pm1.92$ & $0.19\pm0.21$ & $7.54\pm10.00$ & $78.11\pm6.56$ & $71.43$ & $62.42$ & $27.98$ & $39.33$ \\
        SynthCLAIM* & $60.23\pm22.83$ & $83.52\pm28.37$ & $0.72\pm0.44$ & $16.78\pm25.33$ & $49.37\pm28.87$ & $52.60$ & $46.87$ & $2.60$ & $3.03$ \\
        UZH & \cellcolor{bronze!30}$89.35\pm5.53$ & $98.58\pm2.34$ & \cellcolor{silver!20}$0.03\pm0.05$ & $2.60\pm3.62$ & \cellcolor{goldenyellow!60}$91.74\pm3.60$ & $81.92$ & \cellcolor{goldenyellow!60}$82.96$ & \cellcolor{goldenyellow!60}$60.19$ & \cellcolor{goldenyellow!60}$70.78$ \\
        \hline
        Top-3 ensemble & $90.12\pm5.72$ & $99.05\pm1.68$ & $0.06\pm0.09$ & $1.94\pm3.14$ & $92.13\pm4.53$ & $83.93$ & $88.70$ & $57.44$ & $61.35$ \\
        Orgs ensemble & $90.08\pm5.80$ & $99.03\pm1.51$ & $0.09\pm0.11$ & $1.96\pm3.09$ & $88.18\pm6.62$ & $88.02$ & $65.37$ & $49.78$ & $47.95$ \\
        \hline
    \end{tabular}
}
\end{table*}

%% file: tables/table_2024_task_2_det_results_ct_mr_internal.tex
\begin{table*}[!htbp]
\caption{Results (mean $\pm$ standard deviation) of the CoW ROI detection task from 2024 on the TopCoW internal test data in terms of intersection over union (IoU) and boundary IoU. The top three values for each metric are marked in gold, silver and bronze colors. If a team only submitted to one of the tracks the columns of the other track are filled with a `-'.}
\label{table:task_2_det_results_cta_mra_internal_2024}
\centering
    \begin{tabular}{lll|ll}
    \hline
    \multicolumn{5}{c}{\bfseries CoW ROI detection performance on the 2024 TopCoW internal test data}\\
    \hline
     & \multicolumn{2}{c}{\bfseries CTA (n=70)}& \multicolumn{2}{c}{\bfseries MRA (n=70)}\\
    \cline{2-3} \cline{4-5}
    \bfseries Team & \bfseries IoU (\%) $\uparrow$ & \makecell[l]{\bfseries Boundary\\\bfseries IoU (\%) $\uparrow$} & \bfseries IoU (\%) $\uparrow$ & \makecell[l]{\bfseries Boundary\\\bfseries IoU (\%) $\uparrow$} \\
    \hline
    ARG & \cellcolor{bronze!30}$74.20\pm7.08$ & \cellcolor{bronze!30}$61.93\pm9.53$ & \cellcolor{bronze!30}$80.30\pm6.08$ & \cellcolor{bronze!30}$70.08\pm8.93$ \\
    CLAIM & $72.45\pm7.79$ & $59.61\pm10.40$ & $76.58\pm6.34$ & $65.12\pm8.62$ \\
    IMR & $74.19\pm7.63$ & $61.86\pm10.26$ & $77.40\pm9.97$ & $66.77\pm11.43$ \\
    junqiangchen & $72.13\pm9.59$ & $59.95\pm11.61$ & $1.06\pm2.27$ & $0.96\pm1.59$ \\
    NIC-VICOROB & \cellcolor{silver!20}$76.42\pm6.52$ & \cellcolor{silver!20}$65.14\pm8.80$ & \cellcolor{silver!20}$81.54\pm6.77$ & \cellcolor{silver!20}$71.97\pm9.56$ \\
    pamaad & $33.59\pm6.54$ & $16.24\pm6.48$ & $76.58\pm9.46$ & $65.63\pm11.30$ \\
    UB-VTL & $63.14\pm11.20$ & $48.05\pm13.47$ & - & - \\
    UZH & \cellcolor{goldenyellow!60}$79.31\pm7.52$ & \cellcolor{goldenyellow!60}$69.47\pm10.38$ & \cellcolor{goldenyellow!60}$84.84\pm5.11$ & \cellcolor{goldenyellow!60}$77.08\pm7.27$ \\
    \hline
    \end{tabular}
\end{table*}

%% file: tables/table_2024_task_3_edg_results_ct_mr_internal.tex
\begin{table*}[!htbp]
\caption{Results of the CoW variant graph classification task from 2024 on the TopCoW internal test set in terms of variant-balanced accuracy (VarBalAcc) for both the anterior and the posterior variant. The top three values for each metric are marked in gold, silver and bronze colors. Teams marked with * were segmentation-based and \# were classification-based submissions.}
\label{table:task_3_edg_results_cta_mra_internal_2024}
\centering
    \begin{tabular}{lcc|cc}
    \hline
    \multicolumn{5}{c}{\bfseries Variant classification performance on the 2024 TopCoW internal test data}\\
    \hline
     & \multicolumn{2}{c|}{\bfseries CTA (n=70)}& \multicolumn{2}{c}{\bfseries MRA (n=70)}\\
    \cline{2-3} \cline{4-5}
    \bfseries Team & \makecell[l]{\bfseries Anterior\\\bfseries VarBalAcc (\%) $\uparrow$} & \makecell[l]{\bfseries Posterior\\\bfseries VarBalAcc (\%) $\uparrow$} & \makecell[l]{\bfseries Anterior\\\bfseries VarBalAcc (\%) $\uparrow$} & \makecell[l]{\bfseries Posterior\\\bfseries VarBalAcc (\%) $\uparrow$} \\
    \hline
    ARG\# & $24.66$ & $17.79$ & $25.52$ & $19.31$ \\
    CLAIM* & \cellcolor{silver!20}$73.94$ & \cellcolor{goldenyellow!60}$88.76$ & \cellcolor{goldenyellow!60}$89.14$ & \cellcolor{goldenyellow!60}$74.60$ \\
    IMR* & \cellcolor{goldenyellow!60}$87.28$ & \cellcolor{silver!20}$62.17$ & \cellcolor{silver!20}$86.46$ & \cellcolor{silver!20}$69.24$ \\
    junqiangchen\# & $25.00$ & $19.05$ & $25.00$ & $21.12$ \\
    pamaad\# & $25.00$ & $16.14$ & $25.00$ & $20.12$ \\
    UZH\# & \cellcolor{bronze!30}$36.26$ & \cellcolor{bronze!30}$39.75$ & \cellcolor{bronze!30}$37.28$ & \cellcolor{bronze!30}$22.47$ \\
    \hline
    \end{tabular}
\end{table*}

%% file: tables/table_2024_task_1_seg_results_ct_mr_external.tex
\begin{table*}[!htbp]
\caption{Results (mean $\pm$ standard deviation) of the CoW multiclass segmentation on the external test data for the top 6 teams in terms of class-average Dice similarity coefficient, centerline Dice (clDice) on merged binary mask, class-average 0-th Betti number ($\beta_0$) error, class-average Hausdorff Distance 95th Percentile (HD95), average F1 score for detection of communicating arteries and 3rd-A2, variant-balanced accuracy (VarBalAcc) of graph classification,
and CoW variant topology match rate (TMR) for both the anterior and posterior variants. The arrow indicates the favorable direction. The top three values for each metric are marked in gold, silver, and bronze colors. The bottom two rows for each dataset are the results for the top-3 teams ensemble and the organizer teams (`UZH', `DKFZ', `HITSZ') ensemble.
Team `CLAIM' used additional training data which included some external MRA test images from Lausanne and IXI-HH without our ground truth labels.}
\label{table:task_1_seg_results_cta_mra_external_2024}
\centering
\resizebox{1\textwidth}{!}{
    \begin{tabular}{llllllcccc}
        \hline
        \multicolumn{10}{c}{\bfseries ISLES CTA multiclass segmentation performance (n=26)}\\
        \hline
        \bfseries Team & \makecell[l]{\bfseries Per case\\\bfseries class-avg\\\bfseries Dice (\%) $\uparrow$} & \bfseries clDice (\%) $\uparrow$ & \makecell[l]{\bfseries Per case\\\bfseries class-avg\\\bfseries $\beta_0$ error $\downarrow$} &  \makecell[l]{\bfseries Per case\\\bfseries class-avg\\\bfseries HD95 $\downarrow$} & \makecell[l]{\bfseries Average F1\\\bfseries score (\%) $\uparrow$} & \makecell[l]{\bfseries Anterior\\\bfseries VarBalAcc (\%) $\uparrow$} & \makecell[l]{\bfseries Posterior\\\bfseries VarBalAcc (\%) $\uparrow$} & \makecell[l]{\bfseries Anterior\\\bfseries TMR (\%) $\uparrow$} & \makecell[l]{\bfseries Posterior\\\bfseries TMR (\%) $\uparrow$}\\
        \hline
        CLAIM & $85.23\pm4.64$ & \cellcolor{bronze!30}$98.69\pm1.65$ & \cellcolor{bronze!30}$0.07\pm0.09$ & \cellcolor{bronze!30}$2.94\pm3.62$ & \cellcolor{silver!20}$93.23\pm9.56$ & $80.67$ & $84.17$ & $21.85$ & \cellcolor{silver!20}$76.11$ \\
        DKFZ & \cellcolor{bronze!30}$89.00\pm5.48$ & \cellcolor{goldenyellow!60}$98.94\pm0.97$ & $0.09\pm0.09$ & $3.16\pm4.47$ & $91.65\pm6.85$ & $78.57$ & \cellcolor{silver!20}$88.33$ & $47.69$ & $59.72$ \\
        HITSZ & $87.72\pm5.46$ & $98.24\pm2.32$ & $0.11\pm0.09$ & $2.97\pm4.27$ & \cellcolor{bronze!30}$92.12\pm6.11$ & $84.24$ & $86.11$ & \cellcolor{bronze!30}$51.89$ & $54.72$ \\
        IMR & $88.39\pm4.93$ & $98.66\pm1.20$ & $0.09\pm0.11$ & $3.26\pm3.83$ & $90.50\pm5.51$ & \cellcolor{bronze!30}$88.45$ & $84.72$ & $35.50$ & $71.39$ \\
        NIC-VICOROB & \cellcolor{goldenyellow!60}$90.72\pm4.78$ & $98.65\pm1.57$ & \cellcolor{silver!20}$0.04\pm0.05$ & \cellcolor{silver!20}$2.55\pm4.00$ & $68.01\pm39.39$ & \cellcolor{goldenyellow!60}$92.02$ & \cellcolor{goldenyellow!60}$89.72$ & \cellcolor{goldenyellow!60}$65.55$ & \cellcolor{goldenyellow!60}$80.83$ \\
        UZH & \cellcolor{silver!20}$89.98\pm4.89$ & \cellcolor{silver!20}$98.70\pm1.27$ & \cellcolor{goldenyellow!60}$0.04\pm0.06$ & \cellcolor{goldenyellow!60}$2.53\pm3.33$ & \cellcolor{goldenyellow!60}$94.76\pm5.89$ & \cellcolor{silver!20}$89.08$ & \cellcolor{bronze!30}$87.78$ & \cellcolor{silver!20}$62.61$ & \cellcolor{bronze!30}$73.33$ \\
        \hline
        Top-3 ensemble & $91.51\pm4.46$ & $99.04\pm0.99$ & $0.06\pm0.08$ & $2.13\pm3.15$ & $94.17\pm6.82$ & $88.45$ & $91.11$ & $60.50$ & $78.89$ \\
        Orgs ensemble & $89.97\pm4.55$ & $98.96\pm0.98$ & $0.08\pm0.08$ & $2.53\pm3.55$ & $93.07\pm6.32$ & $82.14$ & $89.72$ & $51.26$ & $61.11$ \\
        \hline
        \multicolumn{10}{c}{}\\
        \multicolumn{10}{c}{\bfseries LargeIA CTA multiclass segmentation performance (n=20)}\\
        \hline
        \bfseries Team & \makecell[l]{\bfseries Per case\\\bfseries class-avg\\\bfseries Dice (\%) $\uparrow$} & \bfseries clDice (\%) $\uparrow$ & \makecell[l]{\bfseries Per case\\\bfseries class-avg\\\bfseries $\beta_0$ error $\downarrow$} &  \makecell[l]{\bfseries Per case\\\bfseries class-avg\\\bfseries HD95 $\downarrow$} & \makecell[l]{\bfseries Average F1\\\bfseries score (\%) $\uparrow$} & \makecell[l]{\bfseries Anterior\\\bfseries VarBalAcc (\%) $\uparrow$} & \makecell[l]{\bfseries Posterior\\\bfseries VarBalAcc (\%) $\uparrow$} & \makecell[l]{\bfseries Anterior\\\bfseries TMR (\%) $\uparrow$} & \makecell[l]{\bfseries Posterior\\\bfseries TMR (\%) $\uparrow$}\\
        \hline
        CLAIM & $81.88\pm6.44$ & \cellcolor{goldenyellow!60}$98.79\pm1.39$ & \cellcolor{bronze!30}$0.11\pm0.13$ & \cellcolor{bronze!30}$4.21\pm4.14$ & $73.68\pm16.17$ & \cellcolor{goldenyellow!60}$89.29$ & \cellcolor{goldenyellow!60}$63.73$ & $33.33$ & \cellcolor{goldenyellow!60}$49.84$ \\
        DKFZ & \cellcolor{bronze!30}$85.56\pm5.71$ & $98.39\pm1.97$ & $0.17\pm0.18$ & $5.31\pm5.11$ & \cellcolor{bronze!30}$74.51\pm9.88$ & $58.33$ & $45.40$ & $25.00$ & $23.17$ \\
        HITSZ & $84.82\pm7.57$ & \cellcolor{bronze!30}$98.50\pm1.52$ & $0.13\pm0.20$ & $4.53\pm5.69$ & $71.72\pm18.91$ & $61.90$ & \cellcolor{bronze!30}$53.02$ & \cellcolor{silver!20}$48.81$ & $30.08$ \\
        IMR & $84.43\pm6.86$ & $97.67\pm2.55$ & $0.15\pm0.12$ & $6.42\pm5.50$ & $71.57\pm17.85$ & $67.86$ & $38.02$ & $16.67$ & \cellcolor{bronze!30}$32.46$ \\
        NIC-VICOROB & \cellcolor{goldenyellow!60}$88.17\pm5.70$ & \cellcolor{silver!20}$98.55\pm1.62$ & \cellcolor{silver!20}$0.04\pm0.05$ & \cellcolor{goldenyellow!60}$3.38\pm4.48$ & \cellcolor{goldenyellow!60}$82.25\pm10.57$ & \cellcolor{bronze!30}$72.62$ & \cellcolor{silver!20}$57.78$ & \cellcolor{bronze!30}$46.43$ & \cellcolor{silver!20}$49.44$ \\
        UZH & \cellcolor{silver!20}$86.14\pm6.76$ & $97.75\pm1.97$ & \cellcolor{goldenyellow!60}$0.03\pm0.05$ & \cellcolor{silver!20}$3.59\pm4.57$ & \cellcolor{silver!20}$77.84\pm17.02$ & \cellcolor{silver!20}$78.57$ & $44.68$ & \cellcolor{goldenyellow!60}$71.43$ & $24.13$ \\
        \hline
        Top-3 ensemble & $87.95\pm6.32$ & $98.48\pm1.62$ & $0.09\pm0.11$ & $4.49\pm5.18$ & $76.89\pm16.89$ & $64.29$ & $47.06$ & $44.05$ & $43.73$ \\
        Orgs ensemble & $87.27\pm6.63$ & $98.36\pm1.66$ & $0.12\pm0.15$ & $4.14\pm5.14$ & $75.13\pm15.66$ & $64.29$ & $55.40$ & $46.43$ & $32.06$ \\
        \hline
        \multicolumn{10}{c}{}\\
        \multicolumn{10}{c}{\bfseries Lausanne MRA multiclass segmentation performance (n=20)}\\
        \hline
        \bfseries Team & \makecell[l]{\bfseries Per case\\\bfseries class-avg\\\bfseries Dice (\%) $\uparrow$} & \bfseries clDice (\%) $\uparrow$ & \makecell[l]{\bfseries Per case\\\bfseries class-avg\\\bfseries $\beta_0$ error $\downarrow$} &  \makecell[l]{\bfseries Per case\\\bfseries class-avg\\\bfseries HD95 $\downarrow$} & \makecell[l]{\bfseries Average F1\\\bfseries score (\%) $\uparrow$} & \makecell[l]{\bfseries Anterior\\\bfseries VarBalAcc (\%) $\uparrow$} & \makecell[l]{\bfseries Posterior\\\bfseries VarBalAcc (\%) $\uparrow$} & \makecell[l]{\bfseries Anterior\\\bfseries TMR (\%) $\uparrow$} & \makecell[l]{\bfseries Posterior\\\bfseries TMR (\%) $\uparrow$}\\
        \hline
        ARG & $87.98\pm5.83$ & \cellcolor{goldenyellow!60}$99.02\pm1.09$ & $0.10\pm0.13$ & $3.69\pm4.63$ & $83.23\pm9.59$ & $67.86$ & $72.22$ & \cellcolor{bronze!30}$46.43$ & $46.83$ \\
        CLAIM & $88.38\pm5.33$ & \cellcolor{bronze!30}$98.89\pm1.40$ & \cellcolor{goldenyellow!60}$0.03\pm0.05$ & \cellcolor{goldenyellow!60}$2.16\pm4.21$ & \cellcolor{goldenyellow!60}$96.46\pm4.55$ & \cellcolor{goldenyellow!60}$91.67$ & \cellcolor{bronze!30}$85.56$ & \cellcolor{goldenyellow!60}$60.71$ & \cellcolor{goldenyellow!60}$83.17$ \\
        DKFZ & \cellcolor{bronze!30}$88.77\pm6.22$ & \cellcolor{silver!20}$98.98\pm1.16$ & $0.09\pm0.11$ & \cellcolor{bronze!30}$3.32\pm5.26$ & $82.54\pm10.63$ & $72.62$ & $68.89$ & $29.76$ & $45.87$ \\
        HITSZ & $88.24\pm5.78$ & $98.79\pm1.40$ & $0.12\pm0.12$ & $3.62\pm4.60$ & $83.20\pm11.04$ & $76.19$ & $68.89$ & $35.71$ & $43.49$ \\
        IMR & \cellcolor{silver!20}$89.26\pm5.48$ & $98.64\pm1.76$ & \cellcolor{bronze!30}$0.08\pm0.11$ & \cellcolor{silver!20}$3.22\pm4.51$ & \cellcolor{bronze!30}$89.15\pm7.19$ & \cellcolor{silver!20}$90.48$ & \cellcolor{silver!20}$89.68$ & \cellcolor{silver!20}$47.62$ & \cellcolor{silver!20}$73.81$ \\
        UZH & \cellcolor{goldenyellow!60}$89.69\pm5.62$ & $98.71\pm1.42$ & \cellcolor{silver!20}$0.03\pm0.05$ & $3.33\pm4.63$ & \cellcolor{silver!20}$90.91\pm5.41$ & \cellcolor{bronze!30}$84.52$ & \cellcolor{goldenyellow!60}$91.90$ & $41.67$ & \cellcolor{bronze!30}$65.71$ \\
        \hline
        Top-3 ensemble & $90.69\pm5.51$ & $98.98\pm1.20$ & $0.06\pm0.10$ & $2.12\pm4.19$ & $93.08\pm6.08$ & $89.29$ & $91.11$ & $63.10$ & $65.71$ \\
        Orgs ensemble & $89.98\pm5.97$ & $98.84\pm1.24$ & $0.08\pm0.09$ & $2.90\pm5.15$ & $91.70\pm6.39$ & $72.62$ & $74.44$ & $29.76$ & $43.49$ \\
        \hline
        \multicolumn{10}{c}{}\\
        \multicolumn{10}{c}{\bfseries IXI-HH MRA multiclass segmentation performance (n=20)}\\
        \hline
        \bfseries Team & \makecell[l]{\bfseries Per case\\\bfseries class-avg\\\bfseries Dice (\%) $\uparrow$} & \bfseries clDice (\%) $\uparrow$ & \makecell[l]{\bfseries Per case\\\bfseries class-avg\\\bfseries $\beta_0$ error $\downarrow$} &  \makecell[l]{\bfseries Per case\\\bfseries class-avg\\\bfseries HD95 $\downarrow$} & \makecell[l]{\bfseries Average F1\\\bfseries score (\%) $\uparrow$} & \makecell[l]{\bfseries Anterior\\\bfseries VarBalAcc (\%) $\uparrow$} & \makecell[l]{\bfseries Posterior\\\bfseries VarBalAcc (\%) $\uparrow$} & \makecell[l]{\bfseries Anterior\\\bfseries TMR (\%) $\uparrow$} & \makecell[l]{\bfseries Posterior\\\bfseries TMR (\%) $\uparrow$}\\
        \hline
        ARG & $88.26\pm4.91$ & $98.10\pm1.74$ & $0.12\pm0.09$ & \cellcolor{bronze!30}$3.39\pm5.22$ & \cellcolor{silver!20}$90.29\pm7.38$ & \cellcolor{goldenyellow!60}$75.00$ & \cellcolor{bronze!30}$85.71$ & \cellcolor{bronze!30}$34.38$ & $30.71$ \\
        CLAIM & $87.62\pm5.41$ & \cellcolor{bronze!30}$98.20\pm1.63$ & \cellcolor{silver!20}$0.06\pm0.12$ & $3.67\pm6.16$ & \cellcolor{goldenyellow!60}$91.47\pm7.56$ & \cellcolor{goldenyellow!60}$75.00$ & $65.71$ & \cellcolor{silver!20}$37.50$ & \cellcolor{goldenyellow!60}$63.21$ \\
        DKFZ & \cellcolor{silver!20}$89.20\pm5.49$ & \cellcolor{silver!20}$98.38\pm1.89$ & \cellcolor{bronze!30}$0.08\pm0.12$ & \cellcolor{goldenyellow!60}$2.64\pm4.26$ & $83.53\pm10.81$ & \cellcolor{bronze!30}$62.50$ & \cellcolor{goldenyellow!60}$94.29$ & $21.88$ & $41.79$ \\
        HITSZ & \cellcolor{bronze!30}$88.45\pm5.41$ & \cellcolor{goldenyellow!60}$98.52\pm1.60$ & $0.11\pm0.13$ & $3.41\pm5.22$ & $87.89\pm9.96$ & \cellcolor{goldenyellow!60}$75.00$ & $58.57$ & $21.88$ & $36.07$ \\
        IMR & $75.78\pm23.29$ & $90.17\pm22.14$ & $0.19\pm0.24$ & $10.41\pm19.77$ & $77.55\pm17.15$ & \cellcolor{silver!20}$68.75$ & $79.29$ & \cellcolor{goldenyellow!60}$39.06$ & \cellcolor{bronze!30}$49.29$ \\
        UZH & \cellcolor{goldenyellow!60}$89.95\pm3.87$ & $98.11\pm2.18$ & \cellcolor{goldenyellow!60}$0.03\pm0.04$ & \cellcolor{silver!20}$3.18\pm3.77$ & \cellcolor{bronze!30}$89.72\pm7.09$ & \cellcolor{silver!20}$68.75$ & \cellcolor{silver!20}$87.14$ & $29.69$ & \cellcolor{silver!20}$54.64$ \\
        \hline
        Top-3 ensemble & $90.47\pm5.47$ & $98.70\pm1.53$ & $0.07\pm0.11$ & $2.49\pm4.96$ & $90.88\pm6.99$ & $75.00$ & $71.43$ & $34.38$ & $48.93$ \\
        Orgs ensemble & $90.44\pm5.40$ & $98.58\pm1.67$ & $0.08\pm0.13$ & $2.22\pm4.03$ & $91.87\pm6.64$ & $75.00$ & $94.29$ & $21.88$ & $51.79$ \\
        \hline
    \end{tabular}
}
\end{table*}

%% file: tables/table_2024_task_2_det_results_ct_mr_external.tex
\begin{table*}[!htbp]
\caption{Results (mean $\pm$ standard deviation) of the CoW ROI detection task on the external test set for the top 4 teams in terms of intersection over union (IoU) and Boundary IoU. The top three values for each metric are marked in gold, silver and bronze colors.}
\label{table:task_2_det_results_cta_mra_external_2024}
\centering
    \begin{tabular}{lll|ll|ll|ll}
    \hline
    \multicolumn{9}{c}{\bfseries Detection on external test data}\\
    \hline
     & \multicolumn{2}{c}{\bfseries ISLES CTA (n=26)}& \multicolumn{2}{c|}{\bfseries LargeIA CTA (n=20)}& \multicolumn{2}{c}{\bfseries Lausanne MRA (n=20)}& \multicolumn{2}{c}{\bfseries IXI-HH MRA (n=20)}\\
    \cline{2-3} \cline{4-5} \cline{6-7} \cline{8-9}
    \bfseries Team & \bfseries IoU (\%) $\uparrow$ & \makecell[l]{\bfseries Boundary\\\bfseries IoU (\%) $\uparrow$} & \bfseries IoU (\%) $\uparrow$ & \makecell[l]{\bfseries Boundary\\\bfseries IoU (\%) $\uparrow$} & \bfseries IoU (\%) $\uparrow$ & \makecell[l]{\bfseries Boundary\\\bfseries IoU (\%) $\uparrow$} & \bfseries IoU (\%) $\uparrow$ & \makecell[l]{\bfseries Boundary\\\bfseries IoU (\%) $\uparrow$} \\
    \hline
    ARG & \cellcolor{bronze!30}$79.07\pm6.74$ & \cellcolor{bronze!30}$68.79\pm9.36$ & \cellcolor{bronze!30}$76.67\pm6.87$ & \cellcolor{bronze!30}$65.80\pm9.23$ & \cellcolor{bronze!30}$77.71\pm6.39$ & \cellcolor{bronze!30}$65.87\pm9.14$ & \cellcolor{bronze!30}$77.23\pm10.06$ & \cellcolor{bronze!30}$66.03\pm13.72$ \\
    IMR & $62.51\pm26.70$ & $52.60\pm25.33$ & $64.30\pm27.74$ & $55.43\pm25.86$ & $76.85\pm6.82$ & $64.68\pm9.84$ & $73.06\pm19.87$ & $61.91\pm20.09$ \\
    NIC-VICOROB & \cellcolor{silver!20}$80.27\pm6.54$ & \cellcolor{silver!20}$70.58\pm9.03$ & \cellcolor{silver!20}$80.74\pm6.86$ & \cellcolor{silver!20}$71.46\pm9.55$ & \cellcolor{silver!20}$80.08\pm6.60$ & \cellcolor{silver!20}$69.93\pm9.56$ & \cellcolor{silver!20}$83.59\pm7.85$ & \cellcolor{silver!20}$74.83\pm11.02$ \\
    UZH & \cellcolor{goldenyellow!60}$85.85\pm7.09$ & \cellcolor{goldenyellow!60}$78.33\pm10.31$ & \cellcolor{goldenyellow!60}$89.83\pm5.21$ & \cellcolor{goldenyellow!60}$84.33\pm7.58$ & \cellcolor{goldenyellow!60}$89.70\pm6.79$ & \cellcolor{goldenyellow!60}$84.38\pm9.53$ & \cellcolor{goldenyellow!60}$93.14\pm5.66$ & \cellcolor{goldenyellow!60}$89.48\pm8.42$ \\
    \hline
    \end{tabular}
\end{table*}

%% file: tables/table_2024_task_3_edg_results_ct_mr_external.tex
\begin{table*}[!htbp]
\caption{Results of the CoW variant graph classification task on the external test set for the top 4 teams in terms of variant-balanced accuracy (VarBalAcc) for both the anterior and the posterior variant. The top three values for each metric are marked as gold, silver and bronze colors. Teams marked with * were segmentation-based and \# were classification-based submissions.
Team `CLAIM' used additional training data which included some external MRA test images from Lausanne and IXI-HH without our ground truth labels.}
\label{table:task_3_edg_results_cta_mra_external_2024}
\centering
\resizebox{1\textwidth}{!}{
    \begin{tabular}{lcc|cc|cc|cc}
    \hline
    \multicolumn{9}{c}{\bfseries Variant graph classification on external data}\\
    \hline
     & \multicolumn{2}{c}{\bfseries ISLES CTA (n=26)}& \multicolumn{2}{c|}{\bfseries LargeIA CTA (n=20)}& \multicolumn{2}{c}{\bfseries Lausanne MRA (n=20)}& \multicolumn{2}{c}{\bfseries IXI-HH MRA (n=20)}\\
    \cline{2-3} \cline{4-5} \cline{6-7} \cline{8-9}
    \bfseries Team & \makecell[l]{\bfseries Anterior\\\bfseries VarBalAcc (\%) $\uparrow$} & \makecell[l]{\bfseries Posterior\\\bfseries VarBalAcc (\%) $\uparrow$} & \makecell[l]{\bfseries Anterior\\\bfseries VarBalAcc (\%) $\uparrow$} & \makecell[l]{\bfseries Posterior\\\bfseries VarBalAcc (\%) $\uparrow$} & \makecell[l]{\bfseries Anterior\\\bfseries VarBalAcc (\%) $\uparrow$} & \makecell[l]{\bfseries Posterior\\\bfseries VarBalAcc (\%) $\uparrow$} & \makecell[l]{\bfseries Anterior\\\bfseries VarBalAcc (\%) $\uparrow$} & \makecell[l]{\bfseries Posterior\\\bfseries VarBalAcc (\%) $\uparrow$} \\
    \hline
    CLAIM* & \cellcolor{silver!20}$80.67$ & \cellcolor{silver!20}$84.17$ & \cellcolor{goldenyellow!60}$89.29$ & \cellcolor{goldenyellow!60}$63.73$ & \cellcolor{goldenyellow!60}$91.67$ & \cellcolor{silver!20}$85.56$ & \cellcolor{goldenyellow!60}$75.00$ &\cellcolor{silver!20}$65.71$ \\
    IMR* & \cellcolor{goldenyellow!60}$86.97$ & \cellcolor{goldenyellow!60}$84.72$ & \cellcolor{silver!20}$67.86$ & \cellcolor{silver!20}$34.68$ & \cellcolor{silver!20}$88.10$ & \cellcolor{goldenyellow!60}$89.68$ & \cellcolor{silver!20}$68.75$ &\cellcolor{goldenyellow!60}$81.79$ \\
    junqiangchen\# & $25.00$ & \cellcolor{bronze!30}$14.44$ & \cellcolor{bronze!30}$33.33$ & $17.62$ & \cellcolor{bronze!30}$33.33$ & $9.05$ & $25.00$ &\cellcolor{bronze!30}$18.21$ \\
    UZH\# & \cellcolor{bronze!30}$27.52$ & $13.06$ & $30.95$ & \cellcolor{bronze!30}$22.62$ & $30.95$ & \cellcolor{bronze!30}$24.44$ & \cellcolor{bronze!30}$34.38$ &$8.21$ \\
    \hline
    \end{tabular}
}
\end{table*}

%% file: sections/suppl/infertime.tex
\section{Inference Time for Best Segmentation Algorithms}
\label{sec:infer_time_best_seg_algo}

Table~\ref{table:task_1_inf_time}
shows the inference time in seconds by the top segmentation algorithms.

The top algorithms were selected for further benchmarking on external test sets.
During this process, we performed a runtime analysis and measured their inference time using a laptop with GPU.
The runtime depended on the dataset, but none of the datasets took more than two and a half minutes on average on a test image for prediction.
Teams `UZH' and `NIC-VICOROB' had slightly longer inference time of around 2 minutes per test image.
The other five teams had average inference time ranging from 35 seconds to 69 seconds per image.

\input{tables/table_2024_task_1_inference_time_external}

%% file: tables/table_2024_task_1_inference_time_external.tex

\begin{table*}[!htbp]
\caption{Per case inference time in seconds (mean $\pm$ standard deviation) for the CoW multiclass segmentation task on the external test set. The times are reported for the top 6 teams for each track separately. For teams outside the top 6 in a given track, inference times are not included and are indicated with a `-'.}
\label{table:task_1_inf_time}
\centering
    \begin{tabular}{l|ll|ll}
    \hline
    \multicolumn{5}{c}{\bfseries Per case inference time of segmentation algorithms (seconds)}\\
    \hline
    \bfseries Team & \bfseries ISLES CTA & \bfseries LargeIA CTA & \bfseries Lausanne MRA & \bfseries IXI-HH MRA \\
    \hline
    ARG & - & - & $105.0 \pm 17.0$ & $65.5 \pm 5.1$ \\
    CLAIM & $36.6 \pm 4.8$ & $39.9 \pm 3.1$ & $34.3 \pm 3.9$ & $30.0 \pm 0.7$ \\
    DKFZ & $70.7 \pm 11.4$ & $75.5 \pm 9.6$ & $77.1 \pm 14.1$ & $44.1 \pm 3.5$ \\
    HITSZ & $65.5 \pm 18.0$ & $76.5 \pm 11.8$ & $80.6 \pm 14.1$ & $56.9 \pm 5.8$ \\
    IMR & $51.2 \pm 9.0$ & $62.1 \pm 8.8$ & $56.3 \pm 15.0$ & $32.2 \pm 1.4$ \\
    NIC-VICOROB & $123.9 \pm 21.8$ & $129.0 \pm 20.7$ & - & - \\
    UZH & $145.2 \pm 25.1$ & $139.6 \pm 20.8$ & $147.4 \pm 19.9$ & $93.5 \pm 10.2$ \\
    \hline
    \end{tabular}
\end{table*}

%% file: sections/suppl/aneu.tex
\section{Detail Results on Locating Aneurysm}
\label{sec:details_on_aneu}

Table~\ref{table:aneurysm_cases_largeia} lists in detail the selected 12 patients with intracranial aneurysms from the LargeIA CTA dataset,
their aneurysm locations,
and the locations in terms of CoW labels as predicted by the top teams.

\begin{figure*}[p]
    \centering
    \includegraphics[width=0.69\textwidth]{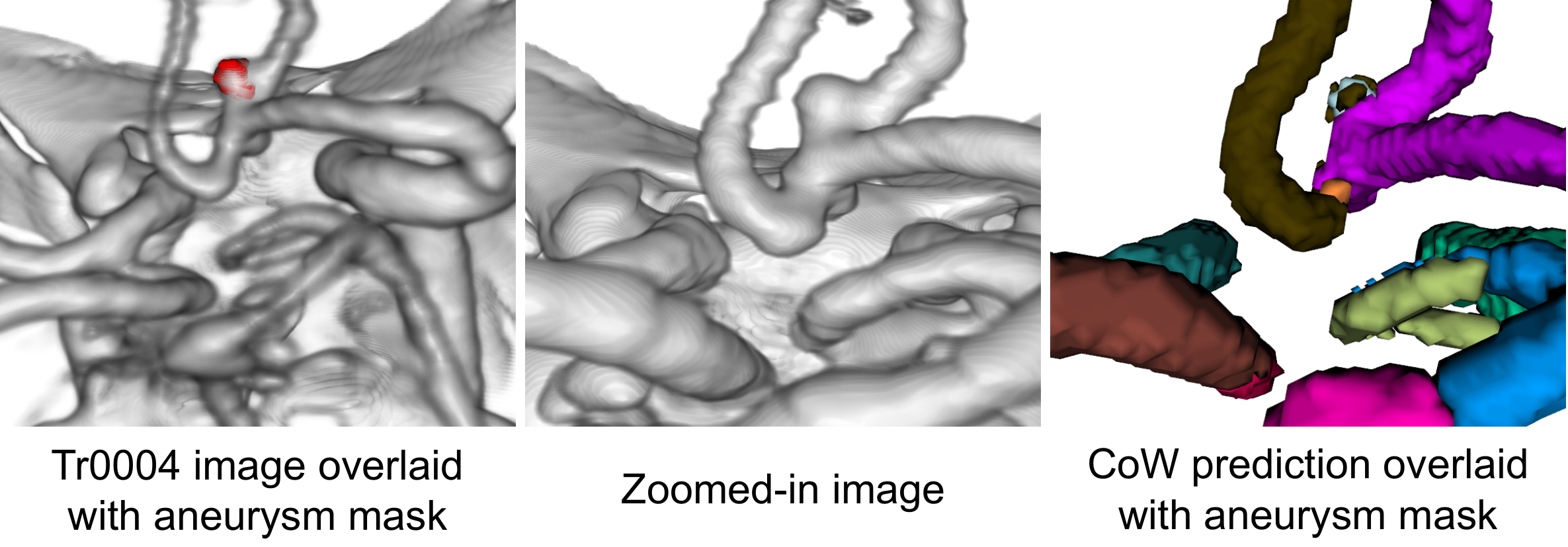}
    \caption{
        Tr0004 contains a common mistake for locating the aneurysm due to the tortuous ACAs and the presence of another infundibulum near the Acom-complex.
        Left: Image overlaid with the ground-truth aneurysm mask in red;
        Middle: Zoomed-in view of the image;
        Right: Aneurysm mask in silver color overlaid with predicted CoW masks from team `CLAIM'.
    }
    \label{fig:aneurysm_frequent_wrong}
\end{figure*}

Fig.~\ref{fig:aneurysm_frequent_wrong} shows the aneurysm and CoW vessels of the patient Tr0004 for which most top teams had an error in locating this aneurysm.

\input{tables/table_2024_aneurysm_cases_largeia}

%% file: tables/table_2024_aneurysm_cases_largeia.tex

\begin{table*}[!htbp]
\caption{Aneurysm locations given by the predicted CoW labels that the aneurysms overlap with or are adjacent to in the overlay of the masks.
The locations in CoW predictions were evaluated for the top teams `UZH', `NIC-VICOROB', `IMR', and `CLAIM' for 12 cases from the LargeIA CTA dataset containing at least one aneurysm inside or near the CoW ROI.
Wrong locations in predictions are marked in magenta color.
}
\label{table:aneurysm_cases_largeia}
\centering
    \begin{tabular}{l|l|l|l|l|l}
    \hline
    \multicolumn{5}{c}{\bfseries Aneurysm location in CoW predictions for LargeIA CTA}\\
    \hline
    \multirow{2}{*}{\bfseries Image} & \multirow{2}{*}{\bfseries Aneurysm location} & \multicolumn{4}{c}{\bfseries Location in CoW prediction} \\
    \cline{3-6}
        &   & \bfseries UZH & \bfseries NIC-VICOROB & \bfseries IMR & \bfseries CLAIM \\
    \hline
    Tr0001 & \makecell[l]{R-ICA} & R-ICA & R-ICA & R-ICA & R-ICA \\
    \hline
    Tr0004 & \makecell[l]{R-A2 origin\\near Acom-complex} & R-ACA & R-ACA, \textcolor{magenta}{L-ACA} & R-ACA, \textcolor{magenta}{L-ACA} & R-ACA, \textcolor{magenta}{L-ACA} \\
    \hline
    Tr0005 & \makecell[l]{Acom-ACA junction\\R-ACA side} & Acom, R-ACA, L-ACA & Acom, R-ACA, L-ACA & Acom, R-ACA, L-ACA & R-ACA, L-ACA \\
    \hline
    Tr0006 & \makecell[l]{L-ICA-Pcom junction} & L-ICA, L-Pcom & L-ICA, L-Pcom & L-ICA, L-Pcom &  L-ICA, L-Pcom \\
    \hline
    Tr0015 & \makecell[l]{L-ICA} & L-ICA & L-ICA & L-ICA & L-ICA \\
    \hline
    Tr0018 & BA tip & BA, R-PCA & BA, R-PCA, L-PCA & BA, R-PCA, L-PCA & BA, R-PCA, L-PCA \\
    \hline
    Tr0019 & \makecell[l]{1. R-ICA bifurcation\\2. L-ICA} & \makecell[l]{1. R-ICA\\2. L-ICA} & \makecell[l]{1. R-ICA, R-ACA\\2. L-ICA} & \makecell[l]{1. R-ICA\\2. L-ICA, \textcolor{magenta}{L-Pcom}} & \makecell[l]{1. R-ICA\\2. L-ICA} \\
    \hline
    Tr0024 & \makecell[l]{Acom-ACA junction} & Acom, R-ACA, L-ACA & Acom, R-ACA, L-ACA & Acom, R-ACA, L-ACA & Acom, R-ACA, L-ACA \\
    \hline
    Tr0025 & \makecell[l]{1. BA trunk \\2. BA tip} & \makecell[l]{1. BA \\2. BA, L-PCA} & \makecell[l]{1. BA \\2. BA, L-PCA}  & \makecell[l]{1. BA \\2. BA, L-PCA} & \makecell[l]{1. BA \\2. BA, R-PCA, L-PCA} \\
    \hline
    Tr0038 & \makecell[l]{L-ICA-Pcom junction} & L-ICA, L-Pcom & L-ICA, L-Pcom & L-ICA, L-Pcom & L-ICA, L-Pcom \\
    \hline
    Tr0069 & \makecell[l]{L-MCA bifurcation} & L-MCA & L-MCA & L-MCA & L-MCA \\
    \hline
    Tr0089 & \makecell[l]{R-MCA bifurcation} & R-MCA & R-MCA & R-MCA & R-MCA \\
    \hline
    \end{tabular}
\end{table*}

%% file: sections/suppl/results23.tex
\section{Detailed Results of the 2023 Tasks}

\begin{figure*}[!ht]
    \centering
    \begin{minipage}{0.9\textwidth}
         \centering
         \includegraphics[width=\textwidth]{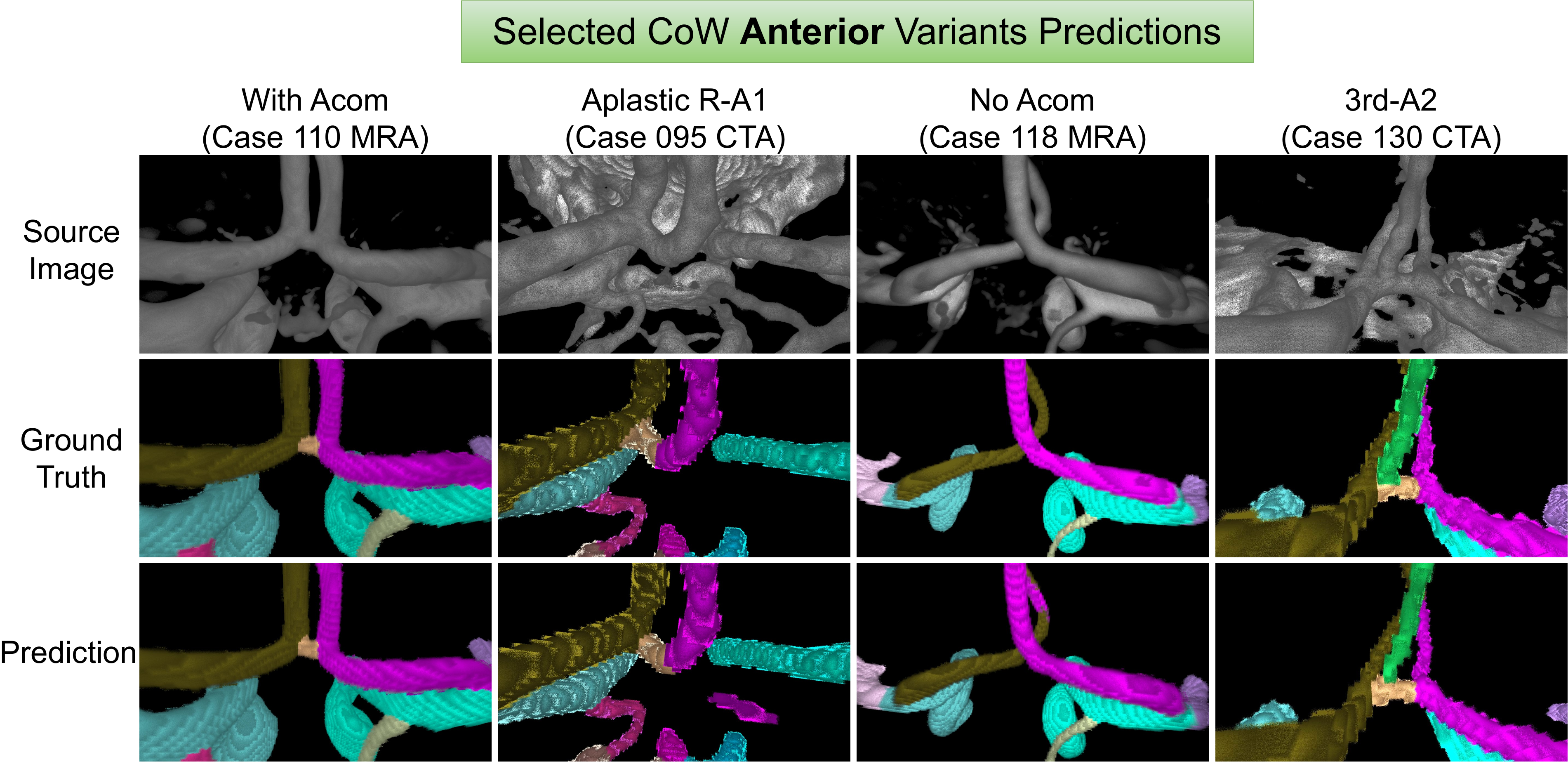}
    \end{minipage}
\par\bigskip 
    \begin{minipage}{0.9\textwidth}
         \centering
         \includegraphics[width=\textwidth]{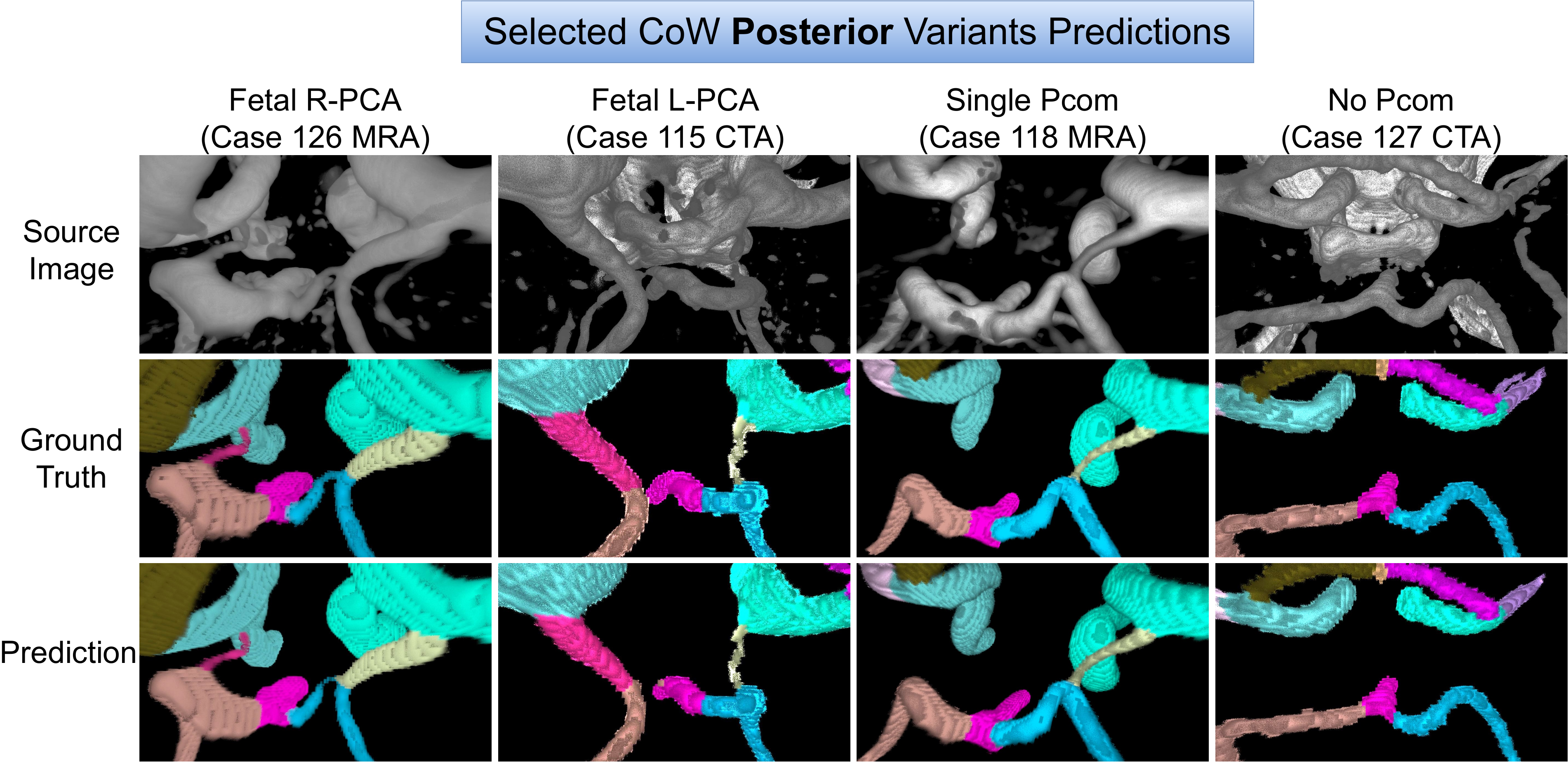}
    \end{minipage}
    \caption{More qualitative results for \textbf{anterior (top sub-figure)} and \textbf{posterior (bottom sub-figure)} variants in multiclass segmentation task.
    The predictions are by 2023 team `DKFZ/WilliWillsWissen'.
    Alternating columns showcasing MRA and CTA.
    }
    \label{fig:TopcowArxiv_PosteriorQualityFig}
\end{figure*}

Table~\ref{table:binary_seg_results} and Table~\ref{table:multiclass_seg_results_1} show the results for the 2023 binary segmentation task and the multiclass segmentation task.
Fig.~\ref{fig:TopcowArxiv_PosteriorQualityFig} shows the qualitative examples of the predictions from one of the 2023 winning teams for a few selected anterior (Fig.~\ref{fig:TopcowArxiv_PosteriorQualityFig} top) and posterior (Fig.~\ref{fig:TopcowArxiv_PosteriorQualityFig} bottom) variants.

\input{tables/table_2023_binary_seg_results.tex}

\input{tables/table_2023_multiclass_seg_results_leaderboard.tex}

%% file: tables/table_2023_binary_seg_results.tex

\begin{table*}[!htbp]
\caption{Binary segmentation task results (mean $\pm$ standard deviation) from 2023 for MRA and CTA in Dice scores, centerline Dice (clDice) scores and errors in the zero-th Betti number $\beta_0$. The arrow indicates the favorable direction. The top three values for each metric and each track are marked as gold, silver and bronze cells in decreasing order. A `*' behind the team name indicates that the segmentation predictions have been converted from the multiclass submissions and inserted here.  If a team only submitted to one of the tracks the columns of the other track are filled with a `-'.}
\label{table:binary_seg_results}
\centering
    \begin{tabular}{llll|lll}
    \hline
 \multicolumn{7}{c}{\bfseries 2023 binary segmentation performance}\\
    \hline
     & \multicolumn{3}{c}{\bfseries MRA (n=35)}& \multicolumn{3}{c}{\bfseries CTA (n=35)}\\
    \cline{2-4} \cline{5-7}
    \bfseries Team & \bfseries Dice (\%) $\uparrow$ & \bfseries clDice (\%) $\uparrow$ & \bfseries $\beta_0$ error $\downarrow$ & \bfseries Dice (\%) $\uparrow$ & \bfseries clDice (\%) $\uparrow$ & \bfseries $\beta_0$ error $\downarrow$ \\
    \hline
    2i\_mtl & - & - & - & $44.15\pm11.94$ & $48.04\pm11.23$ & $28.77\pm15.12$ \\
    agaldran & $94.33\pm2.64$ & $96.68\pm2.79$ & $1.57\pm1.29$ & $87.73\pm3.26$ & $95.99\pm2.97$ & $2.26\pm1.82$ \\
    DKFZ & \cellcolor{silver!20}$95.54\pm2.72$ & \cellcolor{goldenyellow!60}$98.31\pm2.14$ & \cellcolor{goldenyellow!60}$0.37\pm0.60$ & \cellcolor{bronze!30}$92.16\pm2.76$ & \cellcolor{goldenyellow!60}$98.42\pm1.83$ & \cellcolor{silver!20}$0.57\pm0.81$ \\
    EURECOM & $93.84\pm2.69$ & $94.42\pm3.10$ & $3.77\pm2.79$ & $84.79\pm4.34$ & $89.92\pm5.43$ & $8.14\pm4.66$ \\
    gbCoW & $94.95\pm2.89$ & $98.10\pm2.33$ & $0.86\pm0.94$ & - & - & - \\
    gl$^*$ & $93.67\pm4.84$ & $95.81\pm5.11$ & $0.71\pm0.86$ & $70.94\pm24.88$ & $75.80\pm28.05$ & $2.09\pm1.48$ \\
    HITSZ & $94.05\pm2.91$ & $97.29\pm2.29$ & $0.69\pm0.68$ & \cellcolor{silver!20}$92.28\pm2.83$ & $97.70\pm2.54$ & $0.77\pm0.97$ \\
    IWantToGoToCanada & - & - & - & $90.06\pm2.67$ & $96.56\pm2.65$ & $1.40\pm0.88$ \\
    junqiangchen & $94.09\pm2.08$ & $96.86\pm2.80$ & $3.20\pm2.96$ & $89.49\pm2.87$ & $95.91\pm1.63$ & $3.91\pm2.84$ \\
    lWM & $94.39\pm2.48$ & $97.39\pm2.76$ & $1.43\pm1.36$ & $91.07\pm2.50$ & $97.07\pm2.60$ & $1.74\pm1.27$ \\
    NIC-VICOROB & \cellcolor{goldenyellow!60}$95.60\pm2.32$ & \cellcolor{silver!20}$98.26\pm2.20$ & \cellcolor{bronze!30}$0.60\pm0.60$ & $92.07\pm3.54$ & \cellcolor{silver!20}$97.93\pm2.47$ & \cellcolor{bronze!30}$0.60\pm0.85$ \\
    NIC-VICOROB-2 & $93.13\pm3.67$ & $94.30\pm7.03$ & $17.69\pm57.63$ & $89.68\pm3.87$ & $95.45\pm3.28$ & $3.80\pm2.74$ \\
    refrain & $93.87\pm2.02$ & \cellcolor{bronze!30}$98.20\pm1.84$ & $0.97\pm1.20$ & - & - & - \\
    sjtu\_eiee\_2-426lab & - & - & - & \cellcolor{goldenyellow!60}$92.75\pm3.19$ & \cellcolor{bronze!30}$97.81\pm2.66$ & \cellcolor{goldenyellow!60}$0.40\pm0.55$ \\
    UB-VTL & $91.78\pm2.41$ & $93.27\pm3.50$ & $0.91\pm0.98$ & $72.27\pm6.89$ & $77.49\pm7.92$ & $2.37\pm1.72$ \\
    UW & \cellcolor{bronze!30}$95.37\pm2.20$ & $98.18\pm1.93$ & $0.89\pm0.87$ & - & - & - \\
    UZH$^{*}$ & $95.14\pm2.90$ & $98.06\pm2.30$ & \cellcolor{silver!20}$0.57\pm0.65$ & $90.25\pm5.73$ & $96.98\pm3.48$ & $1.14\pm1.44$ \\
    ysato & $88.05\pm4.94$ & $91.99\pm3.60$ & $2.60\pm1.77$ & - & - & - \\
    \hline
    \end{tabular}
\end{table*}

%% file: tables/table_2023_multiclass_seg_results_leaderboard.tex

\begin{table*}[!htbp]
\caption{Results (mean $\pm$ standard deviation) of the multiclass segmentation task from 2023 for MRA and CTA in terms of class-average Dice scores, centerline Dice (clDice) scores and class-average errors in the zero-th Betti number $\beta_0$. The \textit{clDice} scores were computed on the merged binary class, the \textit{Dice} scores and $\beta_0$ errors were computed for each class separately and the average was taken per case. The arrow indicates the favorable direction. The top three values for each metric and each track are marked as gold, silver and bronze cells in decreasing order. If a team only submitted to one of the tracks the columns of the other track are filled with a `-'.
}
\label{table:multiclass_seg_results_1}
\centering
    \begin{tabular}{llll|lll}
    \hline
 \multicolumn{7}{c}{\bfseries 2023 multiclass segmentation performance}\\
    \hline
     & \multicolumn{3}{c}{\bfseries MRA (n=35)}& \multicolumn{3}{c}{\bfseries CTA (n=35)}\\
    \cline{2-4} \cline{5-7}
    \bfseries Team & \makecell[l]{\bfseries Per case\\\bfseries class-average\\\bfseries Dice (\%) $\uparrow$} & \bfseries clDice (\%) $\uparrow$ & \makecell[l]{\bfseries Per case\\\bfseries class-average\\\bfseries $\beta_0$ error $\downarrow$} & \makecell[l]{\bfseries Per case\\\bfseries class-average\\\bfseries Dice (\%) $\uparrow$} & \bfseries clDice (\%) $\uparrow$ & \makecell[l]{\bfseries Per case\\\bfseries class-average\\\bfseries $\beta_0$ error $\downarrow$} \\
    \hline
    agaldran & $0.01\pm0.07$ & $0.16\pm0.69$ & $1.01\pm0.18$ & $0.15\pm0.66$ & $0.77\pm1.62$ & $1.07\pm0.32$ \\
    DKFZ & \cellcolor{goldenyellow!60}$84.58\pm6.47$ & $97.21\pm3.37$ & \cellcolor{goldenyellow!60}$0.06\pm0.06$ & \cellcolor{goldenyellow!60}$83.32\pm5.65$ & \cellcolor{goldenyellow!60}$98.12\pm1.86$ & \cellcolor{goldenyellow!60}$0.11\pm0.11$ \\
    gbCoW & $80.51\pm14.69$ & \cellcolor{silver!20}$98.10\pm2.33$ & $0.28\pm0.27$ & - & - & - \\
    gl & $81.27\pm9.16$ & $95.81\pm5.11$ & \cellcolor{silver!20}$0.09\pm0.08$ & $54.02\pm21.76$ & $75.80\pm28.05$ & $0.24\pm0.22$ \\
    HITSZ & \cellcolor{bronze!30}$83.76\pm6.95$ & $97.29\pm2.29$ & $0.14\pm0.14$ & \cellcolor{silver!20}$81.67\pm6.81$ & \cellcolor{bronze!30}$97.78\pm1.88$ & \cellcolor{silver!20}$0.16\pm0.14$ \\
    IWantToGoToCanada & - & - & - & $74.34\pm7.90$ & $97.61\pm1.78$ & $0.77\pm0.50$ \\
    junqiangchen & $71.56\pm9.72$ & $96.86\pm2.80$ & $0.87\pm0.41$ & $67.68\pm5.97$ & $95.91\pm1.63$ & $1.09\pm0.37$ \\
    lWM & $79.03\pm8.75$ & $96.23\pm3.11$ & $0.27\pm0.22$ & $72.49\pm8.46$ & $92.78\pm5.13$ & $0.32\pm0.19$ \\
    NIC-VICOROB & $77.13\pm20.15$ & $96.31\pm5.78$ & \cellcolor{bronze!30}$0.11\pm0.09$ & $51.84\pm31.11$ & \cellcolor{silver!20}$97.86\pm2.38$ & $0.43\pm0.38$ \\
    NIC-VICOROB-2 & $65.99\pm8.76$ & $94.12\pm6.12$ & $2.64\pm3.00$ & $62.41\pm8.22$ & $95.65\pm3.38$ & $2.31\pm1.37$ \\
    refrain & $83.72\pm5.79$ & \cellcolor{goldenyellow!60}$98.20\pm1.84$ & $0.19\pm0.17$ & - & - & - \\
    sjtu\_eiee\_2-426lab & - & - & - & $67.46\pm27.78$ & $97.15\pm3.22$ & \cellcolor{bronze!30}$0.23\pm0.27$ \\
    UZH & \cellcolor{silver!20}$83.98\pm7.33$ & \cellcolor{bronze!30}$98.06\pm2.30$ & $0.16\pm0.13$ & \cellcolor{bronze!30}$77.00\pm11.95$ & $96.98\pm3.48$ & \cellcolor{bronze!30}$0.23\pm0.26$ \\
    \hline
    \end{tabular}
\end{table*}

%% file: refs.bib
@book{osborn2013osborn,
  title={Osborn's Brain: Imaging, Pathology, and Anatomy},
  author={Osborn, Anne G.},
  isbn={9781931884211},
  lccn={2012031063},
  year={2013},
  publisher={Amirsys}
}

@article{chuang2013configuration,
  title={Configuration of the circle of Willis is associated with less symptomatic intracerebral hemorrhage in ischemic stroke patients treated with intravenous thrombolysis},
  author={Chuang, Yu-Ming and Chan, Lung and Lai, Yen-Jun and Kuo, Kuei-Hong and Chiou, Yih-Hwa and Huang, Lih-Wen and Kwok, Yam-Ting and Lai, Tzu-Hsien and Lee, Siu-Pak and Wu, Hung-Ming and others},
  journal={Journal of Critical Care},
  volume={28},
  number={2},
  pages={166--172},
  year={2013},
  publisher={Elsevier}
}

@article{van2015completeness,
  title={Completeness of the circle of Willis and risk of ischemic stroke in patients without cerebrovascular disease},
  author={{van Seeters}, Tom and Hendrikse, Jeroen and Biessels, Geert Jan and Velthuis, Birgitta K and Mali, Willem PTM and Kappelle, L Jaap and van der Graaf, Yolanda and SMART Study Group},
  journal={Neuroradiology},
  volume={57},
  pages={1247--1251},
  year={2015},
  publisher={Springer}
}

@article{rinaldo2016relationship,
  title={Relationship of A1 segment hypoplasia to anterior communicating artery aneurysm morphology and risk factors for aneurysm formation},
  author={Rinaldo, Lorenzo and McCutcheon, Brandon A and Murphy, Meghan E and Bydon, Mohamad and Rabinstein, Alejandro A and Lanzino, Giuseppe},
  journal={Journal of Neurosurgery},
  volume={127},
  number={1},
  pages={89--95},
  year={2016},
  publisher={American Association of Neurological Surgeons}
}

@article{banga2018incomplete,
  title={Incomplete circle of Willis is associated with a higher incidence of neurologic events during carotid eversion endarterectomy without shunting},
  author={Banga, P{\'e}ter Vince and Varga, Andrea and Csobay-Nov{\'a}k, Csaba and Kolossv{\'a}ry, M{\'a}rton and Sz{\'a}nt{\'o}, Emese and Oderich, Gustavo S and Entz, L{\'a}szl{\'o} and S{\'o}tonyi, P{\'e}ter},
  journal={Journal of Vascular Surgery},
  volume={68},
  number={6},
  pages={1764--1771},
  year={2018},
  publisher={Elsevier}
}

@article{yang2017relationship,
  title={Relationship of A1 segment hypoplasia with the radiologic and clinical outcomes of surgical clipping of anterior communicating artery aneurysms},
  author={Yang, Fan and Li, Hao and Wu, Jun and Li, Maogui and Chen, Xin and Jiang, Pengjun and Li, Zhengsong and Cao, Yong and Wang, Shuo},
  journal={World Neurosurgery},
  volume={106},
  pages={806--812},
  year={2017},
  publisher={Elsevier}
}

@article{krabbe1998circle,
  title={Circle of Willis: morphologic variation on three-dimensional time-of-flight MR angiograms.},
  author={Krabbe-Hartkamp, Monique J and Van der Grond, Jeroen and De Leeuw, FE and de Groot, Jan Cees and Algra, Ale and Hillen, Berend and Breteler, MM and Mali, WP},
  journal={Radiology},
  volume={207},
  number={1},
  pages={103--111},
  year={1998}
}

@article{iqbal2013comprehensive,
  title={A comprehensive study of the anatomical variations of the circle of willis in adult human brains},
  author={Iqbal, Sumaiya},
  journal={Journal of Clinical and Diagnostic Research: JCDR},
  volume={7},
  number={11},
  pages={2423},
  year={2013},
  publisher={JCDR Research \& Publications Private Limited}
}

@article{bullitt2005vessel,
  title={Vessel tortuosity and brain tumor malignancy: a blinded study},
  author={Bullitt, Elizabeth and Zeng, Donglin and Gerig, Guido and Aylward, Stephen and Joshi, Sarang and Smith, J Keith and Lin, Weili and Ewend, Matthew G},
  journal={Academic Radiology},
  volume={12},
  number={10},
  pages={1232--1240},
  year={2005},
  publisher={Elsevier}
}

@article{oktay2018AttentionUNet,
  title={Attention u-net: Learning where to look for the pancreas},
  author={Oktay, Ozan and Schlemper, Jo and Folgoc, Loic Le and Lee, Matthew and Heinrich, Mattias and Misawa, Kazunari and Mori, Kensaku and McDonagh, Steven and Hammerla, Nils Y and Kainz, Bernhard and others},
  journal={arXiv preprint arXiv:1804.03999},
  year={2018}
}

@article{galati2023A2V,
  title={A2V: A semi-supervised domain adaptation framework for brain vessel segmentation via two-phase training angiography-to-venography translation},
  author={Galati, Francesco and Falcetta, Daniele and Cortese, Rosa and Casolla, Barbara and Prados, Ferran and Burgos, Ninon and Zuluaga, Maria A},
  journal={arXiv preprint arXiv:2309.06075},
  year={2023}
}

@article{billot2023synthseg,
  title={Synthseg: Segmentation of brain MRI scans of any contrast and resolution without retraining},
  author={Billot, B. and Greve, D.N. and Puonti, O. and Thielscher, A. and Van Leemput, K. and Fischl, B. and Dalca, A.V. and Iglesias, J.E.},
  journal={Medical Image Analysis},
  volume={86},
  pages={102789},
  year={2023}
}

@article{isensee2021nnUNet,
  title={nnU-Net: a self-configuring method for deep learning-based biomedical image segmentation},
  author={Isensee, F. and Jaeger, P. F. and Kohl, S. A. and Petersen, J. and Maier-Hein, K. H.},
  journal={Nature Methods},
  volume={18},
  number={2},
  year={2021}
}

@article{roy2023mednext,
  title={Mednext: transformer-driven scaling of convnets for medical image segmentation},
  author={Roy, S. and Koehler, G. and Ulrich, C. and Baumgartner, M. and Petersen, J. and Isensee, F. and Jaeger, P. F. and Maier-Hein, K. H.},
  journal={International Conference on Medical Image Computing and Computer-Assisted Intervention (MICCAI)},
  year={2023},
  pages={405-415},
  publisher={Springer}
}

@article{lee2022uxnet,
  title={3d ux-net: A large kernel volumetric convnet modernizing hierarchical transformer for medical image segmentation},
  author={Lee, Ho Hin and Bao, Shunxing and Huo, Yuankai and Landman, Bennett A},
  journal={arXiv preprint arXiv:2209.15076},
  year={2022}
}

@inproceedings{hatamizadeh2021SwinUNETR,
  title={Swin UNETR: Swin Transformers for Semantic Segmentation of Brain Tumors in MRI Images},
  author={Hatamizadeh, Ali and Nath, Vishwesh and Tang, Yucheng and Yang, Dong and Roth, Holger R and Xu, Daguang},
  booktitle={International MICCAI BrainLesion Workshop},
  pages={272--284},
  year={2021},
  organization={Springer}
}

@inproceedings{milletari2016VNet,
  title={V-net: Fully convolutional neural networks for volumetric medical image segmentation},
  author={Milletari, Fausto and Navab, Nassir and Ahmadi, Seyed-Ahmad},
  booktitle={2016 fourth International Conference on 3D Vision (3DV)},
  pages={565--571},
  year={2016},
  organization={IEEE}
}

@article{hilbert2020bravenet,
  title={BRAVE-NET: fully automated arterial brain vessel segmentation in patients with cerebrovascular disease},
  author={Hilbert, Adam and Madai, Vince I and Akay, Ela M and Aydin, Orhun U and Behland, Jonas and Sobesky, Jan and Galinovic, Ivana and Khalil, Ahmed A and Taha, Abdel A and Wuerfel, Jens and others},
  journal={Frontiers in artificial intelligence},
  volume={3},
  pages={552258},
  year={2020},
  publisher={Frontiers Media SA}
}

@article{hilbert2022anatomical,
  title={Anatomical labeling of intracranial arteries with deep learning in patients with cerebrovascular disease},
  author={Hilbert, Adam and Rieger, Jana and Madai, Vince I and Akay, Ela M and Aydin, Orhun U and Behland, Jonas and Khalil, Ahmed A and Galinovic, Ivana and Sobesky, Jan and Fiebach, Jochen and others},
  journal={Frontiers in Neurology},
  volume={13},
  pages={1000914},
  year={2022},
  publisher={Frontiers Media SA}
}

@article{wasserthal2023totalsegmentator,
  title={Totalsegmentator: Robust segmentation of 104 anatomic structures in ct images},
  author={Wasserthal, Jakob and Breit, Hanns-Christian and Meyer, Manfred T and Pradella, Maurice and Hinck, Daniel and Sauter, Alexander W and Heye, Tobias and Boll, Daniel T and Cyriac, Joshy and Yang, Shan and others},
  journal={Radiology: Artificial Intelligence},
  volume={5},
  number={5},
  year={2023},
  publisher={Radiological Society of North America}
}

@inproceedings{quickshearbib,
author = {Schimke, Nakeisha and Hale, John},
title = {Quickshear Defacing for Neuroimages},
year = {2011},
publisher = {USENIX Association},
address = {USA},
booktitle = {Proceedings of the 2nd USENIX Conference on Health Security and Privacy},
pages = {11},
numpages = {1},
location = {San Francisco, CA},
series = {HealthSec'11}
}

@article{isensee2019hdbet,
  title={Automated brain extraction of multisequence MRI using artificial neural networks},
  author={Isensee, Fabian and Schell, Marianne and Pflueger, Irada and Brugnara, Gianluca and Bonekamp, David and Neuberger, Ulf and Wick, Antje and Schlemmer, Heinz-Peter and Heiland, Sabine and Wick, Wolfgang and others},
  journal={Human Brain Mapping},
  volume={40},
  number={17},
  pages={4952--4964},
  year={2019},
  publisher={Wiley Online Library}
}

@article{pidhorskyi2018syglass,
  title={sy{G}lass: Interactive exploration of multidimensional images using virtual reality head-mounted displays},
  author={Pidhorskyi, Stanislav and Morehead, Michael and Jones, Quinn and Spirou, George and Doretto, Gianfranco},
  journal={arXiv preprint arXiv:1804.08197},
  year={2018}
}

@article{bouthillier1996segments,
  title={Segments of the internal carotid artery: a new classification},
  author={Bouthillier, Alain and Van Loveren, Harry R and Keller, Jeffrey T},
  journal={Neurosurgery},
  volume={38},
  number={3},
  pages={425--433},
  year={1996},
  publisher={LWW}
}

@article{kaltenecker2023virtual,
  title={Virtual reality-empowered deep-learning analysis of brain cells},
  author={Kaltenecker, Doris and Al-Maskari, Rami and Negwer, Moritz and
  Hoeher, Luciano and Kofler, Florian and Zhao, Shan and Todorov,
  Mihail and Rong, Zhouyi and Paetzold, Johannes Christian and
  Wiestler, Benedikt and Piraud, Marie and Rueckert, Daniel and
  Geppert, Julia and Morigny, Pauline and Rohm, Maria and Menze,
  Bjoern H and Herzig, Stephan and Berriel Diaz, Mauricio and
  Ert{\"u}rk, Ali},
  journal={Nature Methods},
  pages={1--10},
  year={2024},
  publisher={Nature Publishing Group US New York}
}

@article{rordendcm2niix,
  title={The first step for neuroimaging data analysis: DICOM to NIfTI conversion},
  author={Li, Xiangrui and Morgan, Paul S and Ashburner, John and Smith, Jolinda and Rorden, Christopher},
  journal={Journal of Neuroscience Methods},
  volume={264},
  pages={47--56},
  year={2016},
  publisher={Elsevier}
}

@inproceedings{menten2023skeletonization,
  title={A skeletonization algorithm for gradient-based optimization},
  author={Menten, Martin J and Paetzold, Johannes C and Zimmer, Veronika A and Shit, Suprosanna and Ezhov, Ivan and Holland, Robbie and Probst, Monika and Schnabel, Julia A and Rueckert, Daniel},
  booktitle={Proceedings of the IEEE/CVF International Conference on Computer Vision (ICCV)},
  pages={21394--21403},
  year={2023}
}

@article{bogunovic2013anatomical,
  title={Anatomical labeling of the circle of willis using maximum a posteriori probability estimation},
  author={Bogunovi{\'c}, Hrvoje and Pozo, Jos{\'e} Mar{\'\i}a and C{\'a}rdenes, Rub{\'e}n and San Rom{\'a}n, Luis and Frangi, Alejandro F},
  journal={IEEE Transactions on Medical Imaging},
  volume={32},
  number={9},
  pages={1587--1599},
  year={2013},
  publisher={IEEE}
}

@inproceedings{robben2013anatomical,
  title={Anatomical labeling of the Circle of Willis using maximum a posteriori graph matching},
  author={Robben, David and Sunaert, Stefan and Thijs, Vincent and Wilms, Guy and Maes, Frederik and Suetens, Paul},
  booktitle={Medical Image Computing and Computer-Assisted Intervention--MICCAI 2013: 16th International Conference, Nagoya, Japan, September 22-26, 2013, Proceedings, Part I 16},
  pages={566--573},
  year={2013},
  organization={Springer}
}

@article{robben2016simultaneous,
  title={Simultaneous segmentation and anatomical labeling of the cerebral vasculature},
  author={Robben, David and T{\"u}retken, Engin and Sunaert, Stefan and Thijs, Vincent and Wilms, Guy and Fua, Pascal and Maes, Frederik and Suetens, Paul},
  journal={Medical Image Analysis},
  volume={32},
  pages={201--215},
  year={2016},
  publisher={Elsevier}
}

@inproceedings{chen2020automated,
  title={Automated intracranial artery labeling using a graph neural network and hierarchical refinement},
  author={Chen, Li and Hatsukami, Thomas and Hwang, Jenq-Neng and Yuan, Chun},
  booktitle={Medical Image Computing and Computer Assisted Intervention--MICCAI 2020: 23rd International Conference, Lima, Peru, October 4--8, 2020, Proceedings, Part VI 23},
  pages={76--85},
  year={2020},
  organization={Springer}
}

@article{liebeskind2003collateral,
  title={Collateral circulation},
  author={Liebeskind, David S},
  journal={Stroke},
  volume={34},
  number={9},
  pages={2279--2284},
  year={2003},
  publisher={Am Heart Assoc}
}

@article{kim2016BMJ,
  title={Clinical significance of the circle of Willis in intracranial atherosclerotic stenosis},
  author={Kim, Kang Min and Kang, Hyun-Seung and Lee, Woong Jae and Cho, Young Dae and Kim, Jeong Eun and Han, Moon Hee},
  journal={Journal of Neurointerventional Surgery},
  volume={8},
  number={3},
  pages={251--255},
  year={2016},
  publisher={British Medical Journal Publishing Group}
}

@article{hong2023automated,
  title={Automated in-depth cerebral arterial labelling using cerebrovascular vasculature reframing and deep neural networks},
  author={Hong, Suk-Woo and Song, Ha-Na and Choi, Jong-Un and Cho, Hwan-Ho and Baek, In-Young and Lee, Ji-Eun and Kim, Yoon-Chul and Chung, Darda and Chung, Jong-Won and Bang, Oh-Young and others},
  journal={Scientific Reports},
  volume={13},
  number={1},
  pages={3255},
  year={2023},
  publisher={Nature Publishing Group UK London}
}

@article{dumais2022eicab,
  title={eICAB: A novel deep learning pipeline for Circle of Willis multiclass segmentation and analysis},
  author={Dumais, F{\'e}lix and Caceres, Marco Perez and Janelle, F{\'e}lix and Seifeldine, Kassem and Ar{\`e}s-Bruneau, No{\'e}mie and Gutierrez, Jose and Bocti, Christian and Whittingstall, Kevin},
  journal={NeuroImage},
  volume={260},
  pages={119425},
  year={2022},
  publisher={Elsevier}
}

@inproceedings{kirchhoff2024skeleton,
  title={Skeleton recall loss for connectivity conserving and resource efficient segmentation of thin tubular structures},
  author={Kirchhoff, Yannick and Rokuss, Maximilian R and Roy, Saikat and Kovacs, Balint and Ulrich, Constantin and Wald, Tassilo and Zenk, Maximilian and Vollmuth, Philipp and Kleesiek, Jens and Isensee, Fabian and others},
  booktitle={European Conference on Computer Vision},
  pages={218--234},
  year={2024},
  organization={Springer}
}

@article{shi2023nextou,
  title={NexToU: Efficient Topology-Aware U-Net for Medical Image Segmentation},
  author={Shi, Pengcheng and Guo, Xutao and Yang, Yanwu and Ye, Chenfei and Ma, Ting},
  journal={arXiv preprint arXiv:2305.15911},
  year={2023}
}

@inproceedings{shi2024cbDice,
  title={Centerline boundary dice loss for vascular segmentation},
  author={Shi, Pengcheng and Hu, Jiesi and Yang, Yanwu and Gao, Zilve and Liu, Wei and Ma, Ting},
  booktitle={International Conference on Medical Image Computing and Computer-Assisted Intervention},
  pages={46--56},
  year={2024},
  organization={Springer}
}

@article{zhang2023towards,
  title={Towards connectivity-aware pulmonary airway segmentation},
  author={Zhang, Minghui and Gu, Yun},
  journal={IEEE Journal of Biomedical and Health Informatics},
  volume={28},
  number={1},
  pages={321--332},
  year={2023},
  publisher={IEEE}
}

@article{vos2025evaluation,
  title={Evaluation of techniques for automated classification and artery quantification of the circle of Willis on TOF-MRA images: The CROWN challenge},
  author={Vos, Iris N and Ruigrok, Ynte M and Bennink, Edwin and Velthuis, Mireille RE and Paic, Barbara and Ophelders, Maud EH and Buser, Myrthe AD and van der Velden, Bas HM and Chen, Geng and Coupet, Matthieu and others},
  journal={Medical Image Analysis},
  pages={103650},
  year={2025},
  publisher={Elsevier}
}

@article{de2024isles,
  title={ISLES'24: Final Infarct Prediction with Multimodal Imaging and Clinical Data. Where Do We Stand?},
  author={{de la Rosa}, Ezequiel and Su, Ruisheng and Reyes, Mauricio and Riedel, Evamaria O and Baazaoui, Hakim and Wiest, Roland and Kofler, Florian and Yang, Kaiyuan and Robben, David and Mojtahedi, Mahsa and others},
  journal={arXiv preprint arXiv:2408.10966},
  year={2024}
}

@dataset{LargeIAzenodo,
  author       = {Bo, Zi-Hao},
  title        = {Large IA Segmentation dataset},
  month        = feb,
  year         = 2021,
  publisher    = {Zenodo},
  doi          = {10.5281/zenodo.6801398},
  url          = {https://doi.org/10.5281/zenodo.6801398},
}

@article{LargeIAcellpattern2021,
  title={Toward human intervention-free clinical diagnosis of intracranial aneurysm via deep neural network},
  author={Bo, Zi-Hao and Qiao, Hui and Tian, Chong and Guo, Yuchen and Li, Wuchao and Liang, Tiantian and Li, Dongxue and Liao, Dan and Zeng, Xianchun and Mei, Leilei and others},
  journal={Patterns},
  volume={2},
  number={2},
  pages={100197},
  year={2021},
  publisher={Elsevier}
}

@dataset{LausanneOpenNeuro,
  author = {Tommaso {Di Noto} and Guillaume Marie and Sebastien Tourbier and Yasser Alemán-Gómez and Oscar Esteban and Guillaume Saliou and Meritxell Bach Cuadra and Patric Hagmann and Jonas Richiardi},
  title = {Lausanne TOF-MRA Aneurysm Cohort},
  year = {2022},
  doi = {doi:10.18112/openneuro.ds003949.v1.0.1},
  publisher = {OpenNeuro}
}

@article{LausanneNeuroInfoPaper2023,
  title={Towards automated brain aneurysm detection in TOF-MRA: open data, weak labels, and anatomical knowledge},
  author={Di Noto, Tommaso and Marie, Guillaume and Tourbier, Sebastien and Alem{\'a}n-G{\'o}mez, Yasser and Esteban, Oscar and Saliou, Guillaume and Cuadra, Meritxell Bach and Hagmann, Patric and Richiardi, Jonas},
  journal={Neuroinformatics},
  volume={21},
  number={1},
  pages={21--34},
  year={2023},
  publisher={Springer}
}

@misc{IXIdataset,
  title = {{IXI} Dataset - Brain Development},
  author = "{IXI}",
  howpublished = {\url{https://brain-development.org/ixi-dataset/}},
  year = {2022},
  note = {Accessed: 2022-09-30}
}

@inproceedings{musio2024CoWGraph,
  title={Quantitative evaluation of the Circle of Willis vascular architecture in 3D CT and MR angiography},
  author={Musio, Fabio and Yang, Kaiyuang and Shit, Suprosanna and Prabhakar, Chinmay and Juchler, Norman and Menze, Bjoern and Hirsch, Sven},
  booktitle={8th International Conference on Computational and Mathematical Biomedical Engineering (CMBE24), Arlington, VA, USA, 24-26 June 2024},
  volume={2},
  pages={563--566},
  year={2024},
  organization={Computational \& Mathematical Biomedical Engineering}
}

@inproceedings{shit2021cldice,
  title={cl{D}ice-a novel topology-preserving loss function for tubular structure segmentation},
  author={Shit, Suprosanna and Paetzold, Johannes C and Sekuboyina, Anjany and Ezhov, Ivan and Unger, Alexander and Zhylka, Andrey and Pluim, Josien PW and Bauer, Ulrich and Menze, Bjoern H},
  booktitle={Proceedings of the IEEE/CVF Conference on Computer Vision and Pattern Recognition (CVPR)},
  pages={16560--16569},
  year={2021}
}

@inproceedings{cheng2021boundaryiou,
  title={Boundary IoU: Improving object-centric image segmentation evaluation},
  author={Cheng, Bowen and Girshick, Ross and Doll{\'a}r, Piotr and Berg, Alexander C and Kirillov, Alexander},
  booktitle={Proceedings of the IEEE/CVF conference on computer vision and pattern recognition},
  pages={15334--15342},
  year={2021}
}

@article{reinke2024understanding,
  title={Understanding metric-related pitfalls in image analysis validation},
  author={Reinke, Annika and Tizabi, Minu D and Baumgartner, Michael and Eisenmann, Matthias and Heckmann-N{\"o}tzel, Doreen and Kavur, A Emre and R{\"a}dsch, Tim and Sudre, Carole H and Acion, Laura and Antonelli, Michela and others},
  journal={Nature methods},
  volume={21},
  number={2},
  pages={182--194},
  year={2024},
  publisher={Nature Publishing Group US New York}
}

@article{celaya2023generalized,
  title={A generalized surface loss for reducing the hausdorff distance in medical imaging segmentation},
  author={Celaya, Adrian and Riviere, Beatrice and Fuentes, David},
  journal={arXiv preprint arXiv:2302.03868},
  year={2023}
}

@inproceedings{acebes2024centerline,
  title={The Centerline-Cross Entropy Loss for Vessel-Like Structure Segmentation: Better Topology Consistency Without Sacrificing Accuracy},
  author={Acebes, Cesar and Moustafa, Abdel Hakim and Camara, Oscar and Galdran, Adrian},
  booktitle={International Conference on Medical Image Computing and Computer-Assisted Intervention},
  pages={710--720},
  year={2024},
  organization={Springer}
}

@article{nader2024vascular,
  title={A vascular synthetic model for improved aneurysm segmentation and detection via Deep Neural Networks},
  author={Nader, Rafic and Autrusseau, Florent and L'Allinec, Vincent and Bourcier, Romain},
  journal={arXiv preprint arXiv:2403.18734},
  year={2024}
}

@inproceedings{autrusseau2022toward,
  title={Toward a 3d arterial tree bifurcation model for intra-cranial aneurysm detection and segmentation},
  author={Autrusseau, Florent and Nader, Rafic and Nouri, Anass and L’Allinec, Vincent and Bourcier, Romain},
  booktitle={2022 26th International Conference on Pattern Recognition (ICPR)},
  pages={4500--4506},
  year={2022},
  organization={IEEE}
}

@article{dutta2024segmentation,
	title={Are Vision xLSTM Embedded UNet More Reliable in Medical 3D Image Segmentation?},
	author={Dutta, Pallabi and Bose, Soham and Roy, Swalpa Kumar and Mitra, Sushmita},
	url={https://arxiv.org/abs/2406.16993}, 
  	journal={arXiv},
	pp.={1-9},
	year={2024}
}

@article{oquab2023dinov2,
  title={Dinov2: Learning robust visual features without supervision},
  author={Oquab, Maxime and Darcet, Timoth{\'e}e and Moutakanni, Th{\'e}o and Vo, Huy and Szafraniec, Marc and Khalidov, Vasil and Fernandez, Pierre and Haziza, Daniel and Massa, Francisco and El-Nouby, Alaaeldin and others},
  journal={arXiv preprint arXiv:2304.07193},
  year={2023}
}

@article{cardoso2022monai,
  title={Monai: An open-source framework for deep learning in healthcare},
  author={Cardoso, M Jorge and Li, Wenqi and Brown, Richard and Ma, Nic and Kerfoot, Eric and Wang, Yiheng and Murrey, Benjamin and Myronenko, Andriy and Zhao, Can and Yang, Dong and others},
  journal={arXiv preprint arXiv:2211.02701},
  year={2022}
}

@inproceedings{myronenko20183d,
  title={3D MRI brain tumor segmentation using autoencoder regularization},
  author={Myronenko, Andriy},
  booktitle={International MICCAI brainlesion workshop},
  pages={311--320},
  year={2018},
  organization={Springer}
}

@misc{monai2020,
  author={The MONAI Consortium},
  title={Project MONAI},
  year={2020},
  howpublished={\url{https://doi.org/10.5281/zenodo.4323059}},
  note={Zenodo}
}

@article{drees2021scalable,
  title={Scalable robust graph and feature extraction for arbitrary vessel networks in large volumetric datasets},
  author={Drees, Dominik and Scherzinger, Aaron and H{\"a}gerling, Ren{\'e} and Kiefer, Friedemann and Jiang, Xiaoyi},
  journal={BMC bioinformatics},
  volume={22},
  number={1},
  pages={346},
  year={2021},
  publisher={Springer}
}

@article{meyer2009voreen,
  title={Voreen: A rapid-prototyping environment for ray-casting-based volume visualizations},
  author={Meyer-Spradow, Jennis and Ropinski, Timo and Mensmann, J{\"o}rg and Hinrichs, Klaus},
  journal={IEEE Computer Graphics and Applications},
  volume={29},
  number={6},
  pages={6--13},
  year={2009},
  publisher={IEEE}
}

@software{Jocher_Ultralytics_YOLO_2023,
    author = {Jocher, Glenn and Qiu, Jing and Chaurasia, Ayush},
    license = {AGPL-3.0},
    month = jan,
    title = {{Ultralytics YOLO}},
    url = {https://github.com/ultralytics/ultralytics},
    version = {8.0.0},
    year = {2023}
}

@article{imr2024topology,
  title={Topology-Aware Exploration of Circle of Willis for CTA and MRA: Segmentation, Detection, and Classification},
  author={Zhang, Minghui and You, Xin and Zhang, Hanxiao and Gu, Yun},
  journal={arXiv preprint arXiv:2410.15614},
  year={2024}
}

@inproceedings{liu2022VideoSwinT,
  title={Video swin transformer},
  author={Liu, Ze and Ning, Jia and Cao, Yue and Wei, Yixuan and Zhang, Zheng and Lin, Stephen and Hu, Han},
  booktitle={Proceedings of the IEEE/CVF conference on computer vision and pattern recognition},
  pages={3202--3211},
  year={2022}
}

@inproceedings{karras2019styleGan,
  title={A style-based generator architecture for generative adversarial networks},
  author={Karras, Tero and Laine, Samuli and Aila, Timo},
  booktitle={Proceedings of the IEEE/CVF conference on computer vision and pattern recognition},
  pages={4401--4410},
  year={2019}
}

@inproceedings{redmon2016yolo,
  title={You only look once: Unified, real-time object detection},
  author={Redmon, Joseph and Divvala, Santosh and Girshick, Ross and Farhadi, Ali},
  booktitle={Proceedings of the IEEE conference on computer vision and pattern recognition},
  pages={779--788},
  year={2016}
}

@inproceedings{baumgartner2021nndetection,
  title={nnDetection: a self-configuring method for medical object detection},
  author={Baumgartner, Michael and J{\"a}ger, Paul F and Isensee, Fabian and Maier-Hein, Klaus H},
  booktitle={Medical Image Computing and Computer Assisted Intervention--MICCAI 2021: 24th International Conference, Strasbourg, France, September 27--October 1, 2021, Proceedings, Part V 24},
  pages={530--539},
  year={2021},
  organization={Springer}
}

@inproceedings{ctatlas2020ubvtl,
  title={A publicly available, high resolution, unbiased CT brain template},
  author={Muschelli, John},
  booktitle={Information Processing and Management of Uncertainty in Knowledge-Based Systems: 18th International Conference, IPMU 2020, Lisbon, Portugal, June 15--19, 2020, Proceedings, Part III 18},
  pages={358--366},
  year={2020},
  organization={Springer}
}

@article{warfield2004staple,
  title={Simultaneous truth and performance level estimation (STAPLE): an algorithm for the validation of image segmentation},
  author={Warfield, Simon K and Zou, Kelly H and Wells, William M},
  journal={IEEE transactions on medical imaging},
  volume={23},
  number={7},
  pages={903--921},
  year={2004},
  publisher={IEEE}
}

@article{mou2024costa,
  title={COSTA: A Multi-center TOF-MRA Dataset and A Style Self-Consistency Network for Cerebrovascular Segmentation},
  author={Mou, Lei and Yan, Qifeng and Lin, Jinghui and Zhao, Yifan and Liu, Yonghuai and Ma, Shaodong and Zhang, Jiong and Lv, Wenhao and Zhou, Tao and Frangi, Alejandro F and others},
  journal={IEEE transactions on medical imaging},
  year={2024},
  publisher={IEEE}
}

@misc{CAS2023,
  title = {Cerebral artery segmentation Challenge ({CAS}) 2023},
  author = "{CAS2023}",
  howpublished = {\url{https://codalab.lisn.upsaclay.fr/competitions/9804}},
  year = {2023},
  note = {Accessed: 2023-10-01}
}

@article{chatterjee2024smile,
  title={SMILE-UHURA Challenge--Small Vessel Segmentation at Mesoscopic Scale from Ultra-High Resolution 7T Magnetic Resonance Angiograms},
  author={Chatterjee, Soumick and Mattern, Hendrik and D{\"o}rner, Marc and Sciarra, Alessandro and Dubost, Florian and Schnurre, Hannes and Khatun, Rupali and Yu, Chun-Chih and Hsieh, Tsung-Lin and Tsai, Yi-Shan and others},
  journal={arXiv preprint arXiv:2411.09593},
  year={2024}
}

@article{hindenes2023anatomical,
  title={Anatomical variations in the circle of Willis are associated with increased odds of intracranial aneurysms: The Troms{\o} study},
  author={Hindenes, Lars B and Ingebrigtsen, Tor and Isaksen, J{\o}rgen G and H{\aa}berg, Asta K and Johnsen, Liv-Hege and Herder, Marit and Mathiesen, Ellisiv B and Vangberg, Torgil R},
  journal={Journal of the Neurological Sciences},
  volume={452},
  pages={120740},
  year={2023},
  publisher={Elsevier}
}

@article{dice1945measures,
  title={Measures of the amount of ecologic association between species},
  author={Dice, Lee R},
  journal={Ecology},
  volume={26},
  number={3},
  pages={297--302},
  year={1945},
  publisher={JSTOR}
}

@article{taha2015metrics,
  title={Metrics for evaluating 3D medical image segmentation: analysis, selection, and tool},
  author={Taha, Abdel Aziz and Hanbury, Allan},
  journal={BMC medical imaging},
  volume={15},
  number={1},
  pages={29},
  year={2015},
  publisher={Springer}
}

@article{maier2024metrics,
  title={Metrics reloaded: recommendations for image analysis validation},
  author={Maier-Hein, Lena and Reinke, Annika and Godau, Patrick and Tizabi, Minu D and Buettner, Florian and Christodoulou, Evangelia and Glocker, Ben and Isensee, Fabian and Kleesiek, Jens and Kozubek, Michal and others},
  journal={Nature methods},
  volume={21},
  number={2},
  pages={195--212},
  year={2024},
  publisher={Nature Publishing Group US New York}
}

@article{hu2019topology,
  title={Topology-preserving deep image segmentation},
  author={Hu, Xiaoling and Li, Fuxin and Samaras, Dimitris and Chen, Chao},
  journal={Advances in neural information processing systems},
  volume={32},
  year={2019}
}

@article{riedel2026ischemic,
  title={The Ischemic Stroke Lesion Segmentation Challenge (ISLES)’24 Dataset: A Multimodal Stroke Imaging Dataset with Hyperacute CT, Acute Postinterventional MRI, and 3-month Clinical Outcomes},
  author={Riedel, Evamaria Olga and {de la Rosa}, Ezequiel and Baran, The Anh and Hernandez Petzsche, Moritz and Baazaoui, Hakim and Yang, Kaiyuan and Musio, Fabio Antonio and Huang, Houjing and Robben, David and Seia, Joaquin Oscar and others},
  journal={Radiology: Artificial Intelligence},
  volume={8},
  number={3},
  pages={e250603},
  year={2026},
  publisher={Radiological Society of North America}
}

@inproceedings{ceballos2024vessel,
  title={Vessel-aware aneurysm detection using multi-scale deformable 3D attention},
  author={Ceballos-Arroyo, Alberto M and Nguyen, Hieu T and Zhu, Fangrui and Yadav, Shrikanth M and Kim, Jisoo and Qin, Lei and Young, Geoffrey and Jiang, Huaizu},
  booktitle={International Conference on Medical Image Computing and Computer-Assisted Intervention},
  pages={754--765},
  year={2024},
  organization={Springer}
}

@article{musio2026circle,
  title={Circle of Willis Centerline Graphs: A Dataset and Baseline Algorithm},
  author={Musio, Fabio and Juchler, Norman and Yang, Kaiyuan and Shit, Suprosanna and Prabhakar, Chinmay and Menze, Bjoern and Hirsch, Sven},
  journal={Neuroscience Informatics},
  pages={100265},
  year={2026},
  publisher={Elsevier}
}

@article{hamadache2026topology,
  title={Topology-aware multiclass segmentation of the Circle of Willis from MRA and CTA images},
  author={Hamadache, Rachika E and Lisazo, Clara and Yalcin, Cansu and Estrada, Uma M Lal-Trehan and Abramova, Valeriia and Casamitjana, Adri{\`a} and Oliver, Arnau and Llad{\'o}, Xavier},
  journal={Computers in Biology and Medicine},
  volume={204},
  pages={111516},
  year={2026},
  publisher={Elsevier}
}

@inproceedings{gupta2022learning,
  title={Learning topological interactions for multi-class medical image segmentation},
  author={Gupta, Saumya and Hu, Xiaoling and Kaan, James and Jin, Michael and Mpoy, Mutshipay and Chung, Katherine and Singh, Gagandeep and Saltz, Mary and Kurc, Tahsin and Saltz, Joel and others},
  booktitle={European Conference on Computer Vision},
  pages={701--718},
  year={2022},
  organization={Springer}
}

@incollection{baumgartner2022accurate,
  title={Accurate Detection of Mediastinal Lesions with nnDetection},
  author={Baumgartner, Michael and Full, Peter M and Maier-Hein, Klaus H},
  booktitle={MICCAI Challenge on Correction of Brainshift with Intra-Operative Ultrasound},
  pages={79--85},
  year={2022},
  publisher={Springer}
}
